\documentclass[Afour,sageh,times]{sagej}

\renewcommand{\journalname}{}
\renewcommand{\volumeyear}{}
\renewcommand{\volumenumber}{}
\renewcommand{\issuenumber}{}
\renewcommand{\startpage}{}
\renewcommand{\endpage}{}
\makeatletter
\renewcommand{\@maketitle}{%
\vspace*{-34pt}%
\null%
\begin{center}
\if@PCfour
\begin{rm}
\else
\begin{sf}
\fi
\begin{minipage}[t]{\textwidth}
  \vskip 12.5pt%
    {\raggedright\titlesize\textbf{\@title} \par}%
    \vskip 1.5em%
    \vskip 12.5mm%
    \end{minipage}
{\par\large%
      \lineskip .5em%
      {\raggedright\textbf{\@author}
      \par}}
     \vskip 40pt%
    {\noindent\usebox\absbox\par}
    {\vspace{20pt}%
      {\noindent\normalsize\@keywords}\par}
      \if@PCfour
      \end{rm}
      \else
      \end{sf}
      \fi
      \end{center}
      \vspace{22pt}
        \par%
  }
\makeatother

\makeatletter
\def\ps@title{%
  \def\@oddhead{\parbox{\textwidth}{\mbox{}\\[-1pt]%
    \noindent\rule{\textwidth}{0.5pt}}}%
  \let\@evenhead\@oddhead
  \def\@oddfoot{}%
  \let\@evenfoot\@oddfoot
}

\def\ps@sagepage{%
  \let\@mkboth\@gobbletwo
  \def\@evenhead{\parbox{\textwidth}{}}%
  \def\@oddhead{\parbox{\textwidth}{}}%
  \def\@evenfoot{}%
  \def\@oddfoot{}%
}
\makeatother

\usepackage{moreverb, url}
\usepackage{subfigure}
\usepackage{arydshln}
\usepackage{booktabs}

\usepackage[toc]{appendix}
\usepackage{algorithm}
\usepackage{algorithmic}

\usepackage[
  colorlinks,
  bookmarksopen,
  bookmarksnumbered,
  citecolor=red,
  urlcolor=red
]{hyperref}

\newcommand{\BibTeX}{{\rmfamily B\kern-.05em \textsc{i\kern-.025em b}\kern-.08em T\kern-.1667em\lower.7ex\hbox{E}\kern-.125emX}}

\def\volumeyear{2025} 

\begin{document}
  \runninghead{F. Yang, P. Frivik, D. Hoeller, C. Wang, C. Cadena, and M. Hutter}


  \title{Spatially-Enhanced Recurrent Memory for Long-Range Mapless Navigation via End-to-End Reinforcement Learning}

  \author{Fan Yang\affilnum{1}, Per Frivik\affilnum{1}, David Hoeller\affilnum{1},
  Chen Wang\affilnum{2}, Cesar Cadena\affilnum{1}, and Marco Hutter\affilnum{1}}

  \affiliation{\affilnum{1}Robotic Systems Lab, ETH Zurich, Zurich, Switzerland\\ \affilnum{2}Spatial AI \& Robotics Lab, University at Buffalo, Buffalo, NY, USA}

  \corrauth{Fan Yang, fanyang1@ethz.ch}

  \begin{abstract}
    Recent advancements in robot navigation, particularly with end-to-end learning approaches such as reinforcement learning (RL), have demonstrated remarkable efficiency and effectiveness. However, successful navigation still depends on two key capabilities: mapping and planning, whether implemented explicitly or implicitly. Classical approaches rely on explicit mapping pipelines to transform and register egocentric observations into a coherent map for the planning module. In contrast, end-to-end learning often achieves this implicitly—through recurrent neural networks (RNNs) that fuse current and historical observations into a latent space for planning. While existing architectures, such as LSTM and GRU, can capture temporal dependencies, our findings reveal a critical limitation: their inability to effectively perform spatial memorization. This capability is essential for transforming and integrating sequential observations from varying perspectives to build spatial representations that support planning tasks. To address this, we propose Spatially-Enhanced Recurrent Units (SRUs)—a simple yet effective modification to existing RNNs—that enhance spatial memorization. To improve navigation performance, we introduce an attention-based network architecture integrated with SRUs, enabling long-range mapless navigation using a single forward-facing stereo camera. Additionally, we employ regularization techniques to facilitate robust end-to-end recurrent training via RL. Experimental results demonstrate that our approach improves long-range navigation performance by 23.5\% overall compared to existing RNNs. Furthermore, when equipped with SRU memory, our method outperforms both RL baseline approaches—one relying on explicit mapping and the other on stacked historical observations—achieving overall improvements of 29.6\% and 105.0\%, respectively, in diverse environments that require long-horizon mapping and memorization capabilities. Finally, we address the sim-to-real gap by leveraging large-scale pretraining on synthetic depth data, enabling zero-shot transfer for deployment across diverse and complex real-world environments.
  \end{abstract}

  \keywords{Spatial Memory, End-to-End Mapless Navigation, Recurrent Neural Networks, Reinforcement Learning}

  \maketitle  

  \section{Introduction}
  End-to-end learning for robot navigation has recently gained significant attention with its potential to address two major challenges inherent in classical modular approaches: (a) system delays and (b) the difficulty of modeling complex kinodynamic environmental interactions. 
  These challenges have traditionally hindered the development of high-speed platforms with intricate dynamics, such as legged-wheeled robots. 
  However, end-to-end learning approaches face their own challenges, particularly in achieving efficient spatial mapping.
  Unlike classical mapping pipelines, which explicitly transform historical ego-centric observations into a coherent map frame for downstream planning, end-to-end learning relies on neural networks to implicitly learn this process. 
  This requires the network to iteratively build and update an environmental representation of the surroundings and understand the spatial-temporal relationships between observations.

  In autonomous driving, large-scale mapping modules~\citep{mohajerin2019multi, mescheder2019occupancy, wei2023surroundocc, wang2025uniocc} are trained on thousands of hours of data, enabling robust spatial mapping with specifically designed architectures, such as occupancy networks~\citep{mescheder2019occupancy} or occupancy grid maps~\citep{mohajerin2019multi}. However, such approaches are not easily deployable on smaller robotic platforms and often struggle to generalize to environments beyond structured road networks. In contrast, embedded robots often rely on end-to-end learning approaches, either by imitating behaviors from datasets~\citep{shah2023vint, karnan2022voila, cesar2021improving, loquercio2021learning} or by optimizing policies through reinforcement learning (RL)~\citep{wijmans2019dd, zhu2017target, surmann2020deep}. These methods typically employ specific network architectures, such as recurrent neural networks (RNNs), to implicitly learn spatial-temporal mappings~\citep{wijmans2019dd, wijmans2023emergence}. While these approaches have demonstrated success in structured indoor environments with discretized action and observation spaces, their performance often diminishes in more complex, real-world scenarios that involve continuous action spaces and dynamic motions.

  Recently, for real-world deployments, researchers have started integrating explicit mapping pipelines~\citep{miki2022elevation}
  to fuse ego-centric observations and provide environmental information to learning modules for tasks such as perceptive locomotion~\citep{miki2022learning} and navigation~\citep{lee2024learning, francis2020long, weerakoon2022terp}. This raises an important question: can end-to-end learning networks with implicit memory mechanisms, such as RNNs, match or surpass the performance of approaches that rely on explicit mapping pipelines? Specifically, do RNNs have inherent limitations in learning spatial-temporal mappings?

  While RNNs excel at capturing temporal dependencies, showcased by their success in various sequential tasks, such as natural language processing~\citep{sutskever2014sequence} and time-series prediction~\citep{siami2019performance}, their ability to learn spatial transformations and memorization remains a topic of research. RNNs are designed to process sequences of data by maintaining an internal state that captures temporal dependencies. 
  However, it is not yet clear to what extent they can effectively learn spatial transformations and integrate observations from different perspectives. 
  Classical approaches achieve spatial registration through homogeneous transformations in three-dimensional space, aligning observations into a consistent local or global frame. 
  For RNNs to achieve effective spatial registration, they must not only memorize sequences but also learn to transform and integrate observations across time and space.

  In this work, we examine the spatial-temporal memory capabilities of several recurrent architectures, including Long Short-Term Memory (LSTM)~\citep{hochreiter1997long}, Gated Recurrent Unit (GRU)~\citep{cho2014learning}, and recent State-Space Models (SSMs) such as S4~\citep{gu2021efficiently} and Mamba-SSM~\citep{mamba}. 
  We evaluate these models on two criteria: (i) their ability to memorize temporal sequences and (ii) their capacity to register and transform sequential observations across varying spatial perspectives. Our findings indicate that, while these models perform well in capturing temporal dependencies, they exhibit limitations in spatial registration, particularly under conditions of dynamic ego-motion and rapidly changing perspectives.

  To address this limitation, we introduce Spatially-Enhanced Recurrent Units (SRUs), a simple yet effective modification to standard LSTM and GRU units that enhances their spatial registration capabilities when processing sequences of ego-centric observations. Unlike classical mapping pipelines that rely on explicit homogeneous transformations, our approach enables the recurrent units to implicitly learn the transformations from varying observation perspectives effectively. To further enhance the performance of long-range navigation tasks, we propose an attention-based network architecture integrated with SRUs, allowing the model to learn long-range mapless navigation policies using only ego-centric observations via end-to-end RL. Our experiments demonstrate improvements in spatial awareness compared to the baselines. With the SRU memory, the implicit recurrent approach via RL with sparse rewards promotes robust exploration in complex 3D and maze-like environments, outperforming the baseline that rely on explicit mapping and memory modules.

  To prevent premature convergence to suboptimal strategies and fully exploit the capabilities of the proposed attention-based recurrent structure, we find that incorporating regularizations during end-to-end RL training is crucial. Furthermore, to address the sim-to-real gap caused by noisy depth images, we pretrain the image encoder on a large-scale synthetic dataset and augment the data using a fully parallelized depth-noise model, adapted from \cite{handa:etal:2014, Barron:etal:2013A, Bohg:etal:2014}.
  In summary, our main contributions are as follows:

  \begin{itemize}
    \item \textbf{Addressing Spatial Mapping Limitations with SRUs:} We identify that standard RNNs, while effective in capturing temporal dependencies, can struggle with spatial registration of observations from different perspectives. 
    To overcome this, we introduce Spatially-Enhanced Recurrent Units (SRUs) that enhance the ability to learn implicit spatial transformations from sequences of ego-centric observations.

    \item \textbf{End-to-End Reinforcement Learning with SRUs and Attention-based Policy:} We integrate the SRU unit into a proposed attention-based network architecture, enabling improved end-to-end reinforcement learning for long-range mapless navigation tasks using only ego-centric observations.

    \item \textbf{Large-Scale Pretraining for Zero-Shot Sim-to-Real Transfer in Long-Range Mapless Navigation:} By leveraging large-scale synthetic pretraining and a parallelizable depth-noise model, our system bridges the sim-to-real gap, enabling zero-shot deployment on a legged-wheel platform in diverse real-world environments, using a single forward-facing stereo camera for long-range mapless navigation.
  \end{itemize}

  \section{Related Works}
  The navigation and planning problem has been studied extensively for decades.
  Early approaches relied on classic search-based methods, including Dijkstra’s algorithm and A*~\citep{dijkstra1959note, Hart1968} operating on pre-discretized grids, as well as on sample-based techniques such as the Rapidly-exploring Random Tree (RRT) family—including variants like RRT*, RRT-Connect~\citep{lavalle2001rapidly, karaman2011sampling, kuffner2000rrt}, etc.—and probabilistic roadmap (PRM) methods, such as Lazy PRM and SPARS~\citep{kavraki1996probabilistic, bohlin2000path, dobson2014sparse}.
  While these techniques have achieved significant success in robotics and real-world applications~\citep{wellhausen2023artplanner}, they depend on building or the existence of a predefined navigation or occupancy map. 
  Consequently, they often struggle in unknown or dynamic environments~\citep{yang2022far}, particularly when planning under static-world assumptions or when complex kinodynamic constraints are present~\citep{webb2012kinodynamic, ortiz2024idb}. 
  Moreover, these classical methods typically require an additional perception and mapping module, and the predetermined traversability or occupancy maps are usually based on heuristic designs rather than being optimized for a specific robotic platform.

  To address these limitations, recent research has increasingly turned to learning-based approaches, especially for more complex robotic agents (e.g., quadrupeds or wheel-legged systems). 
  For instance, recent works in imitation learning leverage large-scale video data or demonstrations~\citep{pfeiffer2017perception, bojarski2016end, loquercio2021learning, shah2022gnm, shah2023vint} to directly map raw egocentric sensory inputs to navigation actions.
  Given the challenges of capturing dynamic, closed-loop interactions from purely offline data, researchers have also explored model-free reinforcement learning (RL) methods~\citep{shi2019end, wijmans2019dd, choi2019deep, hoeller2021learning, truong2021learning, wu2021learn, ruiz2022towards, fu2022coupling, huang2023goal, bhattacharya2024vision, lee2024learning} that train navigation policies end-to-end by simulating the entire robot dynamics. 
  By replacing the traditional perception, mapping, and planning pipeline with a tailored network—such as architectures based on recurrent networks~\citep{wijmans2019dd, choi2019deep, hoeller2021learning, wu2021learn} or Transformers with attention mechanisms~\citep{ruiz2022towards, huang2023goal, bhattacharya2024vision, zeng2024poliformer}—these approaches have achieved improvements in navigation tasks as well as in robotic locomotion~\citep{miki2022learning, yang2022learning, kareer2023vinl}. 

  A key challenge with end-to-end approaches is learning a robust state representation from the partial observations provided by egocentric sensors. 
  Recent studies have attempted to mitigate this challenge by incorporating explicit mapping and memory mechanisms~\citep{savinov2018semi, cimurs2021goal, fu2022coupling, lee2024learning} or by employing specialized network architectures like RNNs~\citep{hoeller2021learning, wijmans2019dd, choi2019deep}. However, RNNs—originally designed to capture temporal sequences in language tasks~\citep{cho2014learning}—are not inherently well-suited for spatial mapping, particularly when processing sequential egocentric observations from continuously changing perspectives. For instance, while prior studies in indoor navigation have shown that spatial cues can be decoded from RNN memories, this effect is limited to employing binary contact sensing and does not extend to high-dimensional visual inputs~\citep{wijmans2023emergence}. 
  Moreover, recent findings suggest that variations in recurrent network architectures have minimal impact on the final task-level rewards achieved through reinforcement learning~\citep{duarte2023lstm}. This indicates that, despite architectural differences, some fundamental limitations may persist across these RNN units.

  In this paper, we explore a key limitation of existing RNN-based architectures in addressing partial observability—their spatial memorization capabilities—highlighting their shortcomings in learning spatial transformations and integrating observations from different perspectives. We then introduce Spatially-Enhanced Recurrent Units (SRUs) and demonstrate their effectiveness in improving long-range mapless navigation tasks with a specifically designed attention-based network structure via end-to-end reinforcement learning.
  

  \section{Problem Statement}

  Consider a robot operating in a three-dimensional (3D) environment \(E \subset \mathbb{R}^{3}\). At each time step \(t\), the robot is located at a configuration defined by its position and orientation in \(SE(3)\), and receives an observation \(o_t\) through its egocentric sensors. The navigation objective is defined as starting from an initial relative goal position \(p_1 \in \mathbb{R}^{3}\) in the robot’s egocentric frame and reaching a designated goal region \(\mathcal{G} \subset \mathbb{R}^{3}\), such that the relative goal position satisfies \(\|p_t\| < \epsilon\), where \(\epsilon > 0\) represents a specified tolerance, within a finite time horizon \(t \leq T_{\max}\). The robot follows a policy $\pi$ that maps its current state $s_{t}$ to an action $a_{t}$. However, due to the egocentric setup, the agent's current state $s_{t}$ is not fully observable from a single sensor snapshot. Formally, we model the navigation task as a Partially Observable Markov Decision Process (POMDP), characterized by the tuple:
  \[
    (\mathcal{S}, \mathcal{A}, \mathcal{T}, \mathcal{R}, \mathcal{O}, \mathcal{Z}
    , \gamma),
  \]
  with the following components:
  \begin{itemize}
    \item $\mathcal{S}$: the set of all possible states of the environment.
    \item $\mathcal{A}$: the set of actions available to the robot.
    \item $\mathcal{T}: \mathcal{S}\times \mathcal{A}\times \mathcal{S}\rightarrow
      [0,1]$: the state transition function, where $\mathcal{T}(s, a, s')$ denotes the probability of transitioning from state $s$ to state $s'$ when action $a$ is taken.
    \item $\mathcal{R}: \mathcal{S}\times \mathcal{A}\rightarrow \mathbb{R}$: the reward function, assigning a scalar reward to each state-action pair.
    \item $\mathcal{O}$: the set of all possible observations.
    \item $\mathcal{Z}: \mathcal{S}\times \mathcal{A}\times \mathcal{O}\rightarrow
      [0,1]$: the observation function, specifying the probability of receiving an observation given the current state and action.
    \item $\gamma \in [0,1]$: the discount factor that balances immediate reward and future payoffs.
  \end{itemize}
  The action $a_{t}\in \mathcal{A}$ is executed in the robot's local frame at time $t$. At each time step $t$, the robot receives an observation $o_{t}\in \mathcal{O}$ and combines it with its historical observations $\mathcal{H}_{t-1}$ to determine the current state, where $\mathcal{H}_{t-1}$ is defined as:
  \[
    \mathcal{H}_{t-1}:= \{o_{1}, o_{2}, \ldots, o_{t-1}\}.
  \]  
  We define a function $f$ that fuses current and historical observations into an estimate $\hat{s}_{t}$ of the unobservable state $s_{t}\in \mathcal{S}$:
  \[
    \hat{s}_{t}= f(o_{t}, \mathcal{H}_{t-1}).
  \]
  The policy $\pi$ then maps the estimated state $\hat{s}_{t}$ to an action $a_{t}$:
  \[
    a_{t}= \pi(\hat{s}_{t}).
  \]
  Due to the robot's ego-motion, the current observation $o_{t}$ can be captured from a different perspective or observation frame compared to historical observations in $\mathcal{H}_{t-1}$. 
  Therefore, to fuse observations from different viewpoints, the function $f$ typically involves a spatial transformation that aligns the observations into a coherent reference frame to estimate the current state $s_{t}$ for the policy $\pi$.
  In classical mapping pipelines, spatial transformations are typically achieved through homogeneous transformations that combine rotations and translations. However, in end-to-end learning approaches, the function $f$, which can be parameterized by neural network weights, is learned implicitly.


  \section{Methodology}

  \subsection{Overview}
  To tackle the long-range navigation task, we first examine and demonstrate the limitations of existing recurrent architectures (e.g., LSTM, GRU, S4, and Mamba-SSM) in a spatial-temporal memory task. 
  We then introduce the Spatially-Enhanced Recurrent Units (SRUs). 
  Next, we integrate SRUs into an attention-based network architecture to learn long-range mapless navigation via end-to-end reinforcement learning. Furthermore, we discuss the importance of incorporating regularization techniques to prevent early overfitting, which we find to be crucial to enhance SRUs' spatial memorization. Finally, we address the sim-to-real gap by pretraining the depth image encoder on large-scale synthetic depth data and incorporating a parallelizable depth-noise model, enabling zero-shot transfer to real-world environments.

  \subsection{Background: Recurrent Neural Networks}
  Recurrent Neural Networks (RNNs) are a class of neural networks designed to process sequential data by maintaining a hidden state that captures temporal dependencies. 
  Given a sequence of inputs $\mathbf{x}:= (x_{1}, x_{2}, \ldots , x_{T})$, an RNN computes a sequence of hidden states $\mathbf{h}:= (h_{1}, h_{2}, \ldots, h_{T})$ using the following recursive formula:
  \[
    h_{t}:= f(x_{t}, h_{t-1}),
  \]
  where $f$ is a function that combines the previous hidden state $h_{t-1}$ with the current input $x_{t}$ to compute the current hidden state $h_{t}$. 
  This is analogous to the function mentioned earlier, which fuses the current observation $o_{t}$ with the historical observations $\mathcal{H}_{t-1}$ into the estimated current state $\hat{s}_{t}$. 
  The hidden state $h_{t}$ captures the network's internal representation at time $t$, encoding information from the entire input sequence up to that point. 
  However, the vanilla version of RNN can suffer from gradient vanishing and explosion issues \citep{hochreiter1997long,cho2014learning}. 
  To address these problems, several variants have been proposed, including LSTM and GRU, as commonly used in sequential tasks nowadays. 
  These models introduce gating mechanisms that control the flow of information through the network, enabling better long-term memory retention and gradient flow.
  Those types of RNNs adopted gates and residual connections across temporal sequences, resulting in strong performance in various sequential tasks. 
  The standard LSTM unit takes the following form:
  \[
    \begin{aligned}
      \mathtt{i}_{\mathtt{t}} & = \sigma(\mathtt{W}_{\mathtt{x}\mathtt{i}}\mathtt{x}_{\mathtt{t}}+ \mathtt{W}_{\mathtt{h}\mathtt{i}}\mathtt{h}_{\mathtt{t}-1}+ \mathtt{b}_{\mathtt{i}}), \\
      \mathtt{f}_{\mathtt{t}} & = \sigma(\mathtt{W}_{\mathtt{x}\mathtt{f}}\mathtt{x}_{\mathtt{t}}+ \mathtt{W}_{\mathtt{h}\mathtt{f}}\mathtt{h}_{\mathtt{t}-1}+ \mathtt{b}_{\mathtt{f}}), \\
      \mathtt{o}_{\mathtt{t}} & = \sigma(\mathtt{W}_{\mathtt{x}\mathtt{o}}\mathtt{x}_{\mathtt{t}}+ \mathtt{W}_{\mathtt{h}\mathtt{o}}\mathtt{h}_{\mathtt{t}-1}+ \mathtt{b}_{\mathtt{o}}), \\
      \mathtt{g}_{\mathtt{t}} & = \tanh(\mathtt{W}_{\mathtt{x}\mathtt{g}}\mathtt{x}_{\mathtt{t}}+ \mathtt{W}_{\mathtt{h}\mathtt{g}}\mathtt{h}_{\mathtt{t}-1}+ \mathtt{b}_{\mathtt{g}}),  \\
      \mathtt{c}_{\mathtt{t}} & = \mathtt{f}_{\mathtt{t}}\odot \mathtt{c}_{\mathtt{t}-1}+ \mathtt{i}_{\mathtt{t}}\odot \mathtt{g}_{\mathtt{t}},                                  \\
      \mathtt{h}_{\mathtt{t}} & = \mathtt{o}_{\mathtt{t}}\odot \tanh(\mathtt{c}_{\mathtt{t}}),
    \end{aligned}
  \]
  and the GRU unit:
  \[
    \begin{aligned}
      \mathtt{z}_{\mathtt{t}} & = \sigma\bigl(\mathtt{W}_{\mathtt{x}\mathtt{z}}\,\mathtt{x}_{\mathtt{t}}+ \mathtt{W}_{\mathtt{h}\mathtt{z}}\,\mathtt{h}_{\mathtt{t}-1}+ \mathtt{b}_{\mathtt{z}}\bigr), \\
      \mathtt{r}_{\mathtt{t}} & = \sigma\bigl(\mathtt{W}_{\mathtt{x}\mathtt{r}}\,\mathtt{x}_{\mathtt{t}}+ \mathtt{W}_{\mathtt{h}\mathtt{r}}\,\mathtt{h}_{\mathtt{t}-1}+ \mathtt{b}_{\mathtt{r}}\bigr), \\
      \tilde{\mathtt{h}}_{\mathtt{t}} & = \tanh\bigl(\mathtt{W}_{\mathtt{x}\mathtt{h}}\,\mathtt{x}_{\mathtt{t}}+ \mathtt{W}_{\mathtt{h}\mathtt{h}}\,( \mathtt{r}_{\mathtt{t}}\odot \mathtt{h}_{\mathtt{t}-1}) + \mathtt{b}_{\mathtt{h}}\bigr), \\
      \mathtt{h}_{\mathtt{t}} & = (1 - \mathtt{z}_{\mathtt{t}})\odot \tilde{\mathtt{h}}_{\mathtt{t}}+ \mathtt{z}_{\mathtt{t}}\odot \mathtt{h}_{\mathtt{t}-1},
    \end{aligned}
  \]
  where $\mathtt{i}_{\mathtt{t}}$, $\mathtt{f}_{\mathtt{t}}$, $\mathtt{o}_{\mathtt{t}}$, and $\mathtt{g}_{\mathtt{t}}$ are the input, forget, output, and cell gate activations in LSTM, respectively. Similarly, $\mathtt{z}_{\mathtt{t}}$, $\mathtt{r}_{\mathtt{t}}$, and $\tilde{\mathtt{h}}_{\mathtt{t}}$ are the update, reset, and candidate hidden states in GRU, respectively. 
  The weights $\mathtt{W}$ and biases $\mathtt{b}$ are learnable parameters, and $\sigma$ denotes the sigmoid activation function. Note that for all the letters used in the RNN formulations, we use a ``monospaced'' font style to prevent confusion with the symbols used in the remainder of the paper.
  
  More recently, the State-Space Model~\citep{gu2021efficiently} was introduced, which is inspired by the general form of state space models widely used in control theory. 
  Such models take the following form in discrete time:
  \[
    \begin{aligned}
      \mathtt{x}_{\mathtt{t}} & = \bar{\mathtt{A}}\,\mathtt{x}_{\mathtt{t}-1}+ \bar{\mathtt{B}}\,\mathtt{u}_{\mathtt{t}}, \\
      \mathtt{y}_{\mathtt{t}} & = \bar{\mathtt{C}}\,\mathtt{x}_{\mathtt{t}}+ \bar{\mathtt{D}}\,\mathtt{u}_{\mathtt{t}},
    \end{aligned}
  \]
  where $\mathtt{x}_{\mathtt{t}}$ represents the state at time $\mathtt{t}$, $\mathtt{u}_{\mathtt{t}}$ the input, and $\mathtt{y}_{\mathtt{t}}$ the output.
  The matrices $\bar{\mathtt{A}}$, $\bar{\mathtt{B}}$, $\bar{\mathtt{C}}$, and $\bar{\mathtt{D}}$ define the system dynamics. 
  
  In \cite{gu2020hippo}, a so-called HiPPO matrix is proposed for $\bar{\mathtt{A}}$ and $\bar{\mathtt{B}}$ to optimally project historical information into the current state via a polynomial basis. 
  Although this approach has led to a new family of RNN models (e.g., S4, S5, Mamba-SSM) that excel at capturing long-term temporal dependencies, their emphasis on long temporal processing is not the focus of this paper; therefore, we omit further details on these models.

  \subsection{Spatial Mapping Limitations in RNNs}
  \label{sec:spatial-mapping-limitation-sec} Achieving long-range mapless navigation from ego-centric observations requires the robot to perform effective spatial mapping.
  In three-dimensional (3D) space, spatial mapping is commonly achieved using homogeneous transformations, which combine rotations and translations. 
  A general representation of such a transformation is expressed as:
  \[
    \begin{aligned}
      \begin{bmatrix}\mathbf{p}' \\ 1\end{bmatrix} = \begin{bmatrix}R&\mathbf{t}\\ \mathbf{0}^{\top}&1\end{bmatrix} \begin{bmatrix}\mathbf{p} \\ 1\end{bmatrix},
    \end{aligned}
  \]
  where $\mathbf{p}$ and $\mathbf{p}'$ are the coordinates of a point in the original and transformed frames, $R$ is the rotation matrix, and $\mathbf{t}$ is the translation vector. 
  In the context of RNNs, this spatial mapping and transformation is implicitly learned through the function $f$, which integrates the current observation $o_{t}$ with the historical observations $\mathcal{H}_{t-1}$ to estimate the current state $s_{t}$.

\begin{figure}
      \begin{center}
        \subfigure[Temporal Training Loss]{%
        \includegraphics[width=0.4\textwidth]{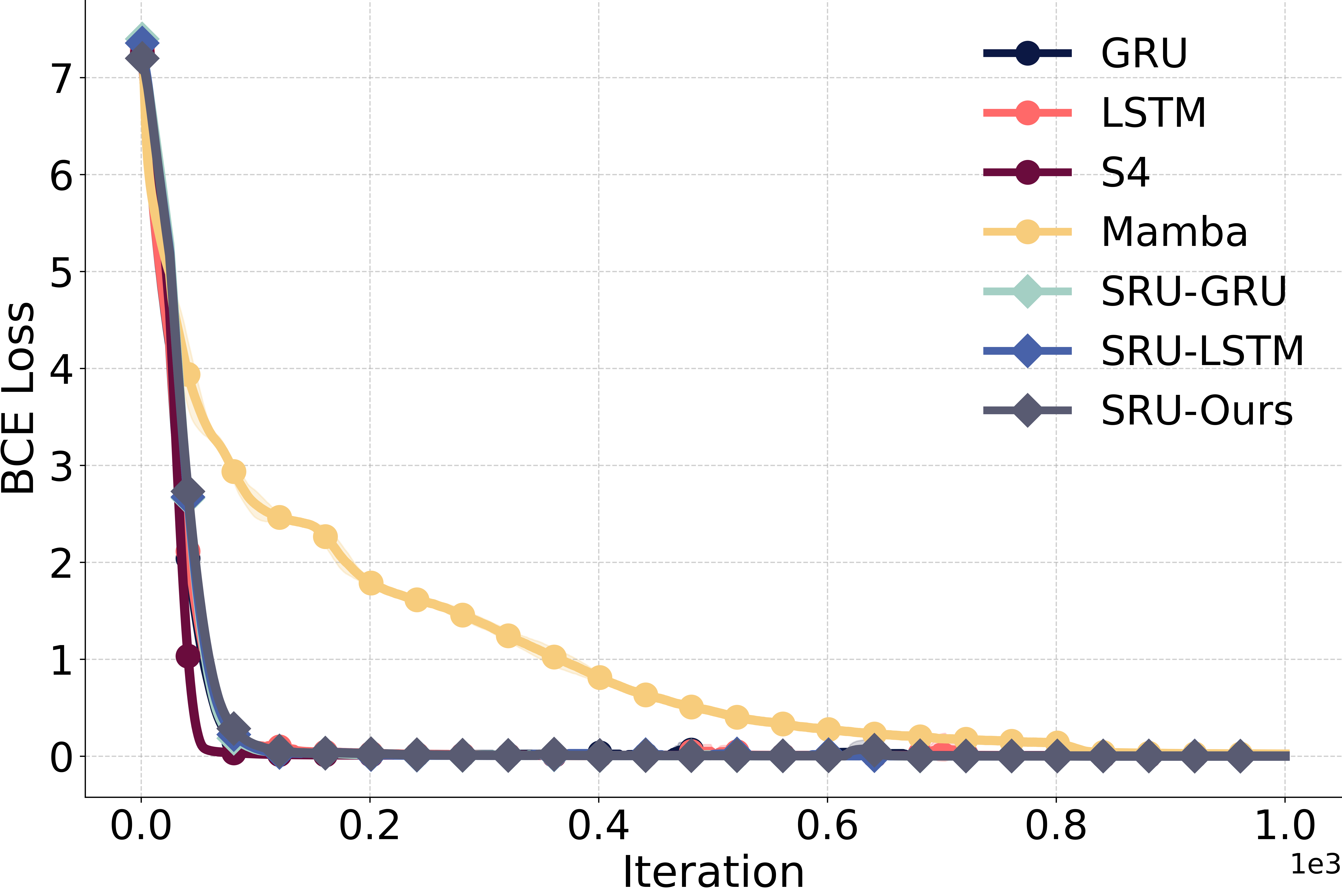}
        \label{fig:temporal_loss}%
        } \subfigure[Spatial Training Loss]{%
        \includegraphics[width=0.4\textwidth]{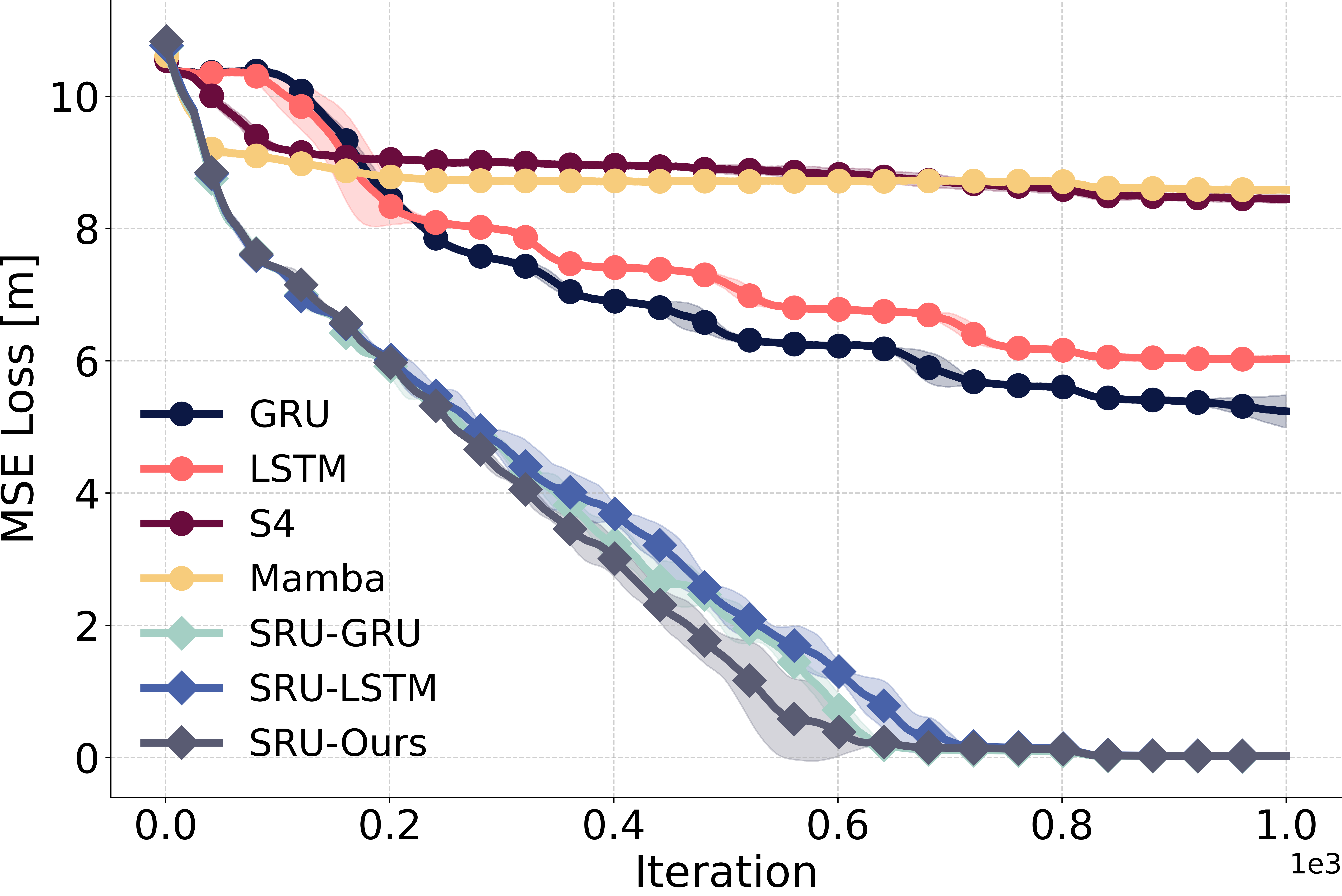}
        \label{fig:spatial_loss}%
        }
        \caption{Training for the Spatial-temporal Memorization: (a) Temporal memorization
        loss shows that standard RNN units (LSTM, GRU, S4, and Mamba-SSM)
        effectively recall sequential information. (b) Spatial memorization loss indicates
        that these units struggle with accurate spatial transformations and memorization
        under changing observation perspectives, resulting in misaligned landmark
        coordinates.}
        \label{fig:spatial_temporal_mapping}
      \end{center}
    \end{figure}

  In this section, we assess the spatial and temporal mapping performance of existing recurrent structures—namely LSTM, GRU, and the recent S4 and Mamba-SSM—on two fronts: (\romannumeral
  1) temporal memorization and (\romannumeral 2) spatial transformation and memorization.
  Consider an abstract scenario relevant to the navigation task, in which a robot is initialized at a pose in \(SE(3)\) and moves randomly within the three-dimensional environment \(E \subset \mathbb{R}^3\). At each time step $t$, the robot receives an observation $o_{t}$ containing the coordinates of observed landmarks $l^{i}_{t}\in \mathbb{R}^{3}$, defined relative to the robot's current frame (indicated by the subscript $t$). 
  Each landmark is also associated with a binary categorical label $c^{i}$, which is temporally relevant and independent of the observation frame. 
  Additionally, the robot is provided with its ego-motion transformation matrix $M^{t-1}_{t}$, representing the transformation from the previous pose at time step $t-1$ to the pose at time step $t$. 
  This transformation enables the network to align and integrate observations into a unified reference frame, akin to classical homogeneous transformations.
  Over a sequence of $T$ time steps, the RNN processes these observations. At the final step $T$, the network is evaluated simultaneously on its ability to:
  \begin{itemize}
    \item Memorize and accurately predict the sequence of binary categorical labels associated with the observed landmarks, ensuring temporal association and order preservation (Temporal Task).
    \item Transform and register the spatial coordinates of all observed landmarks into the final robot frame at $t=T$, achieving spatial alignment and memorization of their positions (Spatial Task).
  \end{itemize}
  The training details are provided in Appendix~\ref{appendix: spatial-temporal-details}.
  The results indicate that while LSTM, GRU, S4, and Mamba-SSM effectively encode temporal sequences and retain landmark categories, as shown in Figure~\ref{fig:temporal_loss}, they face significant challenges in accurately memorizing and transforming landmark coordinates from sequential ego-centric observations. 
  This limitation is reflected in the higher mean squared error (MSE) when recalling observed landmark positions during training, as depicted in Figure~\ref{fig:spatial_loss}.

  \begin{figure}
    \begin{center}
      \subfigure[Spatial Mapping with LSTM]{%
      \includegraphics[width=0.4\textwidth]{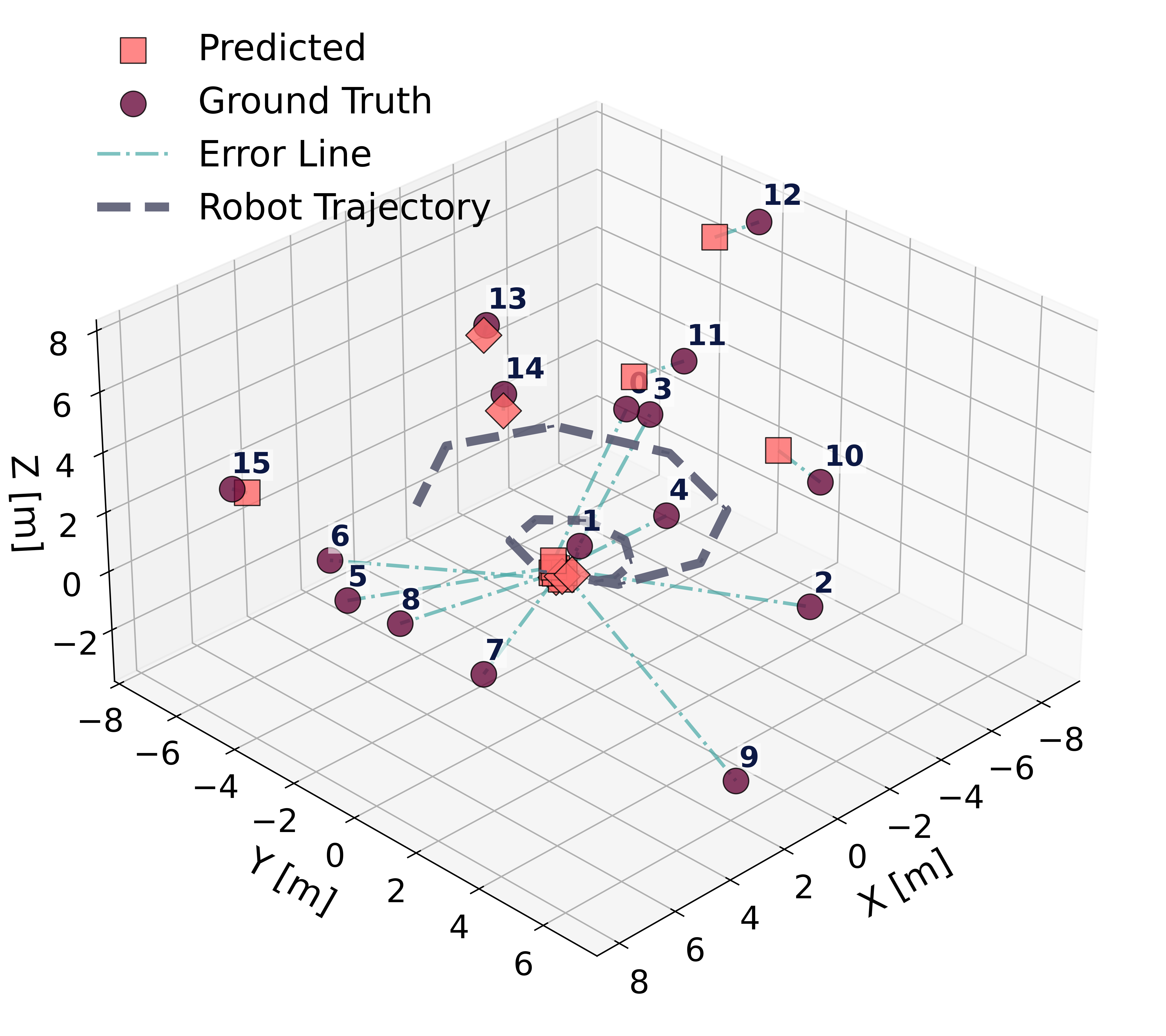}
      \label{fig:lstm_spatial_map}%
      } \subfigure[Spatial Mapping with SRU-LSTM]{%
      \includegraphics[width=0.4\textwidth]{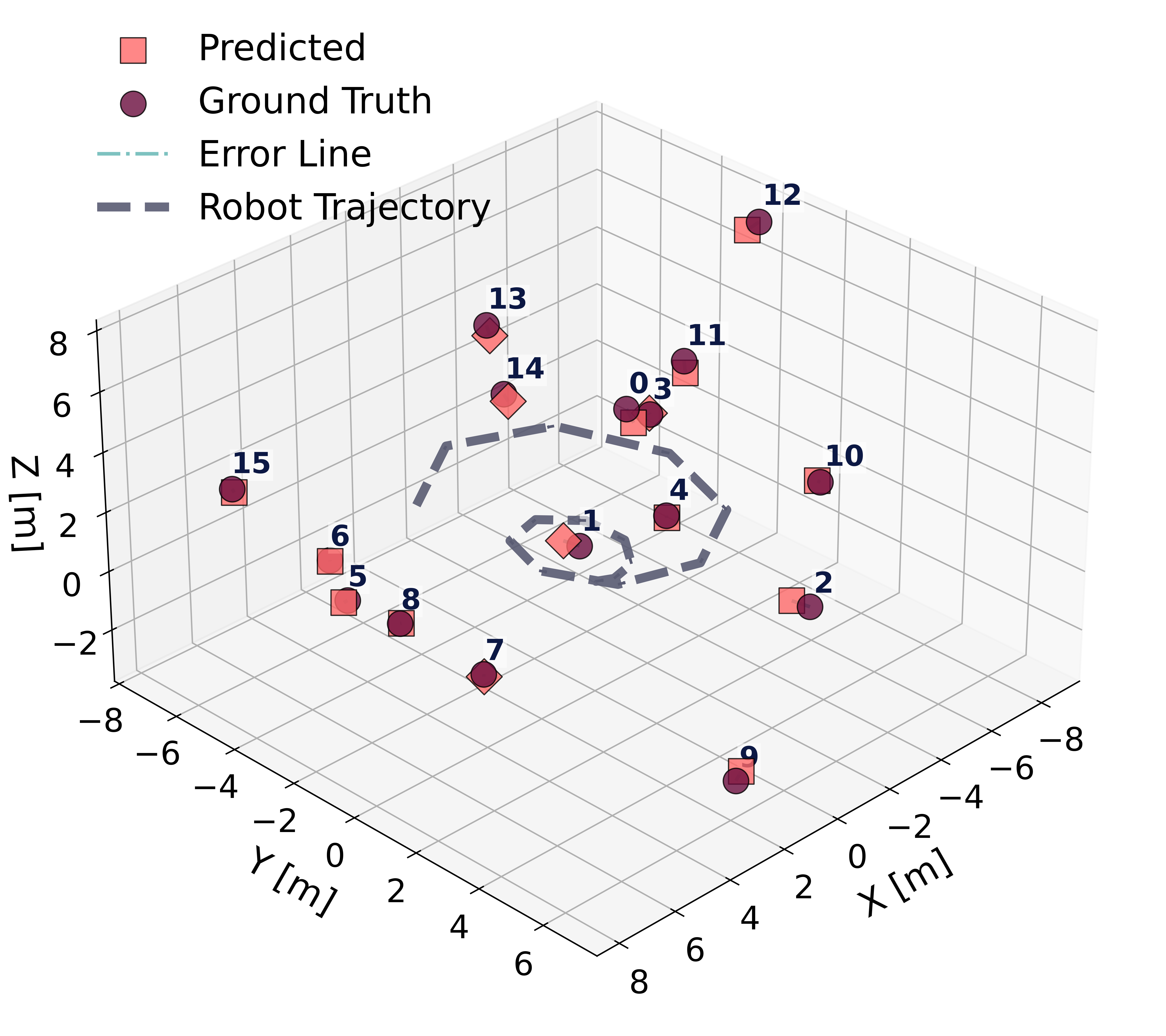}
      \label{fig:sru-lstm_spatial_map}%
      } \subfigure[Spatial Memory Error]{%
      \includegraphics[width=0.4\textwidth]{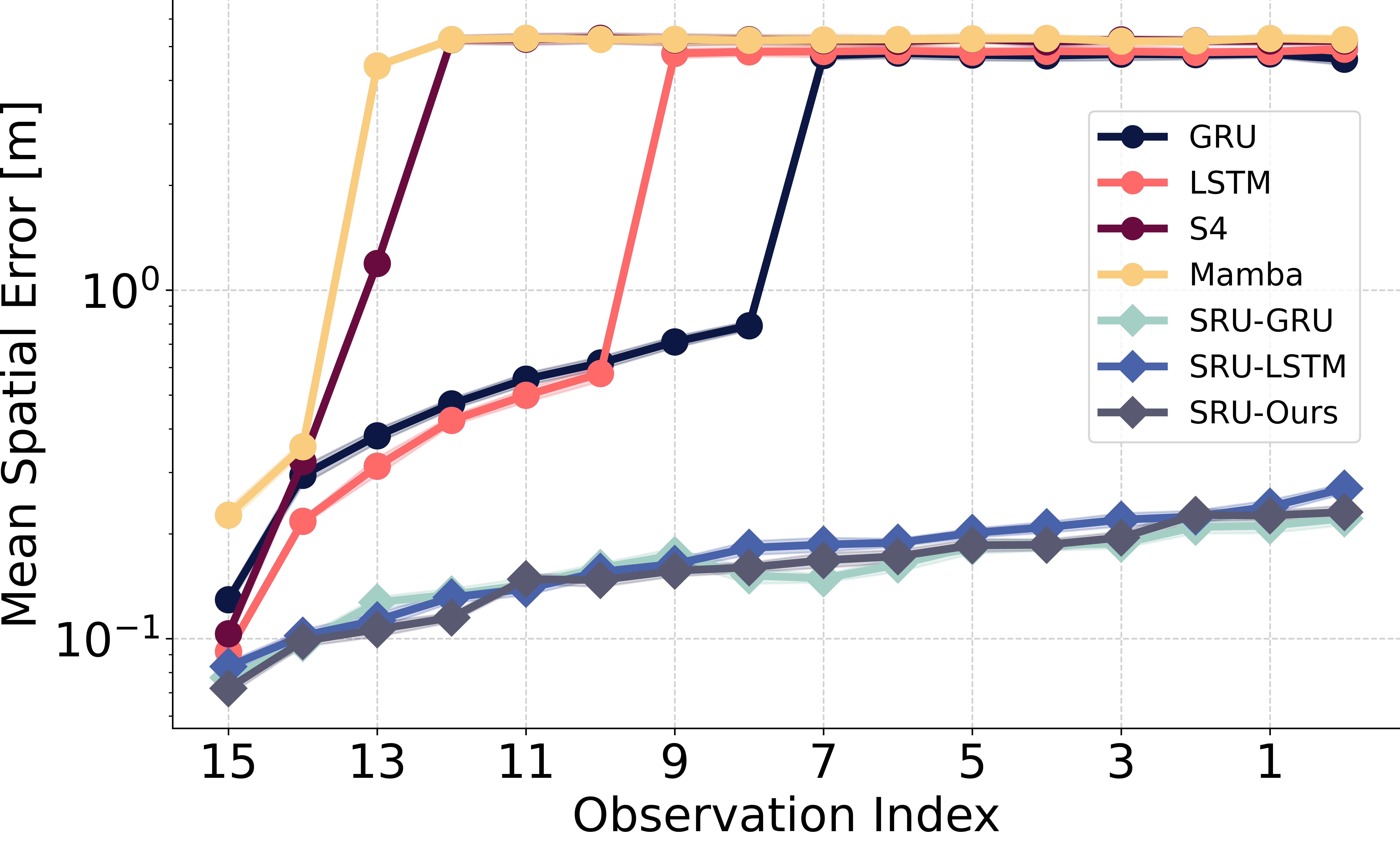}
      \label{fig:spatial_error_plot}%
      }
      \caption{Spatial Mapping Comparison: (a) and (b) depict the spatial mapping
      performance of LSTM and SRU-LSTM units on synthetic data, respectively, as
      the robot follows a spiral path, observing landmarks from different
      perspectives. At the end of the path, the robot is tasked with memorizing
      and transforming the observed landmark coordinates into the final robot
      frame. Numbers indicate observation time steps. (c) illustrates the mean spatial
      memory errors (log scale) across observation step indices, ordered from the final (15) to the
      initial step (1), averaged over various randomly generated trajectories and observations.}
      \label{fig:spatial_mapping_result}
    \end{center}
  \end{figure}


  \subsection{Spatially-Enhanced Recurrent Unit (SRU)}
  \label{sec:sru-sec} 
  To address the limitations of spatial mapping of existing RNN units, we propose a modification to the standard LSTM and GRU architectures by introducing an additional spatial transformation operation. This enhancement results in a new class of units, termed Spatially-Enhanced Recurrent Units (SRUs). The added operation enables the network to implicitly learn spatial transformations, aligning and memorizing observations from varying perspectives while preserving robust temporal memorization capabilities. 
  
  The effectiveness of this approach is demonstrated by the training results of the spatial mapping task mentioned above and illustrated in Figure~\ref{fig:spatial_temporal_mapping}. With the SRU modification, the network effectively transforms and memorizes observed landmark coordinates from different perspectives, as indicated by the spatial loss curve in Figure~\ref{fig:spatial_loss}, while preserving similar temporal memorization performance compared to standard LSTM and GRU units, as shown in Figure~\ref{fig:temporal_loss}. The design of SRUs emerged through iterative experimentation and analysis of spatial mapping performance. The final formulation draws inspiration from the multiplicative form of homogeneous transformations and recent research on the use of the "star operation" (element-wise multiplication) to enhance the representational capacity of neural networks~\citep{ma2024rewrite}.
  
  The following equations detail the modifications to incorporate spatial transformations into both LSTM and GRU units, ensuring a balance between spatial memorization and temporal dependency learning. In each case, we compute an additional spatial transformation term, denoted as $\mathtt{s}_{\mathtt{t}}$, which acts as a mechanism to implicitly transform and align the candidate state with the current observation's perspective. For the modified LSTM, referred to as SRU-LSTM, we define:
  \[
    \begin{aligned}
      \mathtt{s}_{\mathtt{t}} & = \mathtt{W}_{\mathtt{x}\mathtt{s}}\, \mathtt{x}_{\mathtt{t}}+ \mathtt{b}_{\mathtt{s}},                                                                                                                            \\
      \mathtt{g}_{\mathtt{t}} & = \tanh\Bigl( \mathtt{s}_{\mathtt{t}}\odot \bigl(\mathtt{W}_{\mathtt{x}\mathtt{g}}\, \mathtt{x}_{\mathtt{t}}+ \mathtt{W}_{\mathtt{h}\mathtt{g}}\, \mathtt{h}_{\mathtt{t}-1}+ \mathtt{b}_{\mathtt{g}}\bigr) \Bigr).
    \end{aligned}
  \]
  Similarly, for the modified GRU, referred to as SRU-GRU, the formulation is enhanced as follows:
  \[
    \begin{aligned}
      \mathtt{s}_{\mathtt{t}}         & = \mathtt{W}_{\mathtt{x}\mathtt{s}}\, \mathtt{x}_{\mathtt{t}}+ \mathtt{b}_{\mathtt{s}},                                                                                                                                                           \\
      \tilde{\mathtt{h}}_{\mathtt{t}} & = \tanh\Bigl( \mathtt{s}_{\mathtt{t}}\odot \bigl(\mathtt{W}_{\mathtt{x}\mathtt{h}}\, \mathtt{x}_{\mathtt{t}}+ \mathtt{W}_{\mathtt{h}\mathtt{h}}\,(\mathtt{r}_{\mathtt{t}}\odot \mathtt{h}_{\mathtt{t}-1}) + \mathtt{b}_{\mathtt{h}}\bigr) \Bigr).
    \end{aligned}
  \]
  To further enhance spatial-temporal memorization, we extend the SRU-LSTM with a refined gating mechanism~\citep{gu2020improving}, simply referred to as SRU-Ours in the following sections. 
  This mechanism introduces an additional refining function to address gating saturation issues during recurrent training. 
  The final modification, compared to the vanilla LSTM unit, is as follows:
  \[
    \begin{aligned}
      \mathtt{s}_{\mathtt{t}} & = \mathtt{W}_{\mathtt{x}\mathtt{s}}\, \mathtt{x}_{\mathtt{t}}+ \mathtt{b}_{\mathtt{s}},                                                                                                                            \\
      \mathtt{g}_{\mathtt{t}} & = \tanh\Bigl( \mathtt{s}_{\mathtt{t}}\odot \bigl(\mathtt{W}_{\mathtt{x}\mathtt{g}}\, \mathtt{x}_{\mathtt{t}}+ \mathtt{W}_{\mathtt{h}\mathtt{g}}\, \mathtt{h}_{\mathtt{t}-1}+ \mathtt{b}_{\mathtt{g}}\bigr) \Bigr), \\
      \mathtt{r}_{\mathtt{t}} & = \mathtt{i}_{\mathtt{t}}\odot \Bigl(1 - (1 - \mathtt{f}_{\mathtt{t}})^{2}\Bigr) + (1 - \mathtt{i}_{\mathtt{t}}) \odot \mathtt{f}_{\mathtt{t}}^{2},                                                                \\
      \mathtt{c}_{\mathtt{t}} & = \mathtt{r}_{\mathtt{t}}\odot \mathtt{c}_{\mathtt{t}-1}+ (1 - \mathtt{r}_{\mathtt{t}}) \odot \mathtt{g}_{\mathtt{t}}.
    \end{aligned}
  \]
    The effectiveness of SRU is further validated and compared to the standard LSTM unit in the designed spatial mapping task, as illustrated in Figure~\ref{fig:spatial_mapping_result}. In this task, the robot follows a spiral path, observing landmarks from varying perspectives along its trajectory. At the end of the path, the robot is tasked with memorizing and transforming the observed landmark coordinates into the final frame, as well as recalling the associated categories of the observed landmarks. Our experiments demonstrate that the SRU modification enables the network to effectively learn spatial transformations. In contrast, the baseline models struggle to align and memorize landmark coordinates observed in earlier steps, resulting in higher spatial errors, particularly for earlier observations, as depicted in Figure~\ref{fig:spatial_error_plot}. However, both SRU and baseline models achieve 100\% accuracy in recalling the categories of the observed landmarks. Since the temporal task results are identical across all models, they are not visualized Specifically in Figure~\ref{fig:spatial_mapping_result}. Furthermore, the latest recurrent units, such as S4 and Mamba-SSM, which excel at long-term temporal memorization, exhibit even worse spatial memorization capabilities, as shown in both Figure~\ref{fig:spatial_loss} and Figure~\ref{fig:spatial_error_plot}. The SRUs exhibit consistently low spatial errors across all observation steps, underscoring their superior spatial memorization capabilities.

  \begin{figure*}
    \begin{center}
      \includegraphics[width=0.99\textwidth]{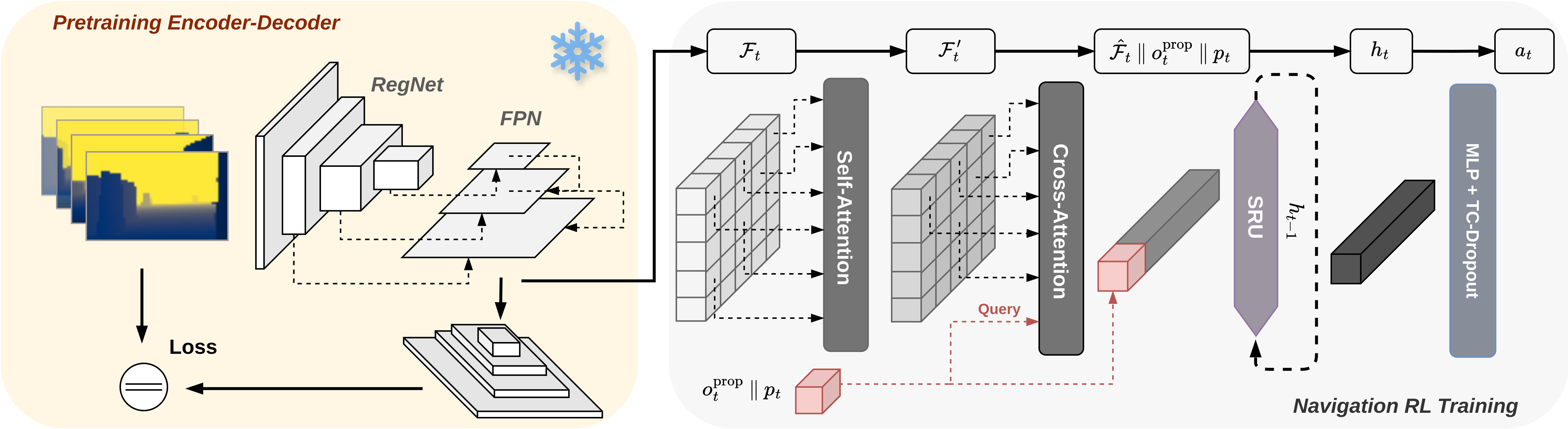}
      \caption{Attention-based recurrent network architecture for navigation: 
      The network integrates a pretrained image encoder and an attention mechanism to compress and emphasize relevant features from encoded observations. 
      These features, combined with proprioceptive inputs, are processed by the SRU unit, which learns spatial transformations and temporal dependencies and fuses them with historical observations to estimate the robot's state. 
      The state is then mapped to actions using an MLP-based head integrated with the temporally consistent (TC) dropout layer for improved robustness and generalization.}
      \label{fig:network_architecture}
    \end{center}
  \end{figure*}

  \subsection{Attention-Based Recurrent Network Architecture for Navigation}
  To leverage the SRUs in the navigation context, we propose an attention-based recurrent network architecture for long-range mapless navigation tasks using raw front-facing stereo depth input, as illustrated in Figure~\ref{fig:network_architecture}. The network consists of a pretrained depth image encoder, two spatial attention layers incorporating both self-attention and cross-attention mechanisms to enhance and compress the encoded visual features, and a recurrent unit (SRU) that learns a spatial-temporal representation of the state by fusing the current observation with historical observations. Finally, a multilayer perceptron (MLP) head computes the actions from the recurrent hidden state, outputting velocity commands for the robot's locomotion controller.


  \begin{figure}
    \begin{center}
      \subfigure[Original Synthetic Image]{%
      \includegraphics[width=0.22\textwidth]{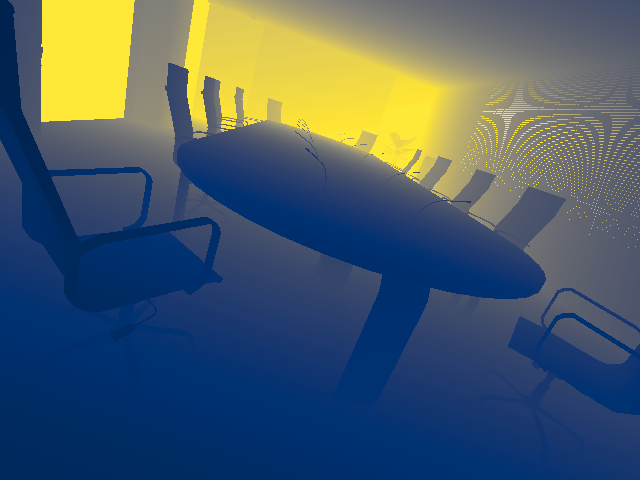}%
      \label{fig:synthetic} } \subfigure[Image with Artificial Noise]{%
      \includegraphics[width=0.22\textwidth]{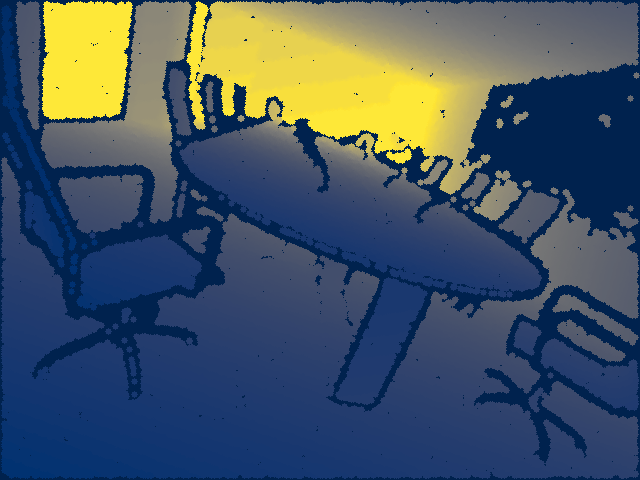}%
      \label{fig:noise} }
      \caption{Simulated stereo depth noise: (a) Synthetic depth image from the TartanAir dataset~\citep{wang2020tartanair}, (b) Image with augmented artificial noise. 
      The depth-noise model introduces edge, filling, and rounding noise to
      the depth images, simulating realistic sensor artifacts.}
      \label{fig:depth_noise}
    \end{center}
  \end{figure}

  \subsubsection{Depth Encoder Pretraining and Simulated Perception Noise}
  For the depth encoder, we adopt a convolutional neural network (CNN) backbone based on RegNet~\citep{radosavovic2020designing}, chosen for its simplicity and efficiency. This is further enhanced with a Feature Pyramid Network (FPN)~\citep{lin2017feature} to capture spatial features across multiple scales. The encoder is pretrained for self-reconstruction on large-scale synthetic depth image data from TartanAir~\citep{wang2020tartanair} using a variational autoencoder (VAE) framework. This pretraining enables the encoder to learn and extract robust and generalizable features from depth images, facilitating effective downstream navigation learning and deployment. However, depth images captured in simulation often differ from those obtained in real-world environments due to various sensor artifacts and noise. To address this sim-to-real gap, we integrate a parallelized depth-noise model, adapted from~\citep{handa2014benchmark, barron2013intrinsic, bohg2014robot}, which introduces configurable noise to the depth images, such as:

  \begin{itemize}
    \item Edge Noise: Distortions at sharp depth discontinuities due to abrupt changes in the scene.
    
    \item Filling Noise: Blurring artifacts introduced during the interpolation of missing or unregistered pixels.
      
    \item Rounding Noise: Quantization effects resulting from sensor resolution limitations, causing rounding errors.
  \end{itemize}
  Figure~\ref{fig:depth_noise} illustrates an example of the simulated stereo depth noise. 
  The depth-noise model is designed for efficient batch processing, enabling parallelized pretraining on large-scale datasets and during RL with simulated depth images. 
  For implementation details, refer to Appendix~\ref{appendix:
  stereo-depth-noise-pseudocode}.

  \begin{figure*}
    \begin{center}
      \subfigure[Robot State | Turning Left]{%
      \includegraphics[width=0.32\textwidth]{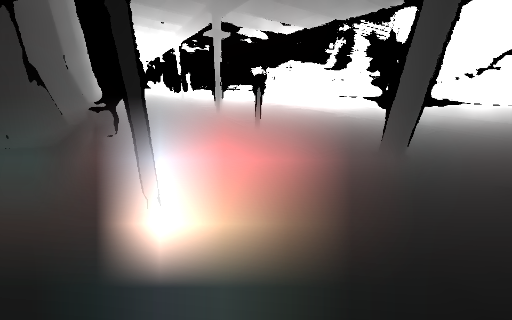}%
      \label{fig:attention_turn_left}%
      } \subfigure[Robot State | Going Straight]{%
      \includegraphics[width=0.32\textwidth]{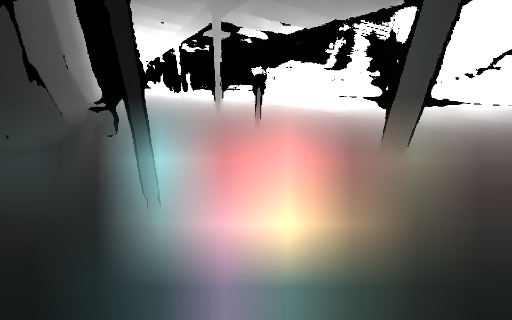}%
      \label{fig:attention_go_straight}%
      } \subfigure[Robot State | Turning Right]{%
      \includegraphics[width=0.32\textwidth]{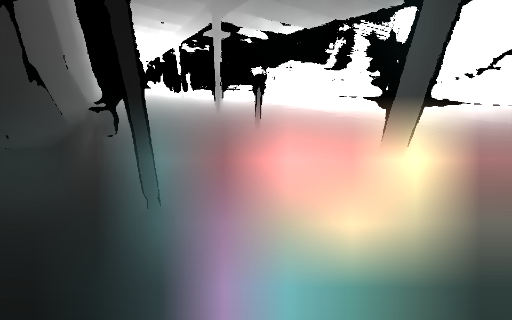}%
      \label{fig:attention_turn_right}%
      }
      \caption{Visualization of cross-attention weights corresponding to different robot states over raw real-world depth input: 
      (a) When the robot turns left, the attention weights highlight the left region, focusing on the left pillar in the depth image; 
      (b) When the robot moves straight, the attention weights emphasize the central region, capturing both pillars; 
      (c) When the robot turns right, the attention weights focus on the right region, concentrating on the right pillar.
      Distinct colors in the attention weights visualizations represent different attention heads.}
      \label{fig:attention_mask_turning}
    \end{center}
  \end{figure*}

  \subsubsection{Attention Layers for Feature Compression}
  During navigation, humans and animals tend to focus on the most relevant spatial cues rather than attempting to memorize all available information~\citep{matthis2018gaze}. This selective attention enables more efficient and effective memory usage. To emulate this, we combine self-attention and cross-attention mechanisms in our architecture. These spatial attention layers process high-dimensional visual inputs, extracting the information most relevant to the robot’s current state. 
  Specifically, given the feature map encoded by the pretrained depth encoder:
  \[
    \mathcal{F}_{t}\in \mathbb{R}^{C \times H \times W},
  \]
  each spatial feature $F^{ij}_{t}$ (where $i \in \{1,\dots,H\}$ and $j \in \{1,\dots ,W\}$) is enriched with global context via a self-attention mechanism, resulting in a refined feature map $\mathcal{F}'_{t}$. Here, $H$ and $W$ denote the height and width of the feature map, respectively, while $C$ represents the number of channels. The self-attention mechanism computes the attention weights across the spatial dimensions of the feature map, enabling the network to fuse and emphasize relevant features while suppressing less important ones. Next, $\mathcal{F}'_{t}$ is processed by a cross-attention layer, where the query is derived from the robot’s egocentric proprioceptive state $o_{t}^{\text{prop}}$---which includes linear velocity $v_{t}$, angular velocity $\omega_{t}$, projected gravity $n_{t}$, and the previous action $a_{t-1}$---as well as the current relative goal position $p_{t}$. 
  This procedure is illustrated in Figure~\ref{fig:network_architecture}.
  This operation compresses the 2-dimensional feature map into a 1-dimensional latent representation
  \[
    \hat{\mathcal{F}}_{t}\in \mathbb{R}^{C \times 1},
  \]  
  which preserves the most relevant spatial features while reducing the dimensionality. This compressed feature $\hat{\mathcal{F}}_{t}$ is then concatenated with the proprioceptive state $o_{t}^{\text{prop}}$ and the relative goal position $p_{t}$ before being passed to the recurrent memory unit. There, it is fused with historical observations to form a spatial-temporal representation of the robot’s current state, integrating both exteroceptive and proprioceptive information.

  Figure~\ref{fig:attention_mask_turning} visualizes the output attention weights of the cross-attention layer across four distinct attention heads (depicted in different colors), overlaid on the depth input. It highlights how the attention mechanism focuses on depth features relevant to the robot's current state. 
  When the robot's movement direction is manually altered, the output attention weights shift accordingly, emphasizing spatial regions and obstacles in the new direction. These behaviors emerge naturally during end-to-end learning, demonstrating the policy's ability to effectively acquire critical spatial cues for navigation.

  \subsubsection{Spatially-Enhanced Recurrent Unit (SRU)}
  The Spatially-Enhanced Recurrent Unit (SRU), as described in Sec.~\ref{sec:sru-sec}, processes the compressed feature map $\hat{\mathcal{F}}_{t}$, along with the robot's current proprioceptive state $o_{t}^{\text{prop}}$ and the previous hidden state $h_{t-1}$, to generate a spatial-temporal representation of the surrounding environment from the robot's egocentric observations. 
  The observation of the current proprioceptive state $o_{t}^{\text{prop}}$ provides essential ego-motion information, equivalent to the transformation matrix $M^{t-1}_{t}$ in Sec.~\ref{sec:spatial-mapping-limitation-sec}.
  The SRU learns to implicitly perform spatial transformations, aligning the current observation feature map $\hat{\mathcal{F}}_{t}$ with the previous hidden state $h_{t-1}$. 
  The resulting hidden state $h_{t}$, which encapsulates the integrated environmental information to estimate the robot's current state $s_{t}$, is subsequently passed through a multi-layer perceptron (MLP) head to compute the robot's action $a_{t}$.

  \subsection{Learning Navigation with Sparse Rewards and Regularizations}
  \label{sec:reward-regularization} 
  The final attention-based network with the SRU is trained end-to-end using RL to achieve long-range mapless navigation, with the objective of maximizing the cumulative reward over the episode. 
  The reward function for the navigation task is designed as a combination of task-level rewards $r^{\text{task}}$, regularization $r^{\text{reg}}$, and penalty $r^{\text{pen}}$ terms, as follows:
  \[
    r_{t}= \alpha_{1}r^{\text{task}}_{t}- \alpha_{2}r^{\text{reg}}_{t}- \alpha_{3}
    r^{\text{pen}}_{t}.
  \]
  Here, $\alpha_{1}$, $\alpha_{2}$, and $\alpha_{3}$ are coefficients used to balance the contributions of the task-level reward, regularization, and penalty terms, respectively. The task-level reward $r^{\text{task}}_{t}$ is the reward signal that encourages the agent to reach the goal. 
  To promote exploration in complex environments, we adopt time-based rewards, similar to \cite{rudin2022advanced, zhang2024resilient, He-RSS-24}, that provide feedback at the end of the episode.
  This approach provides a sparse reward signal, encouraging the agent to reach the goal without being distracted by intermediate rewards. However, with a long episode length $T_{\text{max}}$, e.g., 60 seconds, and a rewarding period $T_{r}$ of only 2 seconds, the network may learn to delay progress until the final step. 
  To mitigate this, we introduce a random check during the episode with a small probability $\delta_{\text{check}}$. This check incentivizes the agent to attempt reaching the goal earlier, without compromising the overall sparsity of the reward signal. The final reward formulation is adapted from \cite{He-RSS-24} and is given as follows:

  \[
    r^{\text{task}}_{t}= \frac{\mathbf{1}(t > T_{\text{max}}- T_{r} \; \lor \;
    \text{random}< \delta_{\text{check}})}{1 + \|\frac{p_{t}}{\sigma}\|_{2}}
  \]
  where $\mathbf{1}(\cdot)$ is the binary indicator function, $T_{r}$ is the rewarding period, $\lor$ represents the logical OR operation, $p_{t}$ is the relative goal position at time $t$ with respect to the current robot pose, and $\sigma$ is a scaling factor controlling the reward's spatial sensitivity. Similar to \cite{He-RSS-24}, we adopt two reward configurations: a tight reward with a small $\sigma$ to encourage
  precise goal-reaching behavior, and a loose reward with a larger $\sigma$ to promote exploration and stabilize training through intermediate guidance. 
  
  The regularization term $r^{\text{reg}}_{t}$ encourages smooth behaviors by penalizing rapid action changes and excessive joint accelerations. 
  This is implemented using L1 regularization on the difference between the current action $a_t$ and a momentum-filtered version of previous actions $a_t^{m}$, as defined below:
  \[
    a_{t}^{m}= \lambda \cdot a_{t-1}^{m}+ (1 - \lambda) \cdot a_{t},
  \]
  where $\lambda$ is the momentum factor. 
  The regularization reward is then given by:
  \[
    \begin{aligned}
      r^{\text{reg}}_{t} & = \beta_{1}\cdot \|{a_{t} - a_{t}^{m}}\|_{1}+ \beta_{2}\cdot \|j_{t}^{\text{acc}}\|_{1},
    \end{aligned}
  \]
  where $\beta_{1}$ and $\beta_{2}$ are regularization coefficients, and $j_{t}^{\text{acc}}$ represents joint-level accelerations from the simulation environment. 
  The penalty term $r^{\text{pen}}_{t}$ discourages unsafe behaviors such as collisions or excessive tilt:
  \[
    r^{\text{pen}}_{t}= \eta_{1}\cdot \mathbf{1}(\text{collision}) + \eta_{2}
    \cdot \max(0, |\theta_{t}| - \theta_{\text{safe}})
  \]
  where $\eta_{1}$ and $\eta_{2}$ are penalty coefficients, $\theta_{t}$ is the robot's current tilt angle, and $\theta_{\text{safe}}$ defines the safe tilt threshold. The reward formulation described above is consistently applied throughout the entire RL training process, which is conducted end-to-end using an Asymmetric Actor-Critic~\citep{pinto2017asymmetric} setup and trained with PPO, without employing any additional environment or reward curricula. Further training details are provided in Appendix~\ref{appendix: rl-training-details}.

  \subsubsection{Training Regularization}
  To mitigate overfitting and enhance robustness, we incorporate two additional regularization strategies during training. 
  These strategies are crucial for training a robust spatial-temporal representation with SRUs, as explained below and demonstrated in the experimental results. The regularization techniques are as follows:
  
  \begin{itemize}
    \item Deep Mutual Learning (DML): As described in \cite{xie2025meta}, DML involves training two policies simultaneously, enabling them to mutually distill knowledge from each other. This approach enhances generalization and mitigates the risk of convergence to suboptimal solutions. The mutual distillation is achieved by incorporating a Kullback–Leibler (KL) divergence loss between the two policies, both of which are trained using standard proximal policy optimization (PPO)~\citep{schulman2017proximal}.
    
     \item Temporally Consistent Dropout (TC-Dropout): Adapting from \cite{hausknecht2022consistent}, we apply a consistent dropout mask across time steps during both rollout and training, ensuring stable memory learning within the recurrent structure.
  \end{itemize}

  As shown in Figure~\ref{fig:spatial_temporal_mapping}, SRUs exhibit a slower convergence rate for spatial memorization compared to learning temporal dependencies, highlighting the inherent complexity and slower pace of learning spatial transformations and forming spatial memory. This discrepancy can lead the network to favor easier-to-learn solutions early in training, relying on temporal features while neglecting the formation of good spatial memorization, resulting in suboptimal performance. 
  To address this, it is crucial to incorporate regularization techniques that mitigate early overfitting and promote the exploration and development of more challenging spatial-temporal features during policy optimization.
  To tackle this challenge, we employ deep mutual learning (DML) strategies tailored for reinforcement learning (RL)~\citep{xie2025meta}. DML involves training multiple policies in parallel, allowing them to distill knowledge from one another. 
  This mutual distillation process enhances the network's generalization capabilities and fosters the learning of robust and essential features. 
  By regularizing each other, the models are less likely to converge prematurely to suboptimal solutions that rely solely on easy-to-learn features, such as temporal dependencies. 
  Instead, DML encourages the formation of spatial-temporal representations, leading to improved overall performance. 
  This approach is critical for leveraging the full potential of the SRU network, as demonstrated in the experimental results.

  Secondly, compared to standard dropout layers, consistent dropout addresses a critical issue in on-policy reinforcement learning, where standard dropout introduces inconsistent masks between the rollout and training stages~\citep{hausknecht2022consistent}. 
  Building on this, we extend consistent dropout with temporal consistency for training the recurrent structure. 
  Specifically, during the data rollout stage, we maintain the same dropout mask across all time steps, ensuring temporal consistency.
  During the training stage, the same dropout mask is applied to the policy network.
  This approach promotes stable memory learning through recurrent connections and enhances the robustness of the learned policy.

  \begin{figure*}
    \begin{center}
      \includegraphics[width=0.99\textwidth]{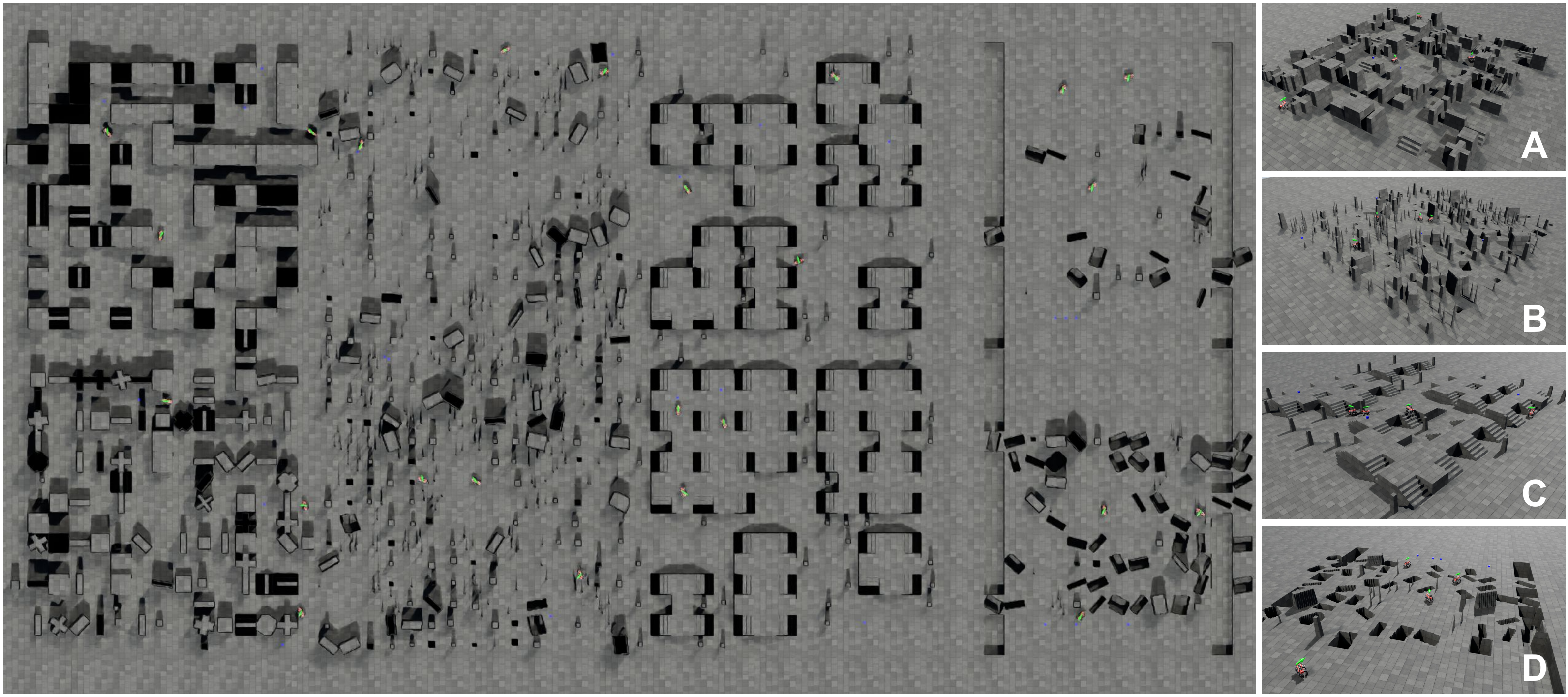}
        \caption{Simulated environments used for training and testing RL-based navigation tasks: (A) Maze, (B) Random Pillars, (C) Stairs, and (D) Pits. These environments are parameterizable and can be randomly generated during both training and testing using the \textit{NVIDIA IsaacLab}~\citep{mittal2023orbit} simulation framework.} 
        \label{fig:sim_enviroment}
    \end{center}
  \end{figure*}

  \section{Experiments}
  To evaluate the effectiveness of the proposed Spatially-Enhanced Recurrent Unit (SRU) and the attention-based network architecture in enhancing long-horizon robot navigation, we conduct experiments in both simulated and real-world environments. 
  We compare the SRU against standard LSTM and GRU units in long-range mapless
  navigation tasks, focusing on their performance in end-to-end reinforcement learning (RL) training and navigation success rates (SR). Additionally, we compare the SRU policy, integrated with our proposed network structure and trained using recurrent RL, against two current state-of-the-art (SOTA) baselines~\citep{huang2023goal, lee2024learning} for robot navigation with RL. 
  Our evaluation highlights the advantages of the implicit recurrent memory provided by SRU in solving long-range mapless navigation tasks across diverse environments.

  Furthermore, we ablate the role of our proposed spatial attention layers in compressing features from encoded observations to improve memorization and overall navigation performance. We compare our approach against the convolution and average pooling method used in \cite{wijmans2019dd}, as well as the attention mechanism introduced in \cite{huang2023goal}. We also investigate the impact of regularization techniques on training the SRU unit end-to-end in RL, evaluating their effectiveness in preventing early convergence to suboptimal solutions and enhancing navigation performance. 
  Finally, we explain and validate the pretrained image encoder’s ability to bridge the sim-to-real gap by demonstrating zero-shot transfer across diverse and complex real-world environments.


  \subsection{Experimental Setup}
  We conduct our experiments in simulated 3D environments using \textit{NVIDIA IsaacLab}~\citep{mittal2023orbit}, which provides a realistic physics engine and fast, parallelizable simulation capabilities. The environments are designed to challenge the robot's navigation capabilities and include maze-like structures, randomly generated pillars, stairs, and environments with negative obstacles, such as holes and pits, as shown in Figure~\ref{fig:sim_enviroment}.
  The robot is equipped with front-facing depth sensors as the only exteroceptive input, capturing the surrounding environment from an egocentric perspective.
  Additionally, a state estimation and localization module provides the robot's proprioceptive state $o_{t}^{\text{prop}}$, including linear and angular velocities ($v_{t}$ and $\omega_{t}$), projected gravity ($n_{t}$), and the relative goal position ($p_{t}$) with respect to the robot's frame at time step $t$. 
  Given the limited field of view (FoV) of the depth camera (Horizontal FoV: $105^{\circ}$, Vertical FoV: $78^{\circ}$) and a maximum range of 10 meters, the robot must rely on its spatial-temporal mapping capabilities to navigate through the terrain and reach the designated goal region effectively.
  The robot's motion is controlled by a set of linear and angular velocities, referred to as the action $a_{t}$, which is the output of the policy network. 
  The navigation policy operates at a frequency of 5 Hz. 
  The robot is equipped with a learning-based locomotion controller~\citep{lee2024learning}, operating at 50hz. This controller takes the action output $a_{t}$ from the high-level navigation policy and executes it to control the robot. 
  The policies are trained end-to-end using reinforcement learning, without employing any distillation or teacher-student setups.

  \begin{figure*}
    \begin{center}
      \subfigure[Navigation Policy with LSTM Unit]{%
      \includegraphics[width=0.99\textwidth]{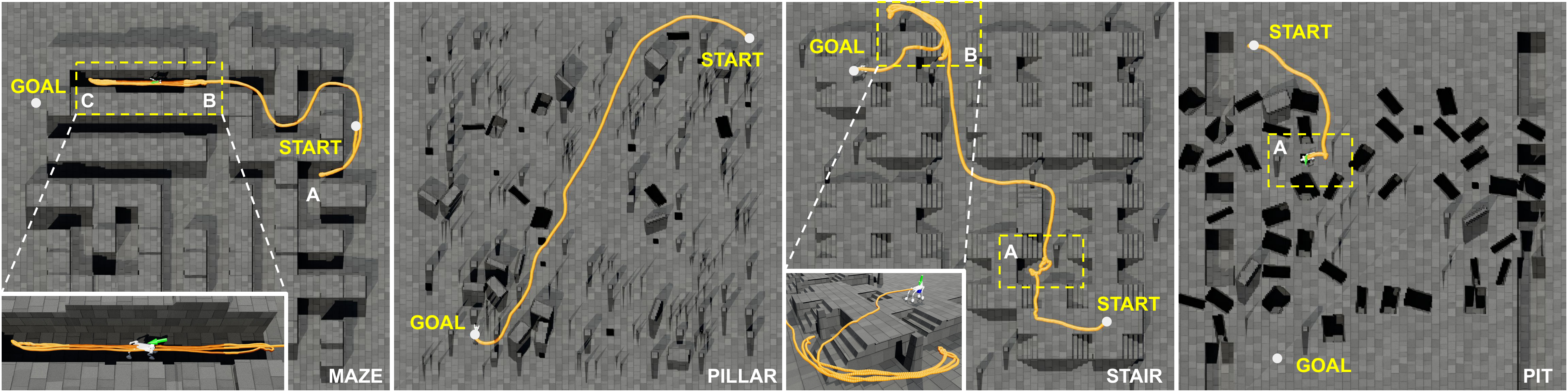}
      \label{fig:rnn_comparison_lstm}%
      } \hspace{0.05\textwidth} \subfigure[Navigation Policy with SRU Unit]{%
      \includegraphics[width=0.99\textwidth]{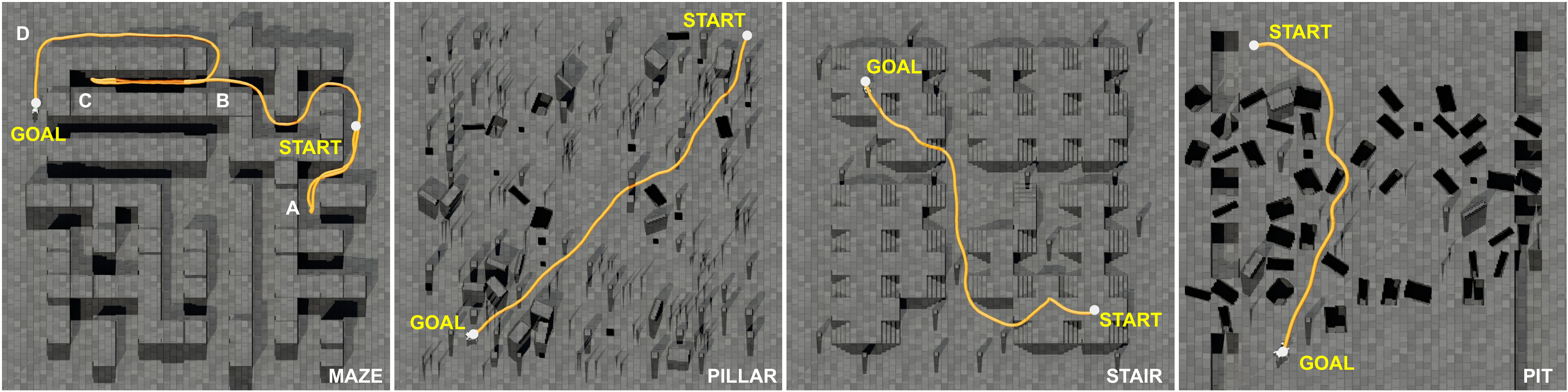}%
      \label{fig:rnn_comparison_sru}%
      }
      \caption{Comparison of navigation trajectories using (a) Navigation Policy with LSTM Unit and (b) Navigation Policy with SRU-Ours. The traversed trajectories are shown in yellow. In maze environments, the LSTM policy becomes trapped in a dead-end corridor, repeatedly looping between points B and C, while the SRU policy successfully navigates through the corridor, traverses region D, and reaches the goal. In stair-like environments, the LSTM policy exhibits frequent back-and-forth movements in areas A and B, indicating unreliable spatial-temporal mapping. In pit environments, the LSTM policy fails to avoid previously encountered pits during turns at area A, whereas the SRU policy effectively recalls their locations and avoids them, even during backward motion.}
      \label{fig:rnn_comparasion}
    \end{center}
  \end{figure*}

  \begin{figure}
    \begin{center}
      \includegraphics[width=0.45\textwidth]{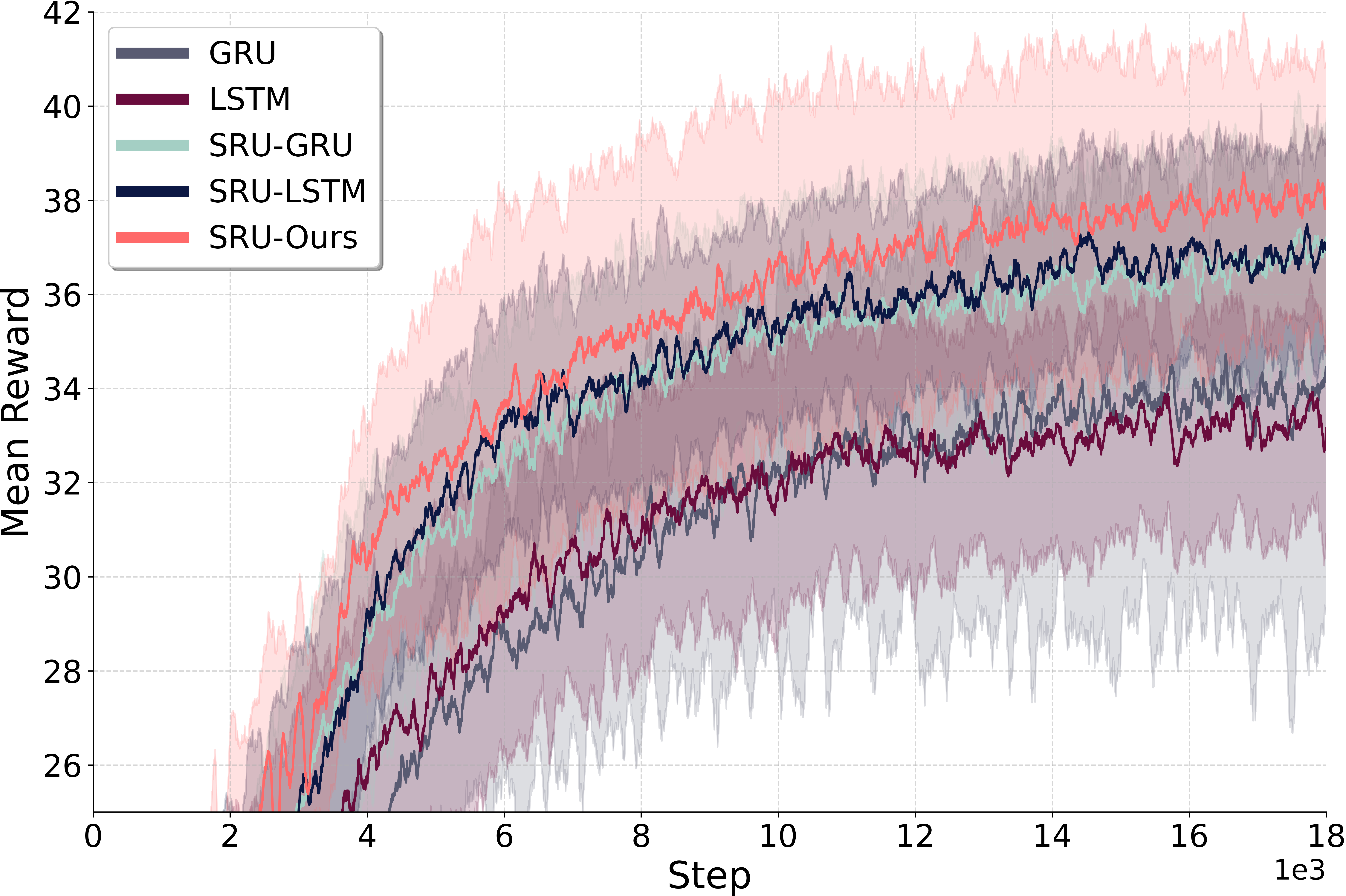} 
      \caption{Training curve comparison between policies integrated with
      different recurrent units: The average return from three random seeds during training. The architecture with SRU units achieves a higher return compared to the baseline LSTM and GRU units.}
      \label{fig:training_curve}
    \end{center}
  \end{figure}

  \begin{table}
    \centering
    \begin{tabular}{l|cccc|c}
      \toprule \multicolumn{6}{c}{Navigation Success Rate - SR \%}  \\
      \midrule \textbf{Model}                                     & \textbf{Maze} & \textbf{Pillar} & \textbf{Stair} & \textbf{Pit}  & \textbf{\textit{Overall}} \\
      \midrule GRU                                                 & 68.1          & 73.6            & 35.7           & 66.7          & 61.0             \\
      LSTM                                                         & 70.3          & 78.2            & 33.1           & 72.7          & 63.5             \\
      \addlinespace[0.5em] \hdashline \addlinespace[0.5em] SRU-GRU & 73.1          & 78.8            & 74.1           & 74.8          & 75.2             \\
      SRU-LSTM                                                     & 75.9          & 76.7            & 79.3           & 74.1          & 76.5             \\
      SRU-Ours                                                     & \textbf{76.0} & \textbf{81.0}   & \textbf{82.8}  & \textbf{75.6} & \textbf{78.9}    \\
      \bottomrule
    \end{tabular}
    \vspace{0.5em}
    \caption{Navigation success rate (SR) for policies integrated with different recurrent units across four environment types: Maze, Random Pillars, Stairs, and Pits. The table compares standard LSTM and GRU units with our proposed spatially enhanced counterparts: SRU-GRU, SRU-LSTM, and SRU-Ours.}
    \label{tab:nav_performance}
  \end{table}

  \subsection{Comparsion with Recurrent Units}

  We evaluate the performance of the proposed Spatially-Enhanced Recurrent Units (SRUs) compared to standard LSTM and GRU units. 
  Given the superior spatial memorization capability of SRUs, as demonstrated in Figure~\ref{fig:spatial_temporal_mapping}, we hypothesize that integrating SRUs will improve the performance of navigation policies in addressing long-range mapless navigation tasks. To test this hypothesis, we train, under same conditions, policy networks integrated with different recurrent units end-to-end using RL in the simulated 3D environments shown in Figure~\ref{fig:sim_enviroment}. All policies are equipped with the same components (attention, training regularization, and a pretrained encoder); only the recurrent network structure differs. We then evaluate their navigation performance. 

  As shown in Figure~\ref{fig:training_curve}, the policy with the SRU memory unit is able to outperform those with standard LSTM and GRU units in terms of average return episode rewards during training, with results averaged across multiple random seeds.
  (Note: GRU training can exhibit instability, so only its successful runs are included in the analysis.), Table~\ref{tab:nav_performance} provides a summary of the navigation performance for policies using different architectures. 
  The best-performing model from each unit (determined by the highest average return rewards) is selected for comparison. 
  The data is averaged over 4800 episodes across 120 randomly generated environments, which are different from the training set. 
  The results are presented in terms of success rate (SR) for each environment. The SRU units consistently outperform the standard LSTM and GRU units, achieving an average 21.8\% improvement in SR across all environments with the SRU modification alone. Furthermore, incorporating the refined gating mechanism in the SRU-Ours model further boosts the results, achieving an overall 23.5\% increase in SR, demonstrating the effectiveness of these enhancements in improving navigation performance. Notably, in stair-like environments, where the 3D structure and significant occlusions pose challenges for navigation without precise spatial memorization and registration capabilities, the navigation policy with SRU units demonstrate over double the performance in success rate compared to standard LSTM and GRU units.

  Figure~\ref{fig:rnn_comparasion} presents example traversed trajectories comparison between the SRU and standard LSTM policies. 
  In maze environments, the LSTM policy gets trapped in a dead-end corridor, looping between points B and C, while the SRU policy successfully passes through the corridor, demonstrating better spatial memorization capability. 
  In stair-like environments, although the LSTM eventually reaches the destination, it exhibits frequent back-and-forth movements (areas A and B), indicating less reliable spatial-temporal memorization and estimation of the current state compared to the policy with SRU. 
  In pit environments, the LSTM policy fails to avoid the previously encountered pits that are no longer visible in the current depth observation, when turning at area A.
  In contrast, the SRU policy effectively recalls the locations of the pit and other previously observed obstacles, enabling it to avoid them during turns and backward motion.

  \subsection{Comparsion against RL-based Navigation Baselines}
  \label{sec:compare-baseline}
  Next, we compare our proposed network structure, trained using recurrent reinforcement learning with SRU, against two state-of-the-art RL baseline methods: the Goal-guided Transformer-based RL approach (GTRL)~\citep{huang2023goal} and the RL approach with Explicit Mapping and Historical Path (EMHP)~\citep{lee2024learning}. GTRL employs a goal-guided Transformer (GoT) architecture to extract task-relevant visual features from stacked historical observations, enabling mapless navigation using only egocentric input. EMHP employs an external mapping pipeline to integrate historical observations for local mapping~\citep{miki2022elevation} and uses an explicit historical traversed path to address POMDP challenges in long-range mapless navigation. While explicit mapping (EMHP) can theoretically achieve high accuracy in spatial-temporal registration, it has two major drawbacks: (i) it introduces significant delays that hinder real-time performance, especially on high-speed, agile platforms~\citep{lee2024learning}, and (ii) it relies on heuristic rules (e.g., fixed context window lengths) to select information, limiting its ability to capture complex spatial-temporal dependencies and abstract information beyond the selected context window.

\begin{figure}
    \centering
    \subfigure[Policy with Explicit Mapping and Historical Path (EMHP)]{%
    \includegraphics[width=0.4\textwidth]{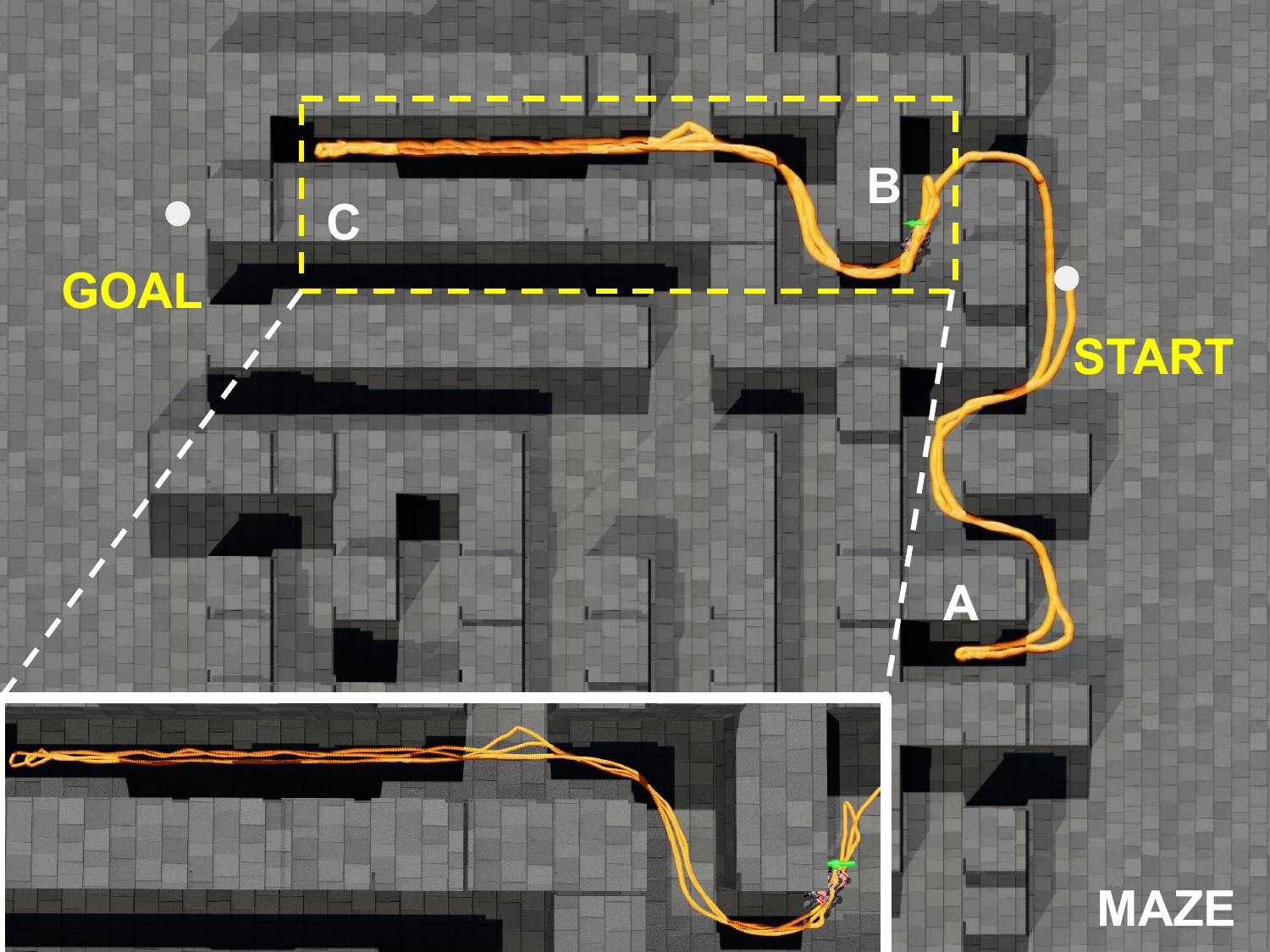} \label{fig:baseline_maze_compare}%
    } \subfigure[Policy with SRU Recurrent Memory]{%
    \includegraphics[width=0.4\textwidth]{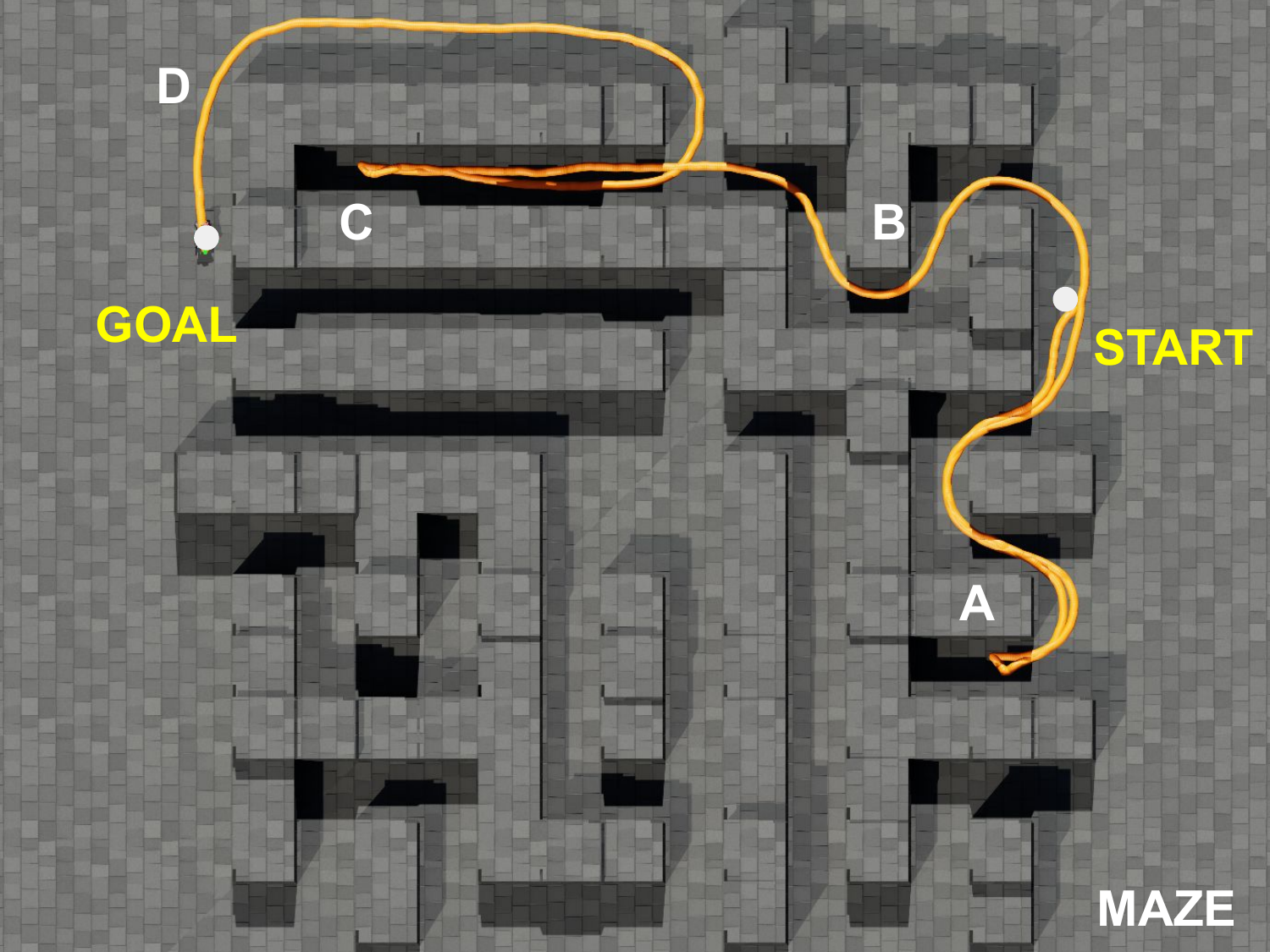}%
    \label{fig:ours_maze_compare}%
    }
    \caption{
    Comparison of proposed mapless method with SRU recurrent memory against the EMHP baseline approach in a maze environment. The robot's traversed trajectory is shown in yellow, with traversal order marked as A, B, C, and D. 
    (a) The EMHP approach starts looping in the long corridor between points B and C, failing to navigate through the dead-end corridors.
    (b) Our approach, with SRU recurrent memory, successfully navigates from start to goal, rerouting through the dead-end corridors and reaching the goal through area D.}
    \label{fig:maze_comparison}
  \end{figure}

  Figure~\ref{fig:baseline_maze_compare} presents a comparison between the EMHP baseline and our SRU-based approach. The EMHP policy collects historical paths for approximately 20 meters, which is insufficient to navigate the long corridor spanning around 30 meters. In contrast, recurrent neural networks offer an unlimited context window and can learn intricate spatial-temporal dependencies optimized for the given task. This allows the SRU-based policy to adapt to long-horizon navigation challenges more effectively. Moreover, our end-to-end architecture processes raw depth sensor inputs, reducing latency and better supporting the agile and fast motion of legged-wheel platforms during deployment. For the GTRL baseline, temporal history is provided by stacking several past observation frames, which are then fused using the transformer-based architecture as described in~\citep{huang2023goal}. Following the original approach in \cite{huang2023goal}, we use the 4-frame history in our experiments. However, the choice of the number of stacked frames remains heuristic: a short history may miss important context, while a longer history increases computational cost quadratically. For a fair comparison, we retrain the GTRL baseline, as described in~\cite{huang2023goal}, within our environment. We replace the RGB input with depth images and utilize the same on-policy optimization method (PPO) for end-to-end RL training. To ensure consistency with the platform utilized in \cite{lee2024learning}, this comparison is conducted using the simulated wheeled ANYmal~\citep{hutter2016anymal} robot model, which differs from the robot model used in the other comparisons in this paper. All policies are trained under identical conditions and evaluated on an independent test environment set to ensure a fair comparison. Note that our policy and the GTRL baseline rely solely on a front-facing camera with a limited field of view, whereas the EMHP approach incorporates a local height scan with a similar range for environmental detection but benefits from a complete 360-degree field of view for mapping.

  The experimental results (see Table~\ref{tab:baseline}) indicate that our architecture, leveraging an implicit memory representation within the recurrent module, outperforms both baselines. Compared to EMHP, it achieves a 29.6\% relative improvement in SR. Against GTRL, it demonstrates a remarkable 105.0\% relative improvement. These results underscore the limitations of explicit mapping or stacked-frame histories for long-horizon navigation under limited context. Figure~\ref{fig:baseline_maze_compare} illustrates a representative comparison: in a long-horizon maze environment, the EMHP approach, despite achieving a relatively higher SR among the two baselines, is constrained by the limited horizon of its explicit memory, fails to navigate through the maze and eventually loops in a long corridor. In contrast, our policy successfully traverses a dead-end corridor and reaches the goal, demonstrating the effectiveness of the SRU unit in learning implicit spatial-temporal mapping for long-horizon navigation tasks. To further validate the effectiveness of SRU for implicit spatial-temporal memorization in mapless navigation, we replaced GTRL’s stacked historical observations with our SRU-based recurrent memory while keeping all other components unchanged. This modified variant, denoted GTRL*, achieves a 73.6\% relative improvement over the original GTRL baseline, increasing the SR from 38.2\% to 66.3\% (Table~\ref{tab:baseline}). This substantial gain highlights the advantage of SRU’s implicit recurrent memory over heuristic stacking of historical observations in capturing spatial-temporal dependencies for improved navigation performance. Notably, compared to GTRL*, our complete SRU-based approach (Ours) achieves an additional 18.1\% relative improvement, attributable to our proposed spatial attention layers. The impact of these attention layers is further discussed later.

\begin{figure}
    \centering
    \includegraphics[width=0.45\textwidth]{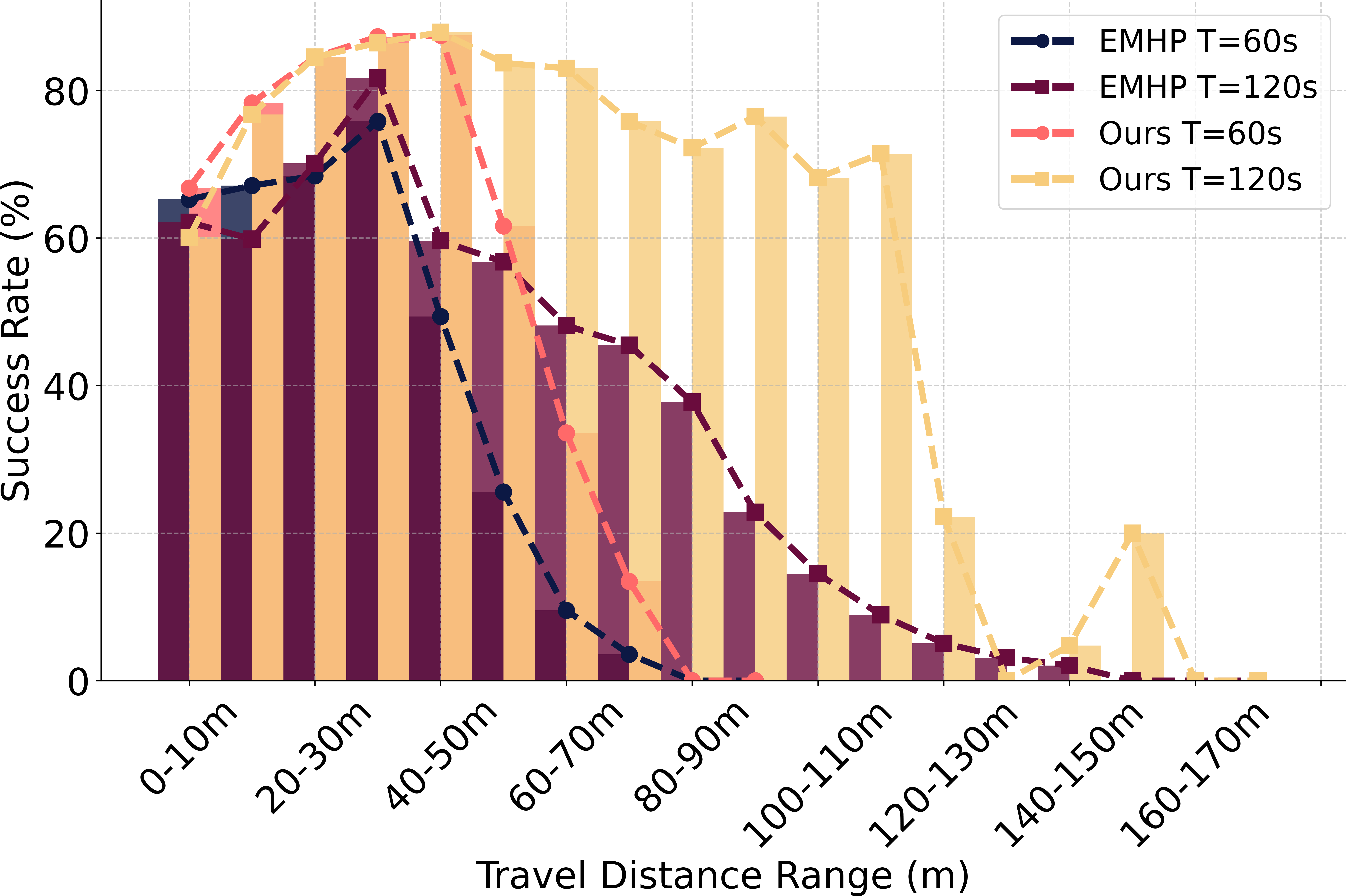}
    \caption{Success rate sorted by travel distance: comparison between the EMHP baseline approach, which uses explicit mapping and a fixed-length historical path, and our approach, which employs the implicit recurrent memory of SRU. 
    Our method maintains a high success rate over longer distances and extends effectively with longer episodic times. 
    In contrast, the baseline's success rate drops significantly for longer travel distances, even when the maximum episodic time is doubled, due to its fixed context window limitation.}
    \label{fig:success_by_distance}
  \end{figure}

  Quantitatively, while the EMHP approach achieves comparable navigation performance, we also analyze the success rate (SR) as a function of travel distance to evaluate its long-range memorization and generalization capabilities. As shown in Figure~\ref{fig:success_by_distance}, with a maximum episodic time of 60 seconds (consistent with training) and the robot's maximum speed set to 1.5 m/s, the EMHP approach's SR drops significantly when the travel distance exceeds 40 meters. 
  In contrast, our SRU-based approach maintains an SR of over 80\% up to 50 meters. 
  When the maximum episode time is extended to 120 seconds, the EMHP's SR still declines to below 60\% at the same 40-meter distance, constrained by its fixed context window. Conversely, our SRU-based approach sustains an SR of over 70\% for distances up to 120 meters, demonstrating the SRU's superior ability to implicitly learn spatial-temporal mappings and generalize to distances beyond the training range. The baseline's reliance on a fixed explicit memory window limits its capacity to capture long-range dependencies, hindering its generalization in extended long-distance navigation tasks.

  Furthermore, we observed that the EMHP policy struggles to effectively learn to climb staircases unless dense reward guidance is provided. We believe this limitation arises from the inherent difficulty explicit memory mechanisms face in capturing intricate spatial-temporal features, which are essential for the robot to develop the maneuvers required to overcome 3D obstacles effectively. Lastly, the end-to-end recurrent setup offers a simpler and more maintainable solution compared to baseline methods. In contrast, baselines rely on an external mapping pipeline and the storage of additional historical paths or observations for each robot, which can introduce complexity and overhead during both training and real-world deployment.

  \begin{figure*}
    \begin{center}
      \subfigure[Office Environment]{%
      \includegraphics[width=0.32\textwidth]{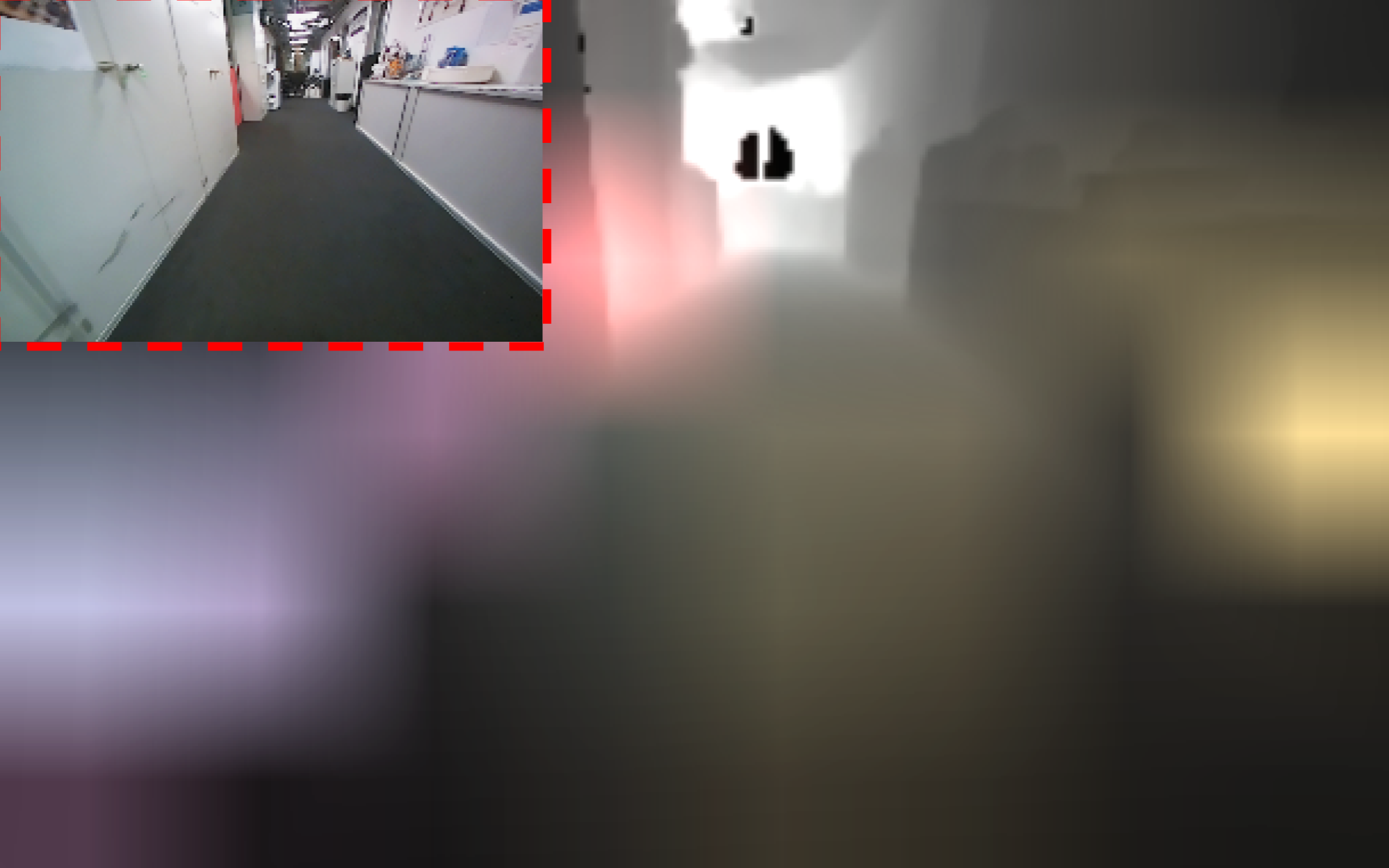}%
      \label{fig:attention_office}%
      } \subfigure[Terrace Environment]{%
      \includegraphics[width=0.32\textwidth]{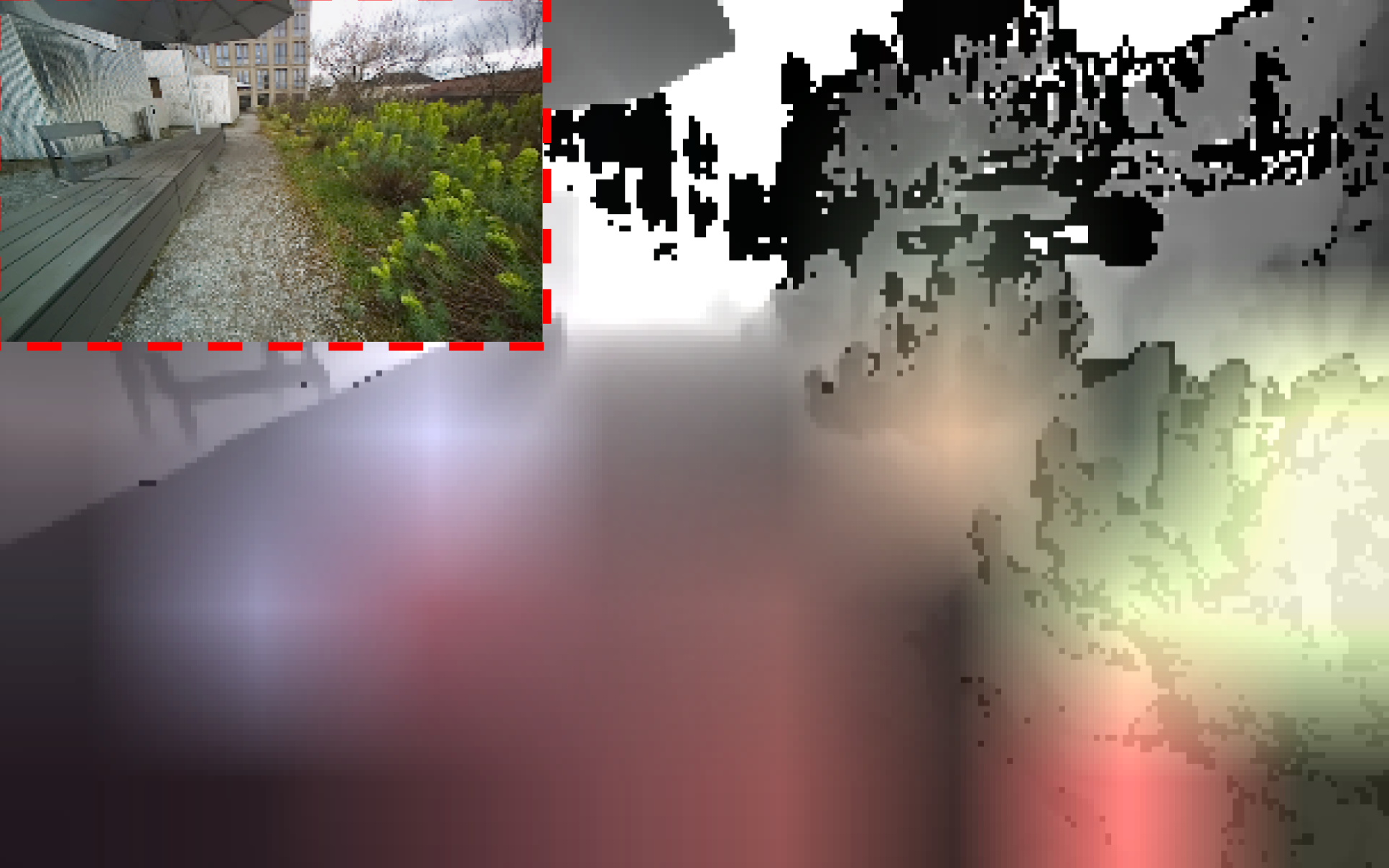}%
      \label{fig:attention_outdoor}%
      } \subfigure[Forest Environment]{%
      \includegraphics[width=0.32\textwidth]{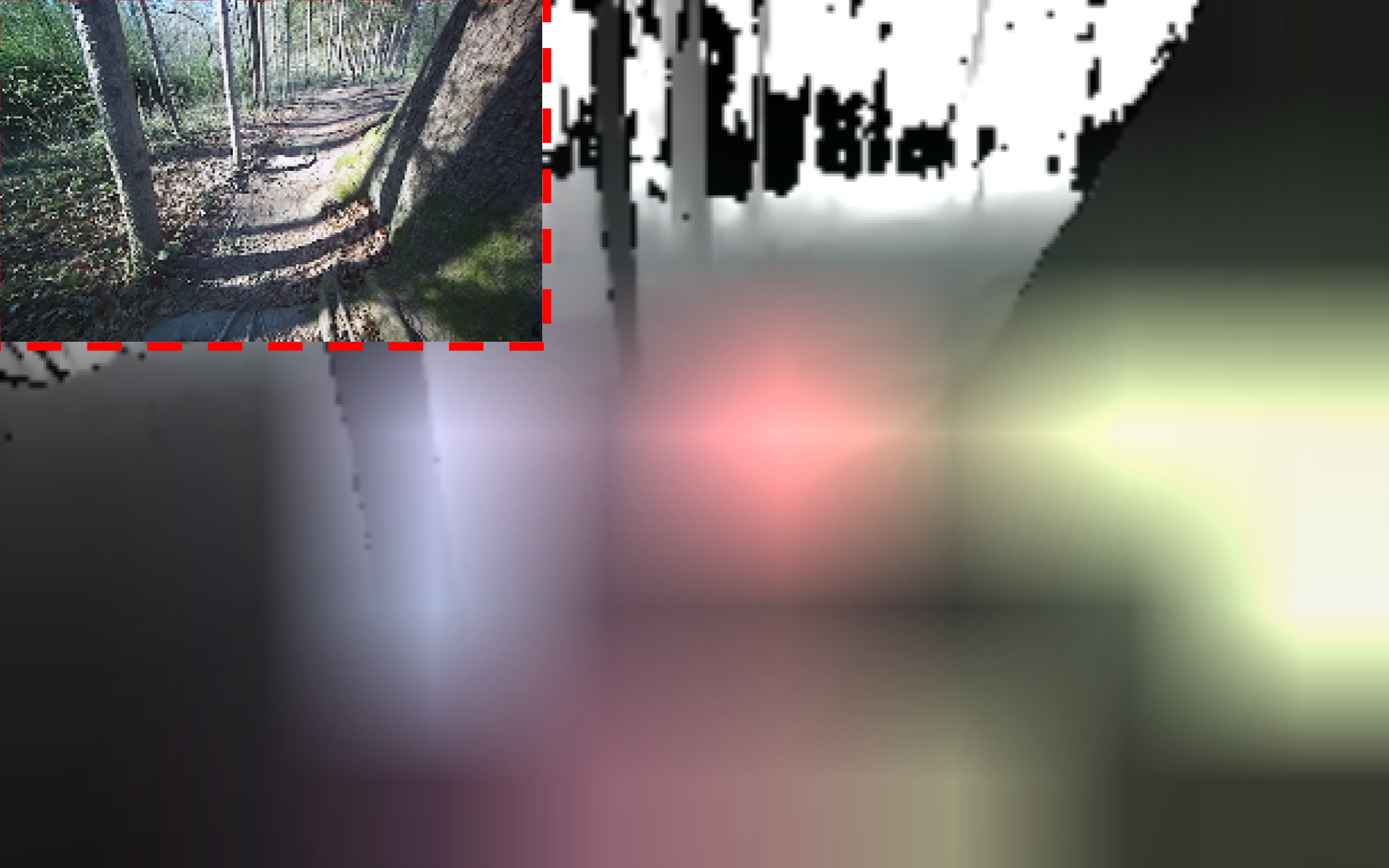}%
      \label{fig:attention_forest}%
      }
      \caption{Visualization of attention weights for the cross-attention layer in three distinct real-world deployment scenarios over raw depth inputs: (a) Office environment, (b) Outdoor terrace environment, and (c) Forest environment. The attention weights dynamically highlight relevant spatial cues for navigation based on the robot's state. The RGB images in the top corners are included for visualization purposes only.} 
      \label{fig:attention_real_world}
    \end{center}
  \end{figure*}

  \begin{table}
    \begin{center}
      \begin{tabular}{c|c}
        \toprule \textbf{Method}                                             & \textbf{SR (\%)} \\
        \midrule GTRL (w. historical obs.) & 38.2             \\
        EMHP (w. explicit mapping / path) & 60.4             \\
        \addlinespace[0.5em] \hdashline \addlinespace[0.5em] GTRL* (w. SRU memory)                                 & 66.3             \\
        Ours (w. SRU memory)                                 & \textbf{78.3}    \\
        \bottomrule
      \end{tabular}
      \vspace{0.5em}
      \caption{Overall navigation success rate (SR) comparison against baselines. Policies with SRU implicit recurrent memory outperform: (i) GTRL (stacked historical observations) and (ii) EMHP (explicit mapping and historical path). GTRL* denotes our modified GTRL variant where stacked observations are replaced by SRU implicit memory, yielding a substantial gain over the original GTRL baseline.\label{tab:baseline}}
    \end{center}
  \end{table}

  \begin{figure}
    \begin{center}
      \includegraphics[width=0.45\textwidth]{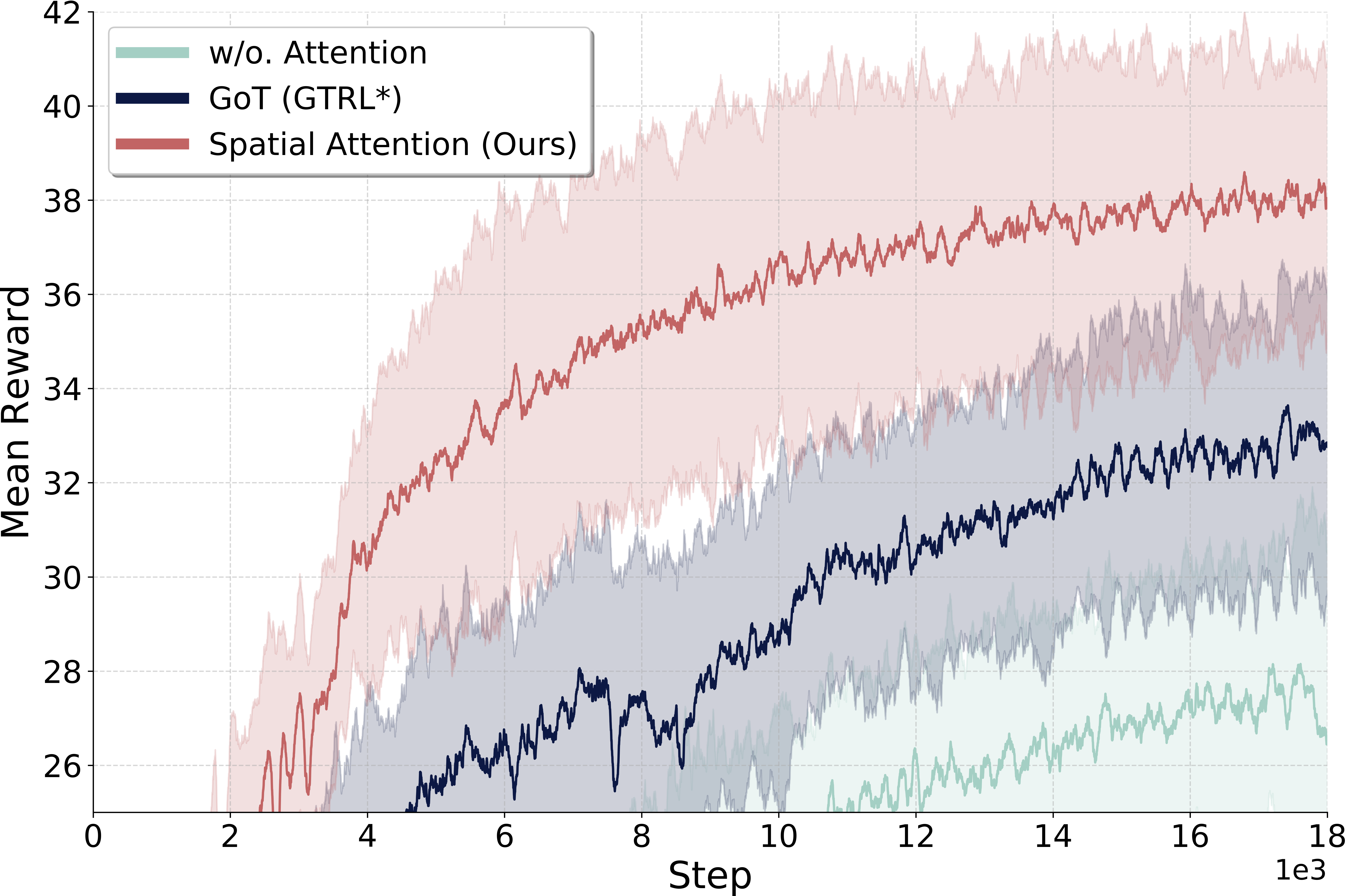}
      \caption{Average training return rewards for attention ablations (all using SRU recurrent memory): (1) without attention (w/o.)~\cite{wijmans2019dd}; (2) Goal-guided Transformer (GoT) attention~\cite{huang2023goal}; and (3) the proposed two-stage spatial attention (Ours). The proposed spatial attention achieves the highest returns, indicating more effective extraction of task-relevant spatial cues for improved recurrent memorization.}
      \label{fig:attention_mask_curve}
    \end{center}
  \end{figure}

  \begin{table}
    \begin{center}
      \begin{tabular}{c|c}
        \toprule
        \textbf{Attention Configuration}          & \textbf{SR (\%)} \\
        \midrule
        w/o. Attention  & 50.5             \\
        GoT (GTRL*)         & 68.4 \\
        Spatial Attention (Ours)   & 78.9             \\
        \bottomrule
      \end{tabular}
      \vspace{0.5em}
      \caption{Navigation success rate (SR) for policies integrated with different attention configurations. The policy with the proposed spatial attention (Ours) achieves the highest SR, outperforming (i) the baseline without attention~\cite{wijmans2019dd} and (ii) the Goal-guided Transformer (GoT)~\cite{huang2023goal} integrated with SRU memory (GTRL* approach in Sec.~\ref{sec:compare-baseline}), highlighting the importance of an attention mechanism for training mapless navigation end-to-end and the effectiveness of the proposed spatial attention structure.\label{tab:attention_mask}}
    \end{center}
  \end{table}

  \subsection{Importance of Spatial Attention Layers}

  We now examine the role of the proposed spatial attention layers in the network architecture and evaluate their impact on navigation performance. These layers are designed to compress and emphasize relevant features from encoded observations, addressing a key challenge faced by recurrent structures: the difficulty of retaining long-term information due to the exponential decay of memory over time. 
  By selectively focusing on the most salient features, the attention mechanism emphasizes the most relevant spatial cues for navigation based on the robot's state and reduces the information density passed into the recurrent memory at each step. We hypothesize that this mechanism can improve the network's memorization and navigation capabilities, enabling it to handle complex, long-range tasks more effectively.
  
  To test this, we conduct an ablation study by: (i) removing the attention layers from our network architecture and replacing them with convolution followed by average pooling for feature compression, as implemented in \cite{wijmans2019dd}, and (ii) comparing the performance of our proposed spatial attention layers against the Goal-guided Transformer (GoT) architecture proposed in \cite{huang2023goal}. The GoT architecture utilizes a modified Vision Transformer (ViT) that integrates the goal state as an additional token. It performs self-attention across both visual feature tokens and the goal token to extract goal-relevant features. In contrast, our approach first applies self-attention exclusively to visual tokens to enhance spatial features. Subsequently, the goal and proprioceptive state are used as queries in the cross-attention layer to compress and extract the most relevant features. For a fair comparison in the ablation experiments, we use identical training settings for all approaches, integrating the SRU memories and the pretrained encoder while varying only the attention layers used to process visual features during RL. The GoT integrated with SRU is the the GTRL* approach, as described in Sec.~\ref{sec:compare-baseline}. Figure~\ref{fig:attention_mask_curve} shows the average return rewards during training. The network without the attention layers exhibits significantly lower performance compared to the two policies utilizing attention mechanisms. Additionally, our proposed spatial attention layers outperform the GoT attention mechanism. Table~\ref{tab:attention_mask} shows the SR performance of the three configurations: (i) without attention, (ii) with GoT attention, and (iii) with our proposed spatial attention (Ours). Our method achieves a 56.2\% relative SR improvement over the no-attention baseline, highlighting the importance of selectively compressing and extracting spatial features for long-range mapless navigation when utilizing implicit recurrent memory. Furthermore, it achieves a 15.4\% relative improvement (18.1\% when trained with the ANYmal robot model, as shown in Table~\ref{tab:baseline}) over the policy utilizing GoT attention. This demonstrates that the proposed two-stage spatial attention mechanism more effectively extracts task-relevant cues, enhancing recurrent memorization and policy optimization.

  Notably, the attention effect emerges naturally during the end-to-end RL training without requiring additional supervision or auxiliary losses. Figure~\ref{fig:attention_real_world} illustrates the attention weights generated by the cross-attention layer over raw visual inputs in three distinct real-world deployment scenarios: an indoor office, an outdoor terrace, and a forest environment. The attention weights, with four attention heads (depicted in different colors), dynamically emphasize the most relevant spatial cues, such as obstacles and navigable free space, based on the robot's state at the time the depth input was recorded. 
  This highlights the effectiveness of training the spatial attention mechanism end-to-end and its ability to generalize across diverse and challenging environments.

  \subsection{Training with Regularizations}
  We evaluate the role of regularization techniques in the end-to-end training of the recurrent network using reinforcement learning.
  Firstly, as shown in Figure~\ref{fig:spatial_temporal_mapping}
  and discussed in Sec.~\ref{sec:reward-regularization}, while the SRU unit effectively enhances the network's ability to learn implicit spatial memorization from sequential observations, the learning curve indicates that spatial memory learning can converge significantly slower than temporal memorization. 
  This discrepancy, combined with the inherent properties of standard policy optimization algorithms like PPO—which restrict deviations from previous optimization steps—and the complex structure of attention networks with RNNs prone to overfitting, suggests that without proper regularization, the network may converge to suboptimal strategies. 
  Such strategies might overly rely on easier-to-learn temporal features to solve navigation tasks, thereby failing to establish robust spatial-temporal memorization. 
  To test this hypothesis, we conduct an ablation study by removing the regularization techniques, specifically deep mutual learning (DML), from the standard PPO training setup and comparing the performance against the setup with DML regularization.

\begin{figure}
    \begin{center}
      \includegraphics[width=0.45\textwidth]{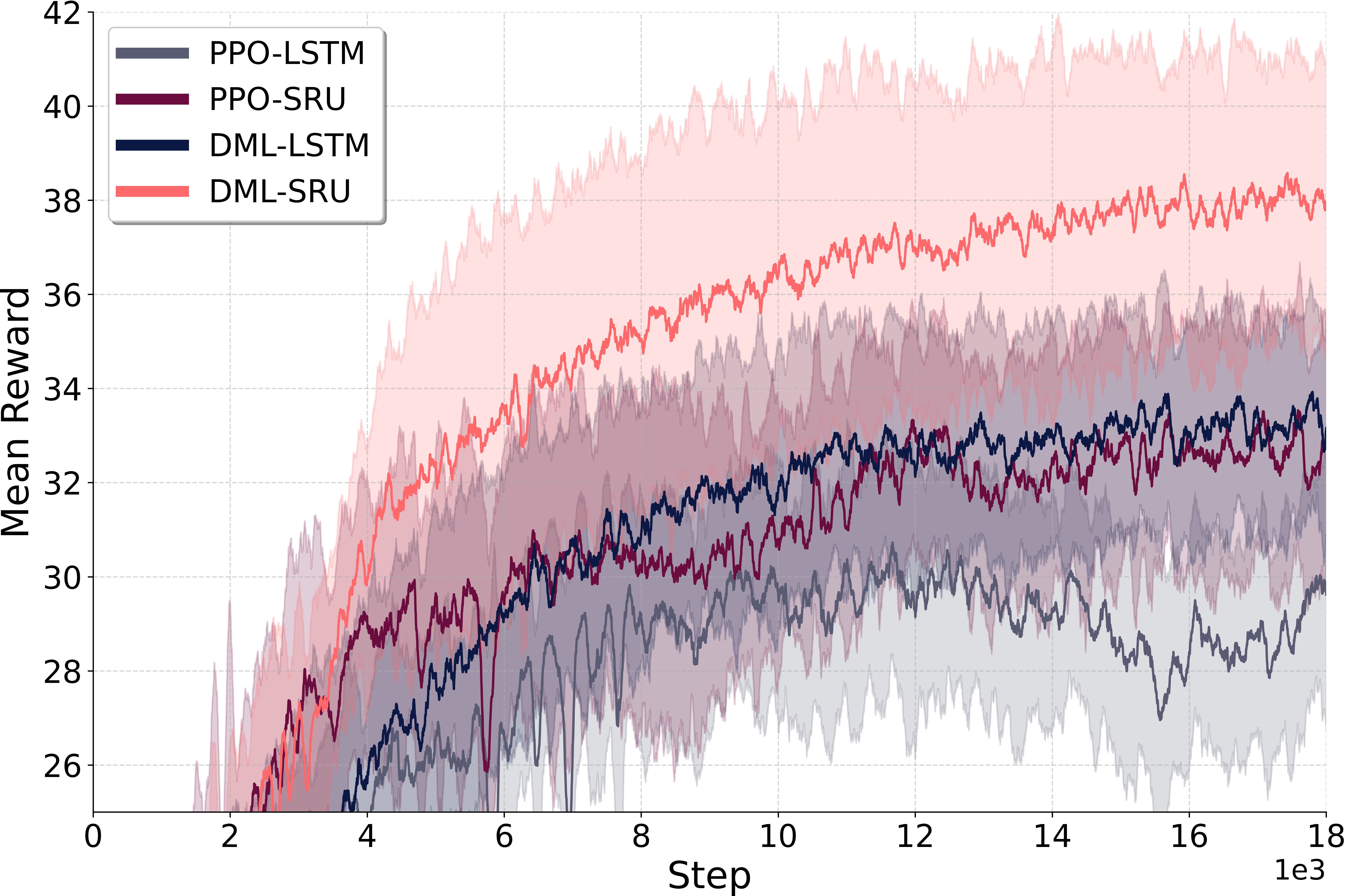}
      \caption{Training curve comparison between policies trained using PPO with deep mutual learning (DML) regularization and PPO: The network with DML regularization techniques achieves higher returns compared to the network trained with vanilla PPO.}
      \label{fig:regularization_curve}
    \end{center}
  \end{figure}

  Figure~\ref{fig:regularization_curve} illustrates that the network without DML exhibits lower average return rewards during training.
  Notably, the performance difference between standard LSTM and SRU modifications becomes more pronounced when regularization techniques are applied. 
  As shown in Table~\ref{tab:regularization}, the SR performance improves from 61.8\% to 65.7\% (a 6.3\% increase) without DML and from 63.5\% to 78.9\% (a significant 24.3\% increase) with DML. 
  This finding underscores that, in certain RL tasks, the network's architecture alone may not be the sole limiting factor. 
  Instead, the optimization process plays a critical role in fully leveraging the network's potential, highlighting the importance of effective training strategies.

  Additionally, we observe that incorporating the consistent dropout layer with temporal consistency into the recurrent training can also positively impact navigation performance, as shown in Table~\ref{tab:regularization_tcd}. 
  This enhancement improves the SR when tested in new, randomly generated environments. 
  These findings align with the discussion in \cite{hausknecht2022consistent}, which highlights the benefits of using dropout in RL training to enhance the network's generalization and robustness.
  
  \subsection{Large-Scale Pretraining for Sim-to-Real Transfer}
  In this section, we evaluate the pretrained image encoder, trained on a large-scale synthetic dataset, for its ability to bridge the sim-to-real gap in real-world perception. 
  Additionally, we assess the effectiveness of the proposed depth noise model in reducing discrepancies between synthetic and real-world data.
  To this end, we conduct zero-shot transfer experiments on legged-wheel platforms across diverse real-world environments to demonstrate the generalization of our approach.

  \begin{table}
    \begin{center}
      \begin{tabular}{l|c}
        \toprule \textbf{RL Training}                                          & \textbf{SR \%} \\
        \midrule LSTM w/o. DML                                                 & 61.8           \\
        LSTM w. DML                                                            & 63.5           \\
        \addlinespace[0.5em] \hdashline \addlinespace[0.5em] SRU-Ours w/o. DML & 65.7           \\
        SRU-Ours w. DML                                                        & 78.9           \\
        \bottomrule
      \end{tabular}
      \vspace{0.5em}
      \caption{Comparison of the overall navigation success rate (SR) with and without DML regularization for policies using LSTM and SRU units. DML significantly enhances SR for SRU (over 20\%) and provides a marginal improvement for LSTM (2.8\%), showcasing DML's effectiveness in unlocking SRU's potential for long-range mapless navigation. \label{tab:regularization}}
    \end{center}
  \end{table} 

  \begin{table}
    \begin{center}
      \begin{tabular}{l|c}
        \toprule \textbf{RL Training} & \textbf{SR (\%)} \\
        \midrule SRU-Ours w/o. TC-D   & 77.2             \\
        SRU-Ours w. TC-D              & 78.9             \\
        \bottomrule
      \end{tabular}
      \vspace{0.5em}
      \caption{Evaluation of the overall navigation success rate (SR) with and without temporally consistent dropout (TC-D): The network with TC-D is able to maintain a similar (or even higher) SR compared to the network without TC-D, while improving robustness and generalization.\label{tab:regularization_tcd}}
    \end{center}
  \end{table}

  \begin{figure*}
    \begin{center}
      \subfigure[Real-world Stereo Depth Image]{%
      \includegraphics[width=0.3\textwidth]{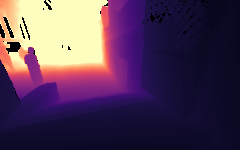}%
      \label{fig:original_stereo_depth}%
      } \hspace{0.02\textwidth} \subfigure[Reconstruction with Large-scale Pretrain]{%
      \includegraphics[width=0.3\textwidth]{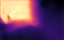}%
      \label{fig:pretrain_large_scale}%
      } \hspace{0.02\textwidth} \subfigure[Reconstruction with RL-images Pretrain]{%
      \includegraphics[width=0.3\textwidth]{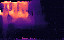}%
      \label{fig:pratain_rl_only}%
      }
      \caption{Comparison of depth image reconstruction using features from encoders
      pretrained on different data sources. (a) Original input stereo depth image from real-world deployment, captured using the \textit{ZEDX} camera. (b) Reconstructed depth image using features
      extracted from the encoder pretrained on large-scale synthetic data with noise augmentation. (c)
      Reconstructed depth image using features extracted from the encoder
      trained exclusively on simulated images collected during RL navigation
      training.}
      \label{fig:pretrain_decoder}
    \end{center}
  \end{figure*}

  \begin{figure}
    \begin{center}
      \subfigure[Latent space distribution with large-scale data pretraining]{%
      \includegraphics[width=0.4\textwidth]{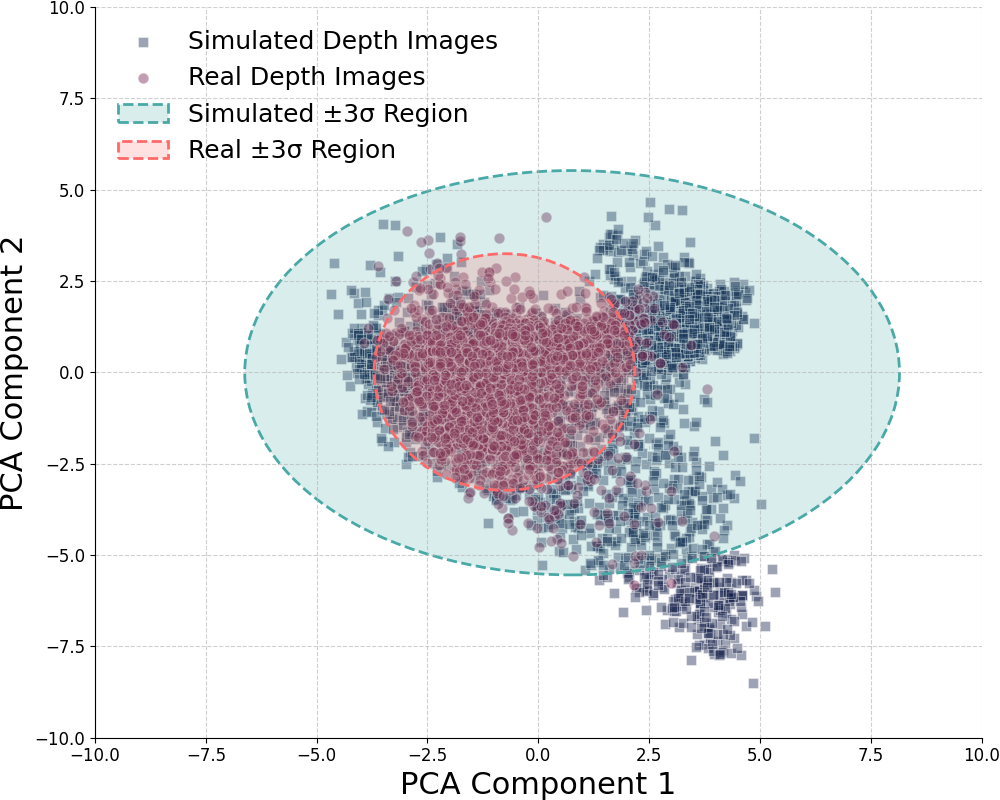}%
      \label{fig:comparison_pretrained}%
      } \hspace{0.05\textwidth} \subfigure[Latent space distribution with RL images
      pretraining]{%
      \includegraphics[width=0.4\textwidth]{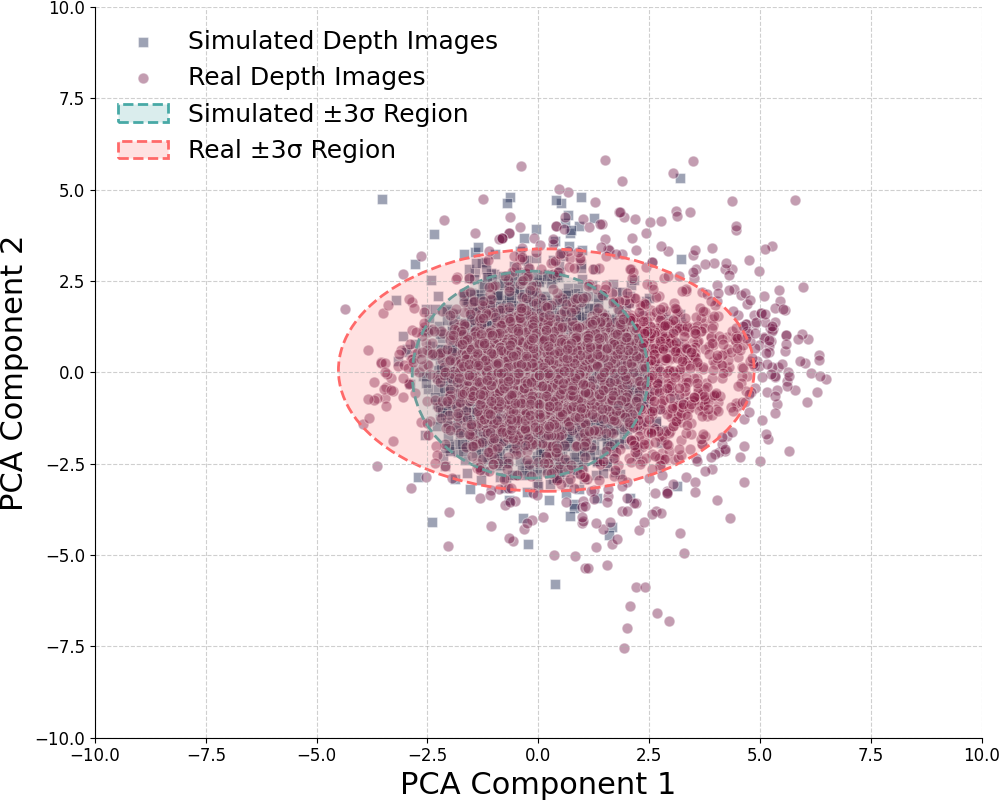}%
      \label{fig:rl_only}%
      }
      \caption{Comparison of latent space distributions: (a) The feature distribution from the encoder pretrained on large-scale synthetic data effectively covers the distribution of real-world data, indicating better generalization. (b) The feature distribution from the encoder trained solely on simulated data collected during RL fails to cover the distribution of real-world depth images, posing challenges in generalizing to real-world data.}
      \label{fig:noise_distribution}
    \end{center}
  \end{figure}

  \begin{figure}
    \begin{center}
      \includegraphics[width=0.4\textwidth]{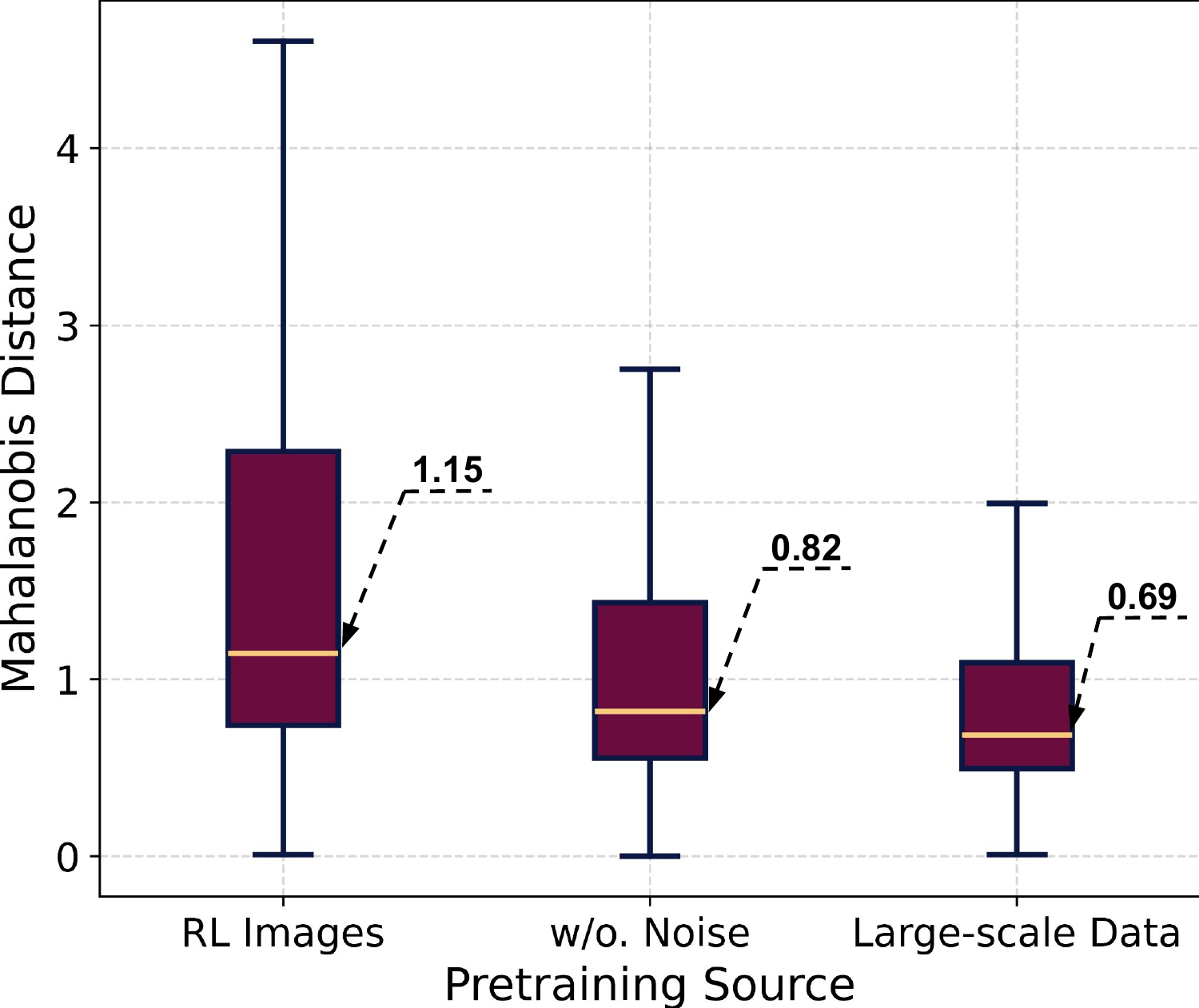}
      \caption{Comparison of Mahalanobis distances between the latent features of real-world images and the latent feature distributions of RL images, using encoders pretrained on different sources: (i) RL images, (ii) large-scale synthetic data without noise augmentation, and (iii) large-scale synthetic data with noise augmentation.}
      \label{fig:mahalanobis_distance_boxplot}
    \end{center}
  \end{figure}

  \subsubsection{Pretrain and Depth Noise.}

  Here, we analyze the latent space distribution of encoders trained under two distinct conditions: (i) an encoder trained exclusively on simulated depth images generated during RL navigation training (RL images), (ii) an encoder pretrained on large-scale synthetic data from \cite{wang2020tartanair}, augmented with the proposed parallelizable depth noise model (Figure~\ref{fig:depth_noise}). To evaluate these encoders, we compare the latent features extracted from their outputs using two data sources: (i) RL images, and (ii) real-world stereo depth images captured by the \textit{ZEDX} camera during deployment (real-world images). This analysis highlights their differences in latent space distributions depending on the pretraining data source used for the encoder.

  Figure~\ref{fig:noise_distribution} illustrates a 2D principal components analysis (PCA)~\citep{dunteman1989principal} projection of the latent features. The latent space distribution of RL-images shows a larger distribution range that encompasses the features extracted from real-world data when derived from the encoder pretrained on large-scale synthetic data (Figure~\ref{fig:comparison_pretrained}). This indicates that the encoder pretrained on large-scale synthetic data effectively captures a wide range of features, enabling it to generalize well to real-world scenarios. In contrast, the encoder trained solely on RL images (Figure~\ref{fig:rl_only}) exhibits a narrower latent space distribution, failing to cover many real-world features. This suggests that an encoder trained exclusively on simulated depth images collected during RL navigation training may struggle to generalize effectively to real-world data when deployed.

  Additionally, Figure~\ref{fig:pretrain_decoder} provides a qualitative comparison of depth reconstruction using features extracted from the same two pretrained encoders. The comparison is based on a real-world stereo depth input captured during deployment. The encoder pretrained on large-scale data demonstrates effective reconstruction of the depth image, with only minor blurring effects (Figure~\ref{fig:pretrain_large_scale}). In contrast, the encoder trained solely on RL images struggles to reconstruct the input depth image effectively, resulting in outputs with significant artifacts and noise (Figure~\ref{fig:pratain_rl_only}).

  To quantitatively assess the distributional disparity of features from encoders trained on different sources, we adapt the method from \cite{lee2018simple} to measure the Mahalanobis distance (MD) for the latent distributions derived from each encoder. In addition to the two pretraining sources mentioned earlier, we also analyze the latent distribution of the encoder pretrained on large-scale synthetic data without noise augmentation. This allows us to evaluate the effectiveness of the proposed depth noise model in further reducing the sim-to-real gap between synthetic depth images and real-world stereo depth. The MDs are computed between the latent features of real-world images and the latent feature distributions of RL images extracted from each encoder. As shown in Figure~\ref{fig:mahalanobis_distance_boxplot}, pretraining on large-scale synthetic data effectively reduces the MD, lowering the median from 1.15 (RL images) to 0.82 (large-scale synthetic data without noise). This demonstrates the pretrained encoder’s effectiveness in covering the distribution of real-world perception inputs. Furthermore, incorporating the proposed depth noise model further reduces the MD to 0.69, underscoring its role in narrowing the differences between synthetic and real-world depth data. These results highlight that the encoder, pretrained on large-scale data and augmented with the proposed depth noise model, can effectively minimize the sim-to-real gap, enabling improved generalization to real-world environments.


  \begin{figure}
    \begin{center}
      \subfigure[With SRU Memory Module]{%
      \includegraphics[width=0.42\textwidth]{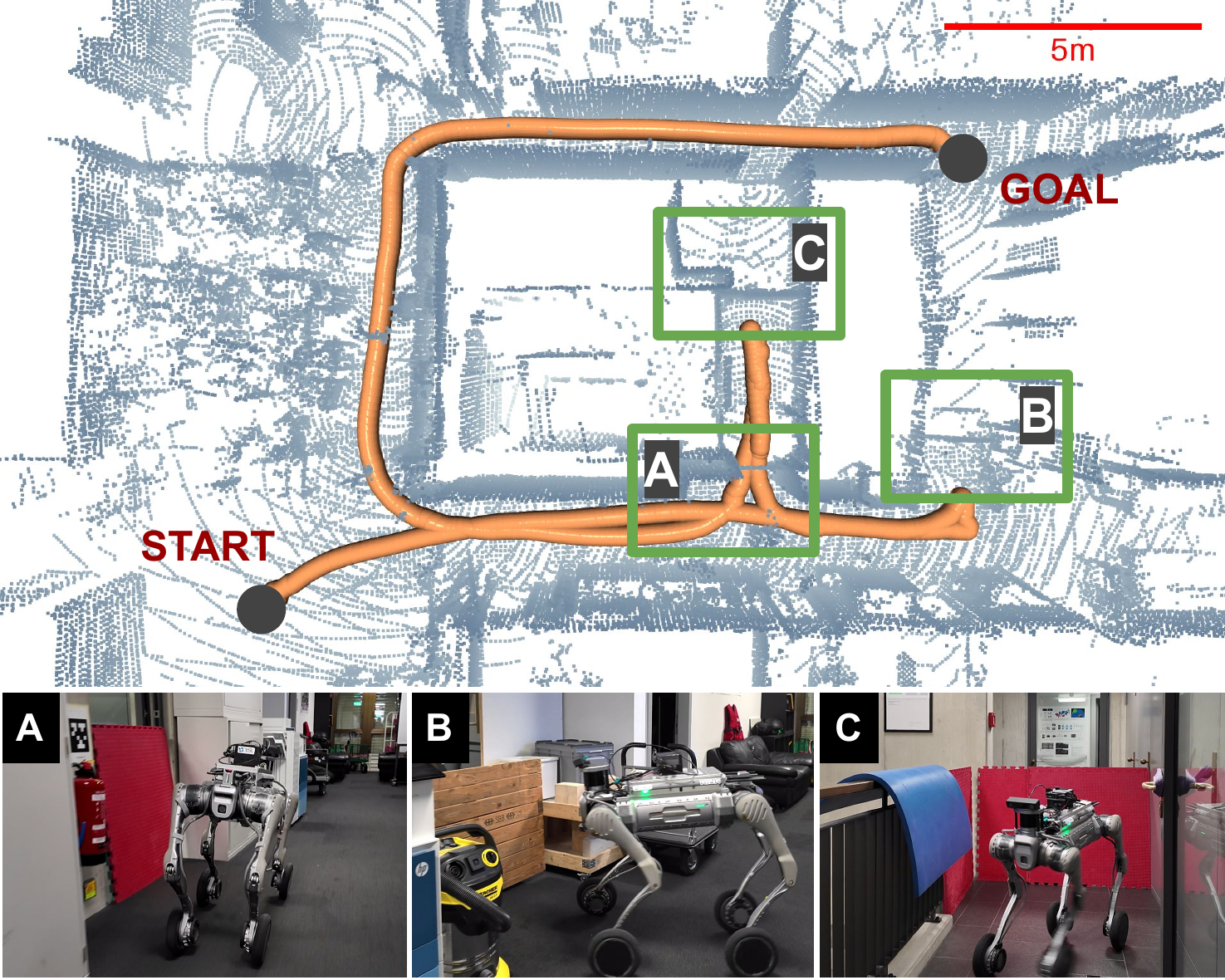}%
      \label{fig:office_real_world_sru}%
      } \subfigure[With Standard LSTM Memory Module]{%
      \includegraphics[width=0.42\textwidth]{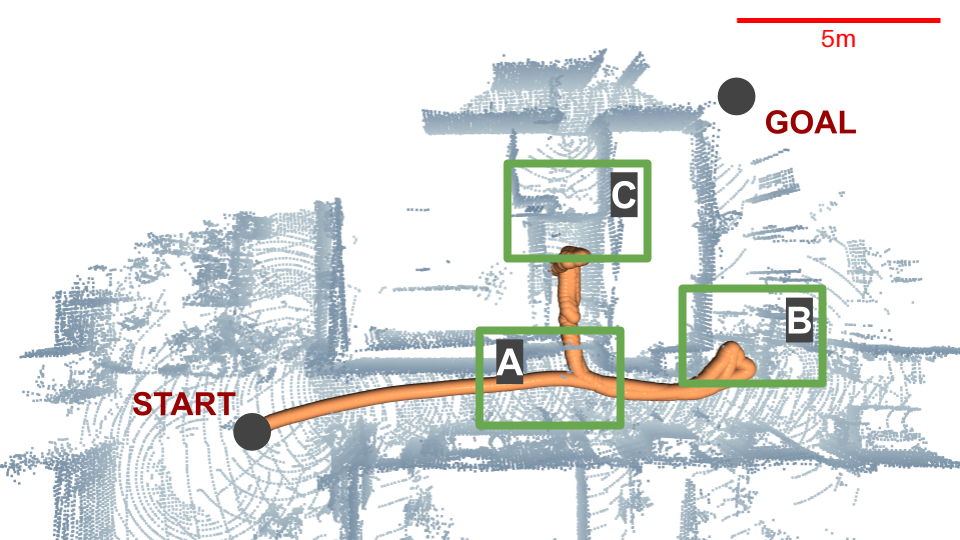}%
      \label{fig:office_real_world_lstm}%
      }
      \caption{Comparison of navigation trajectories (orange) in an office environment. A, B, and C indicate areas that the robot traverses in sequence. (a) shows that the robot using the SRU memory module successfully
      navigates through two dead ends and reaches the goal while adapting to changes in the environment (the blocker located in area A was initially set and later
      removed). (b) illustrates that the baseline model with a standard LSTM fails to reach the goal and repeatedly loops between the dead-end areas C and B.}
      \label{fig:office_real_world}
    \end{center}
  \end{figure}

  \begin{figure}
    \begin{center}
      \subfigure[Main Hall]{%
      \includegraphics[width=0.42\textwidth]{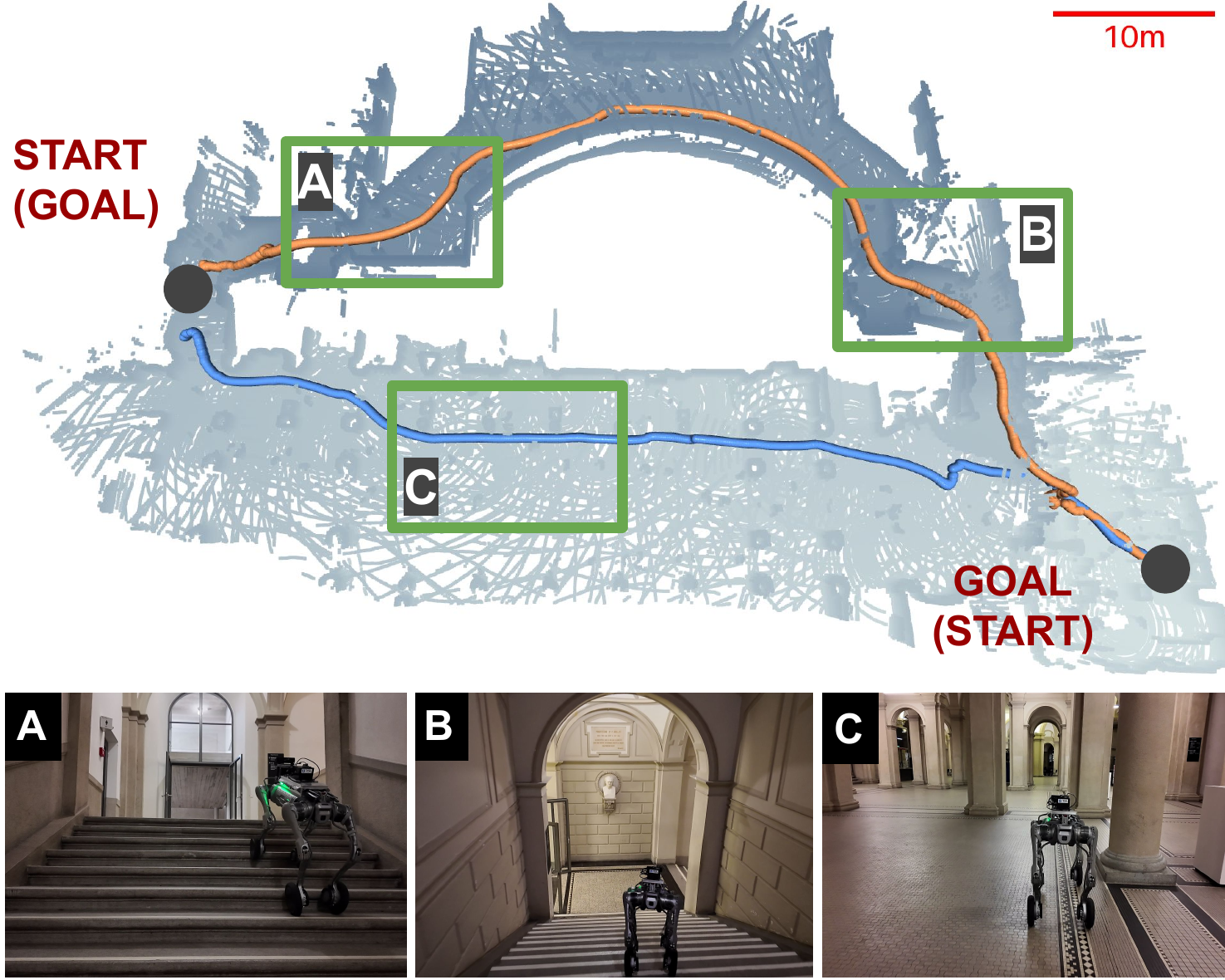}%
      \label{fig:trajectory_main_campus}%
      } \subfigure[Terrace]{%
      \includegraphics[width=0.42\textwidth]{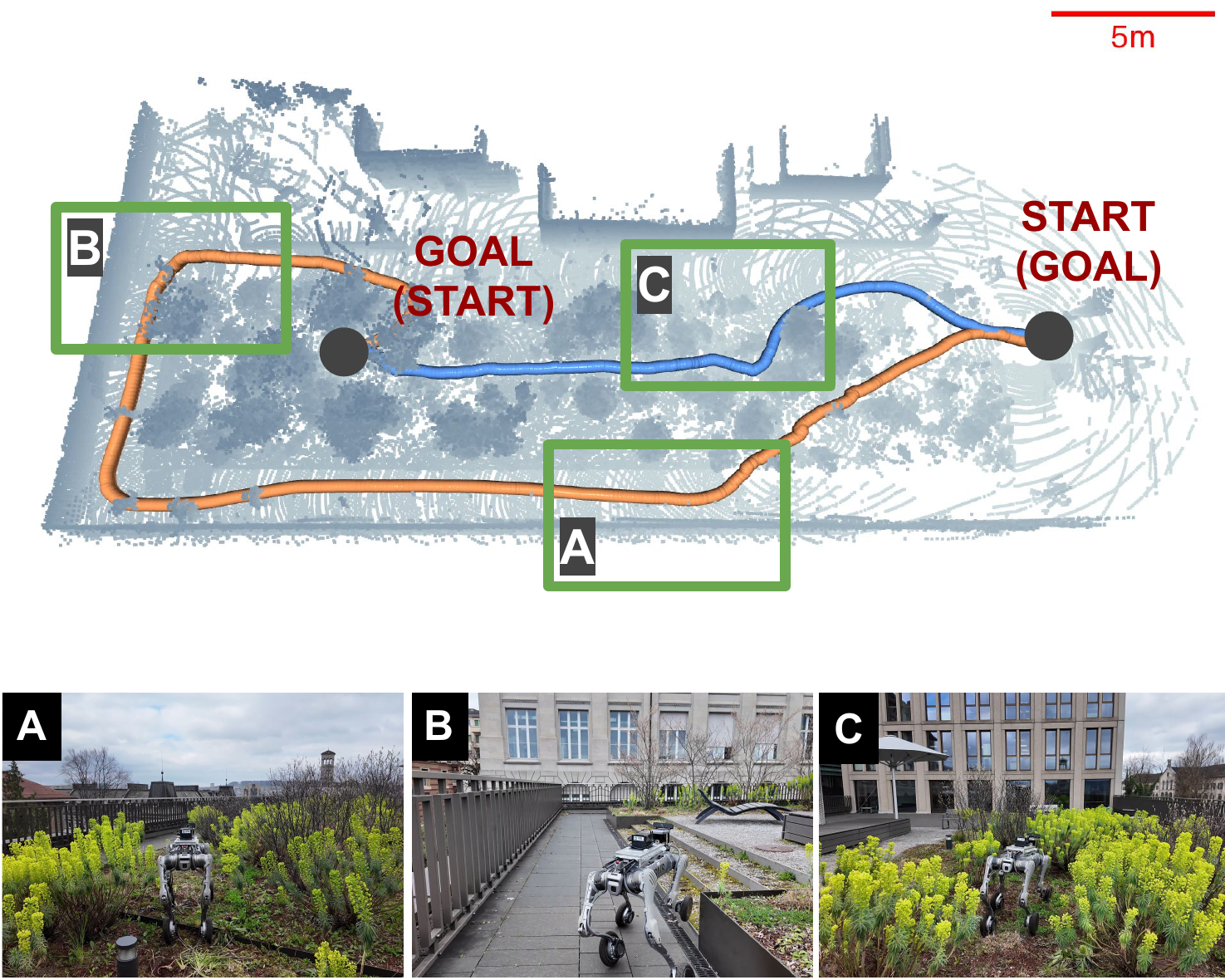}%
      \label{fig:trajectory_terrace}%
      } \subfigure[Forest]{%
      \includegraphics[width=0.42\textwidth]{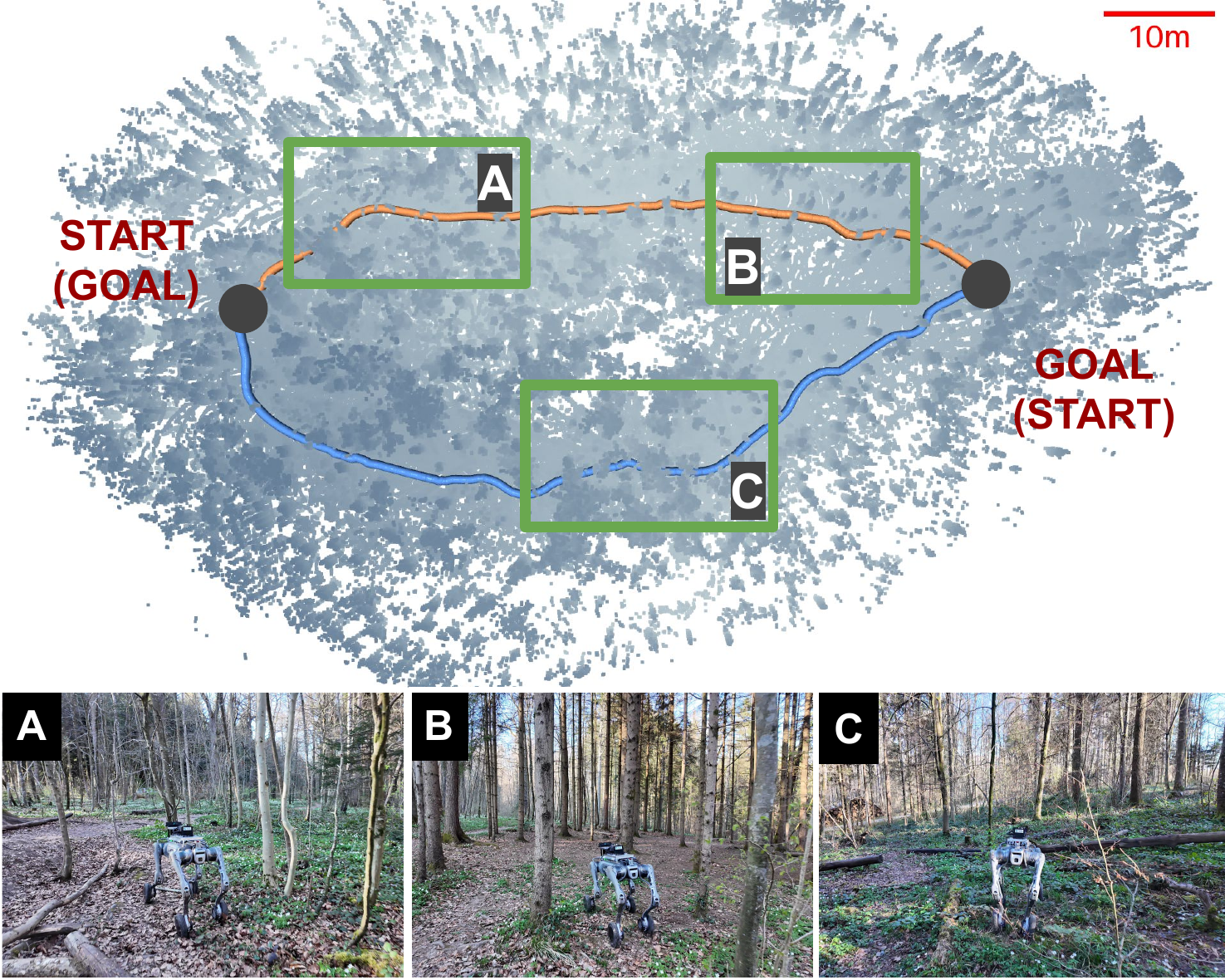}%
      \label{fig:trajectory_forest}%
      }
      \caption{Evaluation of long-range mapless navigation in diverse real-world environments: (a) Main Hall of a university, (b) outdoor terrace, and (c) forest environment with natural obstacles. In each scenario, the robot is tasked with two separate navigation goals (memory reset between goals), resulting in two trajectory segments (orange and blue). Labels A, B, and C mark key areas traversed by the robot.}
      \label{fig:real_world_enviroment}
    \end{center}
  \end{figure}

  \subsubsection{Real-world Tests on Legged-wheel Robot.}
  To evaluate the pretrained image encoder's with the proposed attention-based recurrent network's ability to generalize across in real-world environments, we conduct several zero-shot transfer experiments, on a \textit{Unitree B2W} robot with a learning-based locomotion policy from \textit{RIVR}. 
  The robot is mounted with a \textit{ZEDX}, front-facing stereo depth sensor, and \textit{NVIDIA Jetson AGX Orin} for onboard compute for the policy. 
  The pretrained encoder and network are directly deployed on the robot without any fine-tuning with real-world
  data. 
  For all the test, the robot receives no prior information about the environment, and receives only the stereo depth images from the front-facing camera as the exteroceptive input. 
  Additionally, a LiDAR-based state estimation and localization module~\citep{chen2022dlio} provide the robot's proprioceptive state, including linear and angular velocities $v_{t}$ and $\omega_{t}$, projected gravity $n_{t}$, and relative goal position $p_{t}$ with respect to the robot's frame. 
  The robot is controlled by a set of linear and angular velocities, referred to as action $a_{t}$, which is the input to the locomotion policy.

  Firstly, we conduct an experiment in an office environment, as shown in Figure~\ref{fig:office_real_world}, to compare the navigation performance of our policy with the SRU memory module against a baseline model using a standard LSTM unit. 
  In this experiment, the robot is tasked with navigating from one side of the office to the other while avoiding obstacles. 
  To evaluate the long-term spatial-temporal memorization capabilities of the SRU module, several passageways are temporarily blocked, requiring the robot to backtrack and search for alternative routes to reach the goal. 
  Additionally, dynamic changes are introduced by unblocking certain areas during navigation to further assess the robustness of the SRU-enhanced policy. 
  The policy with SRU demonstrates the ability to explore dead ends, navigate around obstacles, and re-evaluate its path to adapt to dynamic changes in the environment (Figure~\ref{fig:office_real_world_sru}).
  The robot successfully reaches the goal, showcasing the effectiveness of utilizing the SRU memory module to learn robust spatial-temporal memorization from sequential observations. 
  In contrast, the baseline model with a standard LSTM fails to reach the goal and repeatedly loops between dead-end areas, as shown in Figure~\ref{fig:office_real_world_lstm}.

  To further evaluate the generalization and performance of the proposed network architecture in long-horizon navigation tasks, we conduct experiments in a variety of real-world environments—including an indoor campus main hall, outdoor terrace areas, and forest environments—using the same pretrained encoder and navigation policy (see Figure~\ref{fig:real_world_enviroment}). 
  In these experiments, the robot is tasked with navigating to a designated goal and returning to its starting point.
  Note that the policy is designed to maintain episodic memory only between the start and the goal and is reset when a new goal is given. 
  The results demonstrate that the policy generalizes effectively to unseen environments, handling diverse obstacles such as walls, stairs, vegetation, bushes, and trees, as well as navigating uneven terrains. 
  Additionally, the policy adapts to larger-scale scenarios, including extended goal distances of more than 70 meters and traversing over 100 meters, as shown in Figure~\ref{fig:trajectory_forest}. 
  Note that the maximum start-goal distance during RL training is 30 meters. 
  The figures show the trajectories of the robot successfully navigating through these environments, with point clouds generated from the state estimation module~\citep{chen2022dlio} provided solely for visualization. 
  Note that, due to the absence of a dedicated mapping module or loop closure mechanism, the trajectories shown may exhibit some drift and errors.

  \section{Limitations and Future Work}
  While the proposed SRUs in this paper demonstrate significant improvements in spatial-temporal learning capabilities, their recurrent nature remains subject to exponential memory decay, which can limit their ability to retain global context over extended sequences. As a result, the long-range navigation capabilities presented in this paper are centered on local, mapless navigation using egocentric sensing. In this context, "long-range" refers to planning horizons that extend well beyond the local perception radius (e.g., 10 m), enabling rerouting from local dead ends without reliance on an explicit global map. Extending this approach to global-scale navigation—spanning kilometers or hours—would likely require additional mechanisms or architectural enhancements, such as the integration or maintenance of a global map.

  Furthermore, while SRUs enhance the network's capacity for implicit spatial memorization and improve long-range navigation performance, the precise characteristics of the information retained and utilized during end-to-end navigation training remain unclear. This highlights a broader challenge in explainable artificial intelligence, where understanding the internal representations and decision-making processes of neural networks continues to be an active area of research~\citep{mi2024toward}. Future work could explore integrating SRUs with recent advancements in foundation pretraining, such as DINO~\citep{caron2021emerging}, to combine their strengths in scene understanding with the efficiency of recurrent structures, further enhancing the policy's performance in complex real-world environments. Investigating auxiliary losses or additional regularization techniques to further leverage the potential of spatial-temporal memorization in SRUs could also be beneficial. Additionally, extending the application of SRUs to other domains, such as robotic manipulation and 3D reconstruction, could unlock new possibilities and advancements in spatial-temporal learning. In summary, while SRUs are effective, they represent a simple yet practical solution—not necessarily unique or optimal—that proves successful in our end-to-end mapless navigation context. More importantly, this work aims to highlight the potential of implicit spatial memory mechanisms in addressing complex navigation challenges while identifying opportunities for further exploration and optimization in both methodology and application domains.

  \section{Conclusion}
  In this study, we identify and address a limitation of existing recurrent neural network architectures in the context of navigation: while RNNs excel at modeling temporal sequences, they are not inherently designed for spatial memorization or transforming observations from varying perspectives. This limitation makes them less effective in building the spatial representations required for mapless navigation using egocentric perception. To address this, we propose Spatially-Enhanced Recurrent Units (SRUs), which integrate an implicit spatial transformation operation into standard GRU and LSTM structures. These SRUs are incorporated into a novel attention-based architecture, trained end-to-end via reinforcement learning, achieving long-horizon navigation tasks with a single forward-facing depth camera. Our research further highlights the importance of regularization strategies in end-to-end reinforcement learning frameworks. Techniques such as temporally consistent dropout and deep mutual learning are crucial for fully leveraging SRUs' potential and preventing early overfitting. Experiments demonstrate SRUs' superior navigation performance compared to standard LSTM and GRU models. Moreover, we compare our implicit recurrent memory-based approach with a state-of-the-art baseline that utilizes explicit mapping and historical paths. Our findings illustrate the superior effectiveness of recurrent memory structures for long-range mapless navigation tasks. Additionally, through ablation studies, we demonstrate the role of specific design choices, particularly the spatial attention mechanism, in enhancing overall navigation performance. Lastly, we analyze and address the challenge of sim-to-real transfer for stereo depth perception by integrating large-scale pretraining. This approach enables successful zero-shot transfer and robust generalization across diverse real-world environments, including indoor, outdoor, and forest scenarios, underscoring the practical applicability and effectiveness of our proposed methodology.

  \section*{Acknowledgements}
  The authors acknowledge Nikita Rudin, Takahiro Miki, Jonas Frey, Pascal Roth, and Chong Zhang for their valuable feedback and discussions. The authors also extend their gratitude to Marco Trentini for his assistance in conducting real-world experiments and testing the LiDAR-inertial state estimation module. Additionally, the authors recognize the \textit{RIVR} team for their technical support with the legged-wheel robot platform utilized in this research.

  \section*{Declaration of Conflicting Interests}
  The authors declared no potential conflicts of interest with respect to the research, authorship, and/or publication of this article.

    
  \bibliographystyle{SageH}
  \bibliography{citation.bib}

\begin{thebibliography}{81}
\providecommand{\natexlab}[1]{#1}
\providecommand{\url}[1]{\texttt{#1}}
\providecommand{\urlprefix}{URL }
\expandafter\ifx\csname urlstyle\endcsname\relax
  \providecommand{\doi}[1]{DOI:\discretionary{}{}{}#1}\else
  \providecommand{\doi}{DOI:\discretionary{}{}{}\begingroup \urlstyle{rm}\Url}\fi

\bibitem[{Barron and Malik(2013{\natexlab{a}})}]{Barron:etal:2013A}
Barron JT and Malik J (2013{\natexlab{a}}) Intrinsic scene properties from a single rgb-d image.
\newblock \emph{CVPR} .

\bibitem[{Barron and Malik(2013{\natexlab{b}})}]{barron2013intrinsic}
Barron JT and Malik J (2013{\natexlab{b}}) Intrinsic scene properties from a single rgb-d image.
\newblock In: \emph{Proceedings of the IEEE Conference on Computer Vision and Pattern Recognition}. pp. 17--24.

\bibitem[{Bhattacharya et~al.(2024)Bhattacharya, Rao, Parikh, Kunapuli, Wu, Tao, Matni and Kumar}]{bhattacharya2024vision}
Bhattacharya A, Rao N, Parikh D, Kunapuli P, Wu Y, Tao Y, Matni N and Kumar V (2024) Vision transformers for end-to-end vision-based quadrotor obstacle avoidance.
\newblock \emph{arXiv preprint arXiv:2405.10391} .

\bibitem[{Bohg et~al.(2014{\natexlab{a}})Bohg, Romero, Herzog and Schaal}]{Bohg:etal:2014}
Bohg J, Romero J, Herzog A and Schaal S (2014{\natexlab{a}}) Robot arm pose estimation through pixel-wise part classification.
\newblock \emph{ICRA} .

\bibitem[{Bohg et~al.(2014{\natexlab{b}})Bohg, Romero, Herzog and Schaal}]{bohg2014robot}
Bohg J, Romero J, Herzog A and Schaal S (2014{\natexlab{b}}) Robot arm pose estimation through pixel-wise part classification.
\newblock In: \emph{2014 IEEE International Conference on Robotics and Automation (ICRA)}. IEEE, pp. 3143--3150.

\bibitem[{Bohlin and Kavraki(2000)}]{bohlin2000path}
Bohlin R and Kavraki LE (2000) Path planning using lazy prm.
\newblock In: \emph{Proceedings 2000 ICRA. Millennium conference. IEEE international conference on robotics and automation. Symposia proceedings (Cat. No. 00CH37065)}, volume~1. IEEE, pp. 521--528.

\bibitem[{Bojarski et~al.(2016)Bojarski, Del~Testa, Dworakowski, Firner, Flepp, Goyal, Jackel, Monfort, Muller, Zhang et~al.}]{bojarski2016end}
Bojarski M, Del~Testa D, Dworakowski D, Firner B, Flepp B, Goyal P, Jackel LD, Monfort M, Muller U, Zhang J et~al. (2016) End to end learning for self-driving cars.
\newblock \emph{arXiv preprint arXiv:1604.07316} .

\bibitem[{Caron et~al.(2021)Caron, Touvron, Misra, J{\'e}gou, Mairal, Bojanowski and Joulin}]{caron2021emerging}
Caron M, Touvron H, Misra I, J{\'e}gou H, Mairal J, Bojanowski P and Joulin A (2021) Emerging properties in self-supervised vision transformers.
\newblock In: \emph{Proceedings of the IEEE/CVF international conference on computer vision}. pp. 9650--9660.

\bibitem[{C{\`e}sar-Tondreau et~al.(2021)C{\`e}sar-Tondreau, Warnell, Stump, Kochersberger and Waytowich}]{cesar2021improving}
C{\`e}sar-Tondreau B, Warnell G, Stump E, Kochersberger K and Waytowich NR (2021) Improving autonomous robotic navigation using imitation learning.
\newblock \emph{Frontiers in Robotics and AI} 8: 627730.

\bibitem[{Chen et~al.(2023)Chen, Nemiroff and Lopez}]{chen2022dlio}
Chen K, Nemiroff R and Lopez BT (2023) Direct lidar-inertial odometry: Lightweight lio with continuous-time motion correction.
\newblock \emph{2023 IEEE International Conference on Robotics and Automation (ICRA)} : 3983--3989\doi{10.1109/ICRA48891.2023.10160508}.

\bibitem[{Cho et~al.(2014)Cho, Van~Merri{\"e}nboer, Gulcehre, Bahdanau, Bougares, Schwenk and Bengio}]{cho2014learning}
Cho K, Van~Merri{\"e}nboer B, Gulcehre C, Bahdanau D, Bougares F, Schwenk H and Bengio Y (2014) Learning phrase representations using rnn encoder-decoder for statistical machine translation.
\newblock \emph{arXiv preprint arXiv:1406.1078} .

\bibitem[{Choi et~al.(2019)Choi, Park, Kim and Seok}]{choi2019deep}
Choi J, Park K, Kim M and Seok S (2019) Deep reinforcement learning of navigation in a complex and crowded environment with a limited field of view.
\newblock In: \emph{2019 International Conference on Robotics and Automation (ICRA)}. IEEE, pp. 5993--6000.

\bibitem[{Cimurs et~al.(2021)Cimurs, Suh and Lee}]{cimurs2021goal}
Cimurs R, Suh IH and Lee JH (2021) Goal-driven autonomous exploration through deep reinforcement learning.
\newblock \emph{IEEE Robotics and Automation Letters} 7(2): 730--737.

\bibitem[{Dijkstra(1959)}]{dijkstra1959note}
Dijkstra EW (1959) A note on two problems in connexion with graphs.
\newblock \emph{Numerische mathematik} 1(1): 269--271.

\bibitem[{Dobson and Bekris(2014)}]{dobson2014sparse}
Dobson A and Bekris KE (2014) Sparse roadmap spanners for asymptotically near-optimal motion planning.
\newblock \emph{The International Journal of Robotics Research} 33(1): 18--47.

\bibitem[{Dozat(2016)}]{dozat2016incorporating}
Dozat T (2016) Incorporating nesterov momentum into adam .

\bibitem[{Duarte et~al.(2023)Duarte, Lau, Pereira and Reis}]{duarte2023lstm}
Duarte FF, Lau N, Pereira A and Reis LP (2023) Lstm, convlstm, mdn-rnn and gridlstm memory-based deep reinforcement learning.
\newblock In: \emph{ICAART (2)}. pp. 169--179.

\bibitem[{Dunteman(1989)}]{dunteman1989principal}
Dunteman GH (1989) \emph{Principal components analysis}, volume~69.
\newblock Sage.

\bibitem[{Francis et~al.(2020)Francis, Faust, Chiang, Hsu, Kew, Fiser and Lee}]{francis2020long}
Francis A, Faust A, Chiang HTL, Hsu J, Kew JC, Fiser M and Lee TWE (2020) Long-range indoor navigation with prm-rl.
\newblock \emph{IEEE Transactions on Robotics} 36(4): 1115--1134.

\bibitem[{Fu et~al.(2022)Fu, Kumar, Agarwal, Qi, Malik and Pathak}]{fu2022coupling}
Fu Z, Kumar A, Agarwal A, Qi H, Malik J and Pathak D (2022) Coupling vision and proprioception for navigation of legged robots.
\newblock In: \emph{Proceedings of the IEEE/CVF Conference on Computer Vision and Pattern Recognition}. pp. 17273--17283.

\bibitem[{Gu and Dao(2023)}]{mamba}
Gu A and Dao T (2023) Mamba: Linear-time sequence modeling with selective state spaces.
\newblock \emph{arXiv preprint arXiv:2312.00752} .

\bibitem[{Gu et~al.(2020{\natexlab{a}})Gu, Dao, Ermon, Rudra and R{\'e}}]{gu2020hippo}
Gu A, Dao T, Ermon S, Rudra A and R{\'e} C (2020{\natexlab{a}}) Hippo: Recurrent memory with optimal polynomial projections.
\newblock \emph{Advances in neural information processing systems} 33: 1474--1487.

\bibitem[{Gu et~al.(2021)Gu, Goel and R{\'e}}]{gu2021efficiently}
Gu A, Goel K and R{\'e} C (2021) Efficiently modeling long sequences with structured state spaces.
\newblock \emph{arXiv preprint arXiv:2111.00396} .

\bibitem[{Gu et~al.(2020{\natexlab{b}})Gu, Gulcehre, Paine, Hoffman and Pascanu}]{gu2020improving}
Gu A, Gulcehre C, Paine T, Hoffman M and Pascanu R (2020{\natexlab{b}}) Improving the gating mechanism of recurrent neural networks.
\newblock In: \emph{International conference on machine learning}. PMLR, pp. 3800--3809.

\bibitem[{Handa et~al.(2014{\natexlab{a}})Handa, Whelan, McDonald and Davison}]{handa:etal:2014}
Handa A, Whelan T, McDonald J and Davison AJ (2014{\natexlab{a}}) A benchmark for rgb-d visual odometry, 3d reconstruction and slam.
\newblock \emph{ICRA} .

\bibitem[{Handa et~al.(2014{\natexlab{b}})Handa, Whelan, McDonald and Davison}]{handa2014benchmark}
Handa A, Whelan T, McDonald J and Davison AJ (2014{\natexlab{b}}) A benchmark for rgb-d visual odometry, 3d reconstruction and slam.
\newblock In: \emph{2014 IEEE international conference on Robotics and automation (ICRA)}. IEEE, pp. 1524--1531.

\bibitem[{Hart et~al.(1968)Hart, Nilsson and Raphael}]{Hart1968}
Hart P, Nilsson N and Raphael B (1968) A formal basis for the heuristic determination of minimum cost paths.
\newblock \emph{{IEEE} Transactions on Systems Science and Cybernetics} 4(2): 100--107.

\bibitem[{Hausknecht and Wagener(2022)}]{hausknecht2022consistent}
Hausknecht M and Wagener N (2022) Consistent dropout for policy gradient reinforcement learning.
\newblock \emph{arXiv preprint arXiv:2202.11818} .

\bibitem[{He et~al.(2024)He, Zhang, Xiao, He, Liu and Shi}]{He-RSS-24}
He T, Zhang C, Xiao W, He G, Liu C and Shi G (2024) {Agile But Safe: Learning Collision-Free High-Speed Legged Locomotion}.
\newblock In: \emph{Proceedings of Robotics: Science and Systems}. Delft, Netherlands.
\newblock \doi{10.15607/RSS.2024.XX.059}.

\bibitem[{Hochreiter and Schmidhuber(1997)}]{hochreiter1997long}
Hochreiter S and Schmidhuber J (1997) Long short-term memory.
\newblock \emph{Neural computation} 9(8): 1735--1780.

\bibitem[{Hoeller et~al.(2021)Hoeller, Wellhausen, Farshidian and Hutter}]{hoeller2021learning}
Hoeller D, Wellhausen L, Farshidian F and Hutter M (2021) Learning a state representation and navigation in cluttered and dynamic environments.
\newblock \emph{IEEE Robotics and Automation Letters} 6(3): 5081--5088.

\bibitem[{Huang et~al.(2023)Huang, Zhou, He and Lv}]{huang2023goal}
Huang W, Zhou Y, He X and Lv C (2023) Goal-guided transformer-enabled reinforcement learning for efficient autonomous navigation.
\newblock \emph{IEEE Transactions on Intelligent Transportation Systems} 25(2): 1832--1845.

\bibitem[{Hutter et~al.(2016)Hutter, Gehring, Jud, Lauber, Bellicoso, Tsounis, Hwangbo, Bodie, Fankhauser, Bloesch et~al.}]{hutter2016anymal}
Hutter M, Gehring C, Jud D, Lauber A, Bellicoso CD, Tsounis V, Hwangbo J, Bodie K, Fankhauser P, Bloesch M et~al. (2016) Anymal-a highly mobile and dynamic quadrupedal robot.
\newblock In: \emph{2016 IEEE/RSJ international conference on intelligent robots and systems (IROS)}. IEEE, pp. 38--44.

\bibitem[{Karaman and Frazzoli(2011)}]{karaman2011sampling}
Karaman S and Frazzoli E (2011) Sampling-based algorithms for optimal motion planning.
\newblock \emph{The international journal of robotics research} 30(7): 846--894.

\bibitem[{Kareer et~al.(2023)Kareer, Yokoyama, Batra, Ha and Truong}]{kareer2023vinl}
Kareer S, Yokoyama N, Batra D, Ha S and Truong J (2023) Vinl: Visual navigation and locomotion over obstacles.
\newblock In: \emph{2023 IEEE International Conference on Robotics and Automation (ICRA)}. IEEE, pp. 2018--2024.

\bibitem[{Karnan et~al.(2022)Karnan, Warnell, Xiao and Stone}]{karnan2022voila}
Karnan H, Warnell G, Xiao X and Stone P (2022) Voila: Visual-observation-only imitation learning for autonomous navigation.
\newblock In: \emph{2022 International Conference on Robotics and Automation (ICRA)}. IEEE, pp. 2497--2503.

\bibitem[{Kavraki et~al.(1996)Kavraki, Svestka, Latombe and Overmars}]{kavraki1996probabilistic}
Kavraki LE, Svestka P, Latombe JC and Overmars MH (1996) Probabilistic roadmaps for path planning in high-dimensional configuration spaces.
\newblock \emph{IEEE transactions on Robotics and Automation} 12(4): 566--580.

\bibitem[{Kuffner and LaValle(2000)}]{kuffner2000rrt}
Kuffner JJ and LaValle SM (2000) Rrt-connect: An efficient approach to single-query path planning.
\newblock In: \emph{Proceedings 2000 ICRA. Millennium conference. IEEE international conference on robotics and automation. Symposia proceedings (Cat. No. 00CH37065)}, volume~2. IEEE, pp. 995--1001.

\bibitem[{LaValle et~al.(2001)LaValle, Kuffner, Donald et~al.}]{lavalle2001rapidly}
LaValle SM, Kuffner JJ, Donald B et~al. (2001) Rapidly-exploring random trees: Progress and prospects.
\newblock \emph{Algorithmic and computational robotics: new directions} 5: 293--308.

\bibitem[{Lee et~al.(2024)Lee, Bjelonic, Reske, Wellhausen, Miki and Hutter}]{lee2024learning}
Lee J, Bjelonic M, Reske A, Wellhausen L, Miki T and Hutter M (2024) Learning robust autonomous navigation and locomotion for wheeled-legged robots.
\newblock \emph{Science Robotics} 9(89): eadi9641.

\bibitem[{Lee et~al.(2018)Lee, Lee, Lee and Shin}]{lee2018simple}
Lee K, Lee K, Lee H and Shin J (2018) A simple unified framework for detecting out-of-distribution samples and adversarial attacks.
\newblock \emph{Advances in neural information processing systems} 31.

\bibitem[{Lin et~al.(2017)Lin, Doll{\'a}r, Girshick, He, Hariharan and Belongie}]{lin2017feature}
Lin TY, Doll{\'a}r P, Girshick R, He K, Hariharan B and Belongie S (2017) Feature pyramid networks for object detection.
\newblock In: \emph{Proceedings of the IEEE conference on computer vision and pattern recognition}. pp. 2117--2125.

\bibitem[{Loquercio et~al.(2021)Loquercio, Kaufmann, Ranftl, M{\"u}ller, Koltun and Scaramuzza}]{loquercio2021learning}
Loquercio A, Kaufmann E, Ranftl R, M{\"u}ller M, Koltun V and Scaramuzza D (2021) Learning high-speed flight in the wild.
\newblock \emph{Science Robotics} 6(59): eabg5810.

\bibitem[{Ma et~al.(2024)Ma, Dai, Bai, Wang and Fu}]{ma2024rewrite}
Ma X, Dai X, Bai Y, Wang Y and Fu Y (2024) Rewrite the stars.
\newblock In: \emph{Proceedings of the IEEE/CVF Conference on Computer Vision and Pattern Recognition}. pp. 5694--5703.

\bibitem[{Matthis et~al.(2018)Matthis, Yates and Hayhoe}]{matthis2018gaze}
Matthis JS, Yates JL and Hayhoe MM (2018) Gaze and the control of foot placement when walking in natural terrain.
\newblock \emph{Current Biology} 28(8): 1224--1233.

\bibitem[{Mescheder et~al.(2019)Mescheder, Oechsle, Niemeyer, Nowozin and Geiger}]{mescheder2019occupancy}
Mescheder L, Oechsle M, Niemeyer M, Nowozin S and Geiger A (2019) Occupancy networks: Learning 3d reconstruction in function space.
\newblock In: \emph{Proceedings of the IEEE/CVF conference on computer vision and pattern recognition}. pp. 4460--4470.

\bibitem[{Mi et~al.(2024)Mi, Jiang, Luo and Gao}]{mi2024toward}
Mi JX, Jiang X, Luo L and Gao Y (2024) Toward explainable artificial intelligence: A survey and overview on their intrinsic properties.
\newblock \emph{Neurocomputing} 563: 126919.

\bibitem[{Miki et~al.(2022{\natexlab{a}})Miki, Lee, Hwangbo, Wellhausen, Koltun and Hutter}]{miki2022learning}
Miki T, Lee J, Hwangbo J, Wellhausen L, Koltun V and Hutter M (2022{\natexlab{a}}) Learning robust perceptive locomotion for quadrupedal robots in the wild.
\newblock \emph{Science robotics} 7(62): eabk2822.

\bibitem[{Miki et~al.(2022{\natexlab{b}})Miki, Wellhausen, Grandia, Jenelten, Homberger and Hutter}]{miki2022elevation}
Miki T, Wellhausen L, Grandia R, Jenelten F, Homberger T and Hutter M (2022{\natexlab{b}}) Elevation mapping for locomotion and navigation using gpu.
\newblock In: \emph{2022 IEEE/RSJ International Conference on Intelligent Robots and Systems (IROS)}. IEEE, pp. 2273--2280.

\bibitem[{Mittal et~al.(2023)Mittal, Yu, Yu, Liu, Rudin, Hoeller, Yuan, Singh, Guo, Mazhar, Mandlekar, Babich, State, Hutter and Garg}]{mittal2023orbit}
Mittal M, Yu C, Yu Q, Liu J, Rudin N, Hoeller D, Yuan JL, Singh R, Guo Y, Mazhar H, Mandlekar A, Babich B, State G, Hutter M and Garg A (2023) Orbit: A unified simulation framework for interactive robot learning environments.
\newblock \emph{IEEE Robotics and Automation Letters} 8(6): 3740--3747.
\newblock \doi{10.1109/LRA.2023.3270034}.

\bibitem[{Mohajerin and Rohani(2019)}]{mohajerin2019multi}
Mohajerin N and Rohani M (2019) Multi-step prediction of occupancy grid maps with recurrent neural networks.
\newblock In: \emph{Proceedings of the IEEE/CVF Conference on Computer Vision and Pattern Recognition}. pp. 10600--10608.

\bibitem[{Ortiz-Haro et~al.(2024)Ortiz-Haro, H{\"o}nig, Hartmann and Toussaint}]{ortiz2024idb}
Ortiz-Haro J, H{\"o}nig W, Hartmann VN and Toussaint M (2024) idb-a*: Iterative search and optimization for optimal kinodynamic motion planning.
\newblock \emph{IEEE Transactions on Robotics} .

\bibitem[{Pfeiffer et~al.(2017)Pfeiffer, Schaeuble, Nieto, Siegwart and Cadena}]{pfeiffer2017perception}
Pfeiffer M, Schaeuble M, Nieto J, Siegwart R and Cadena C (2017) From perception to decision: A data-driven approach to end-to-end motion planning for autonomous ground robots.
\newblock In: \emph{IEEE International Conference on Robotics and Automation (ICRA)}. IEEE, p. 1527–1533.

\bibitem[{Pinto et~al.(2017)Pinto, Andrychowicz, Welinder, Zaremba and Abbeel}]{pinto2017asymmetric}
Pinto L, Andrychowicz M, Welinder P, Zaremba W and Abbeel P (2017) Asymmetric actor critic for image-based robot learning.
\newblock \emph{arXiv preprint arXiv:1710.06542} .

\bibitem[{Radosavovic et~al.(2020)Radosavovic, Kosaraju, Girshick, He and Doll{\'a}r}]{radosavovic2020designing}
Radosavovic I, Kosaraju RP, Girshick R, He K and Doll{\'a}r P (2020) Designing network design spaces.
\newblock In: \emph{Proceedings of the IEEE/CVF conference on computer vision and pattern recognition}. pp. 10428--10436.

\bibitem[{Rudin et~al.(2022)Rudin, Hoeller, Bjelonic and Hutter}]{rudin2022advanced}
Rudin N, Hoeller D, Bjelonic M and Hutter M (2022) Advanced skills by learning locomotion and local navigation end-to-end.
\newblock In: \emph{2022 IEEE/RSJ International Conference on Intelligent Robots and Systems (IROS)}. IEEE, pp. 2497--2503.

\bibitem[{Ruiz-Serra et~al.(2022)Ruiz-Serra, White, Petrie, Kameneva and McCarthy}]{ruiz2022towards}
Ruiz-Serra J, White J, Petrie S, Kameneva T and McCarthy C (2022) Towards self-attention based visual navigation in the real world.
\newblock \emph{arXiv preprint arXiv:2209.07043} .

\bibitem[{Savinov et~al.(2018)Savinov, Dosovitskiy and Koltun}]{savinov2018semi}
Savinov N, Dosovitskiy A and Koltun V (2018) Semi-parametric topological memory for navigation.
\newblock \emph{arXiv preprint arXiv:1803.00653} .

\bibitem[{Schulman et~al.(2017)Schulman, Wolski, Dhariwal, Radford and Klimov}]{schulman2017proximal}
Schulman J, Wolski F, Dhariwal P, Radford A and Klimov O (2017) Proximal policy optimization algorithms.
\newblock \emph{arXiv preprint arXiv:1707.06347} .

\bibitem[{Shah et~al.(2022)Shah, Sridhar, Bhorkar, Hirose and Levine}]{shah2022gnm}
Shah D, Sridhar A, Bhorkar A, Hirose N and Levine S (2022) Gnm: A general navigation model to drive any robot.
\newblock \emph{arXiv preprint arXiv:2210.03370} .

\bibitem[{Shah et~al.(2023)Shah, Sridhar, Dashora, Stachowicz, Black, Hirose and Levine}]{shah2023vint}
Shah D, Sridhar A, Dashora N, Stachowicz K, Black K, Hirose N and Levine S (2023) Vi{NT}: A foundation model for visual navigation.
\newblock In: \emph{7th Annual Conference on Robot Learning}.

\bibitem[{Shi et~al.(2019)Shi, Shi, Xu and Hwang}]{shi2019end}
Shi H, Shi L, Xu M and Hwang KS (2019) End-to-end navigation strategy with deep reinforcement learning for mobile robots.
\newblock \emph{IEEE Transactions on Industrial Informatics} 16(4): 2393--2402.

\bibitem[{Siami-Namini et~al.(2019)Siami-Namini, Tavakoli and Namin}]{siami2019performance}
Siami-Namini S, Tavakoli N and Namin AS (2019) The performance of lstm and bilstm in forecasting time series.
\newblock In: \emph{2019 IEEE International conference on big data (Big Data)}. IEEE, pp. 3285--3292.

\bibitem[{Surmann et~al.(2020)Surmann, Jestel, Marchel, Musberg, Elhadj and Ardani}]{surmann2020deep}
Surmann H, Jestel C, Marchel R, Musberg F, Elhadj H and Ardani M (2020) Deep reinforcement learning for real autonomous mobile robot navigation in indoor environments.
\newblock \emph{arXiv preprint arXiv:2005.13857} .

\bibitem[{Sutskever et~al.(2014)Sutskever, Vinyals and Le}]{sutskever2014sequence}
Sutskever I, Vinyals O and Le QV (2014) Sequence to sequence learning with neural networks.
\newblock \emph{Advances in neural information processing systems} 27.

\bibitem[{Truong et~al.(2021)Truong, Yarats, Li, Meier, Chernova, Batra and Rai}]{truong2021learning}
Truong J, Yarats D, Li T, Meier F, Chernova S, Batra D and Rai A (2021) Learning navigation skills for legged robots with learned robot embeddings.
\newblock In: \emph{2021 IEEE/RSJ International Conference on Intelligent Robots and Systems (IROS)}. IEEE, pp. 484--491.

\bibitem[{Wang et~al.(2020)Wang, Zhu, Wang, Hu, Qiu, Wang, Hu, Kapoor and Scherer}]{wang2020tartanair}
Wang W, Zhu D, Wang X, Hu Y, Qiu Y, Wang C, Hu Y, Kapoor A and Scherer S (2020) Tartanair: A dataset to push the limits of visual slam.
\newblock In: \emph{2020 IEEE/RSJ International Conference on Intelligent Robots and Systems (IROS)}. IEEE, pp. 4909--4916.

\bibitem[{Wang et~al.(2025)Wang, Huang, Sun, Yan, Xing, Tu and Li}]{wang2025uniocc}
Wang Y, Huang X, Sun X, Yan M, Xing S, Tu Z and Li J (2025) Uniocc: A unified benchmark for occupancy forecasting and prediction in autonomous driving.
\newblock \emph{arXiv preprint arXiv:2503.24381} .

\bibitem[{Webb and Berg(2012)}]{webb2012kinodynamic}
Webb DJ and Berg Jvd (2012) Kinodynamic rrt*: Optimal motion planning for systems with linear differential constraints.
\newblock \emph{arXiv preprint arXiv:1205.5088} .

\bibitem[{Weerakoon et~al.(2022)Weerakoon, Sathyamoorthy, Patel and Manocha}]{weerakoon2022terp}
Weerakoon K, Sathyamoorthy AJ, Patel U and Manocha D (2022) Terp: Reliable planning in uneven outdoor environments using deep reinforcement learning.
\newblock In: \emph{2022 International Conference on Robotics and Automation (ICRA)}. IEEE, pp. 9447--9453.

\bibitem[{Wei et~al.(2023)Wei, Zhao, Zheng, Zhu, Zhou and Lu}]{wei2023surroundocc}
Wei Y, Zhao L, Zheng W, Zhu Z, Zhou J and Lu J (2023) Surroundocc: Multi-camera 3d occupancy prediction for autonomous driving.
\newblock In: \emph{Proceedings of the IEEE/CVF International Conference on Computer Vision}. pp. 21729--21740.

\bibitem[{Wellhausen and Hutter(2023)}]{wellhausen2023artplanner}
Wellhausen L and Hutter M (2023) Artplanner: Robust legged robot navigation in the field.
\newblock \emph{Field Robotics} 3: 413--434.

\bibitem[{Wijmans et~al.(2019)Wijmans, Kadian, Morcos, Lee, Essa, Parikh, Savva and Batra}]{wijmans2019dd}
Wijmans E, Kadian A, Morcos A, Lee S, Essa I, Parikh D, Savva M and Batra D (2019) Dd-ppo: Learning near-perfect pointgoal navigators from 2.5 billion frames.
\newblock \emph{arXiv preprint arXiv:1911.00357} .

\bibitem[{Wijmans et~al.(2023)Wijmans, Savva, Essa, Lee, Morcos and Batra}]{wijmans2023emergence}
Wijmans E, Savva M, Essa I, Lee S, Morcos AS and Batra D (2023) Emergence of maps in the memories of blind navigation agents.
\newblock \emph{AI Matters} 9(2): 8--14.

\bibitem[{Wu et~al.(2021)Wu, Wang, Esfahani and Yuan}]{wu2021learn}
Wu K, Wang H, Esfahani MA and Yuan S (2021) Learn to navigate autonomously through deep reinforcement learning.
\newblock \emph{IEEE Transactions on Industrial Electronics} 69(5): 5342--5352.

\bibitem[{Xie et~al.(2025)Xie, Cao, Zhang, Zhang, Wang and Xu}]{xie2025meta}
Xie Z, Cao J, Zhang Q, Zhang J, Wang C and Xu R (2025) The meta-representation hypothesis.
\newblock \emph{arXiv preprint arXiv:2501.02481} .

\bibitem[{Yang et~al.(2022{\natexlab{a}})Yang, Cao, Zhu, Oh and Zhang}]{yang2022far}
Yang F, Cao C, Zhu H, Oh J and Zhang J (2022{\natexlab{a}}) Far planner: Fast, attemptable route planner using dynamic visibility update.
\newblock In: \emph{2022 ieee/rsj international conference on intelligent robots and systems (iros)}. IEEE, pp. 9--16.

\bibitem[{Yang et~al.(2022{\natexlab{b}})Yang, Zhang, Hansen, Xu and Wang}]{yang2022learning}
Yang R, Zhang M, Hansen N, Xu H and Wang X (2022{\natexlab{b}}) Learning vision-guided quadrupedal locomotion end-to-end with cross-modal transformers.
\newblock In: \emph{International Conference on Learning Representations}.

\bibitem[{Zeng et~al.(2024)Zeng, Zhang, Ehsani, Hendrix, Salvador, Herrasti, Girshick, Kembhavi and Weihs}]{zeng2024poliformer}
Zeng KH, Zhang Z, Ehsani K, Hendrix R, Salvador J, Herrasti A, Girshick R, Kembhavi A and Weihs L (2024) Poliformer: Scaling on-policy rl with transformers results in masterful navigators.
\newblock \emph{arXiv preprint arXiv:2406.20083} .

\bibitem[{Zhang et~al.(2024)Zhang, Jin, Frey, Rudin, Mattamala, Cadena and Hutter}]{zhang2024resilient}
Zhang C, Jin J, Frey J, Rudin N, Mattamala M, Cadena C and Hutter M (2024) Resilient legged local navigation: Learning to traverse with compromised perception end-to-end.
\newblock In: \emph{2024 IEEE International Conference on Robotics and Automation (ICRA)}. IEEE, pp. 34--41.

\bibitem[{Zhu et~al.(2017)Zhu, Mottaghi, Kolve, Lim, Gupta, Fei-Fei and Farhadi}]{zhu2017target}
Zhu Y, Mottaghi R, Kolve E, Lim JJ, Gupta A, Fei-Fei L and Farhadi A (2017) Target-driven visual navigation in indoor scenes using deep reinforcement learning.
\newblock In: \emph{2017 IEEE international conference on robotics and automation (ICRA)}. IEEE, pp. 3357--3364.

\end{thebibliography}


  \newpage 

  \appendix
  \setcounter{figure}{0}                         
  \renewcommand{\thefigure}{\thesection.\arabic{figure}}
  \setcounter{algorithm}{0}                         
  \renewcommand{\thealgorithm}{\thesection.\arabic{algorithm}}
  \setcounter{table}{0}                         
  \renewcommand{\thetable}{\thesection.\arabic{table}}

  \section*{Appendix}

  \section{Training Details for Spatial-Temporal Memorization Task}
  \label{appendix: spatial-temporal-details}

\begin{figure}
    \centering
    \includegraphics[width=0.4\textwidth]{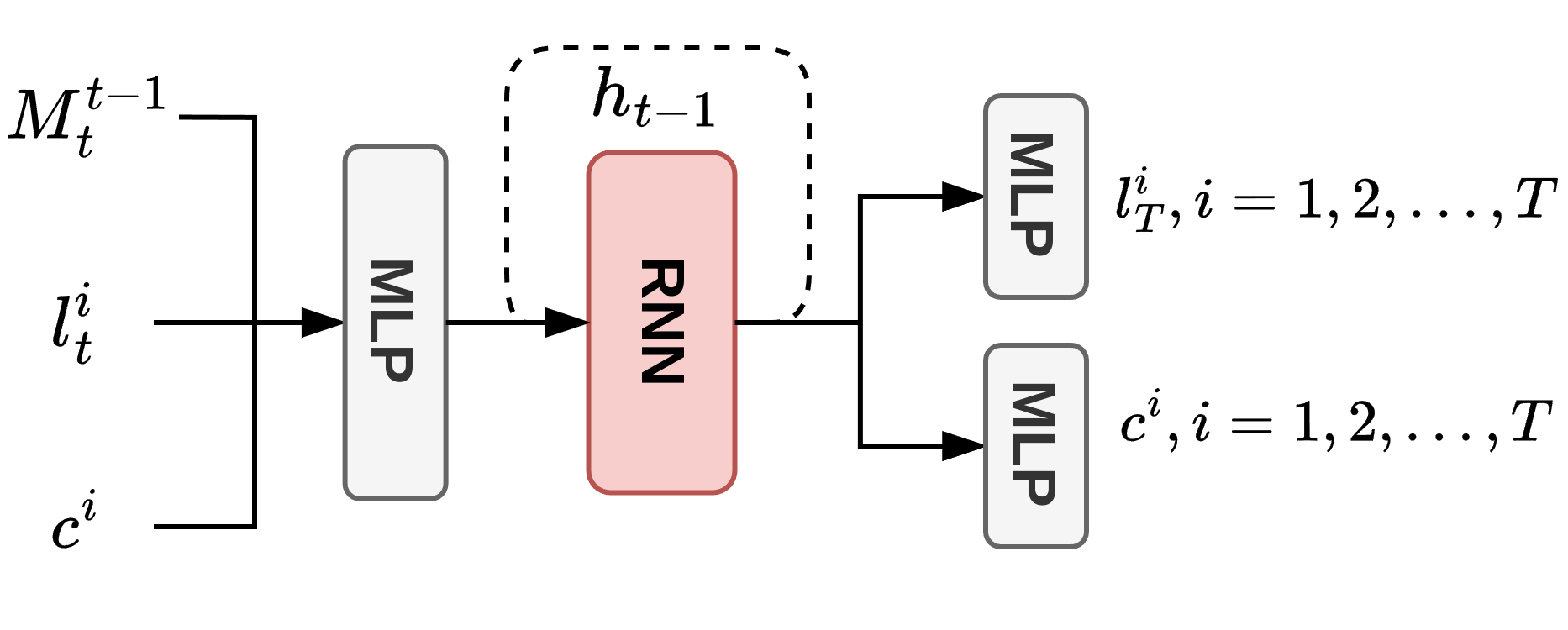}
      \caption{Network architecture for the spatial and temporal memorization task.
      At each step $t$, the agent receives landmark coordinates $l^{i}_{t}$, a binary label $c^{i}$, and ego-centric motion $M^{t-1}_{t}$. 
      These inputs are concatenated, passed through an MLP layer, and processed by a recurrent unit. After $T$ steps, the MLP head is tasked with recalling, from the final hidden state of the recurrent unit, all observed landmark positions $\{l^{i}_{T}\}$
      with respect to the final frame $T$ (spatial task) and sequentially predicting the associated labels $\{c^{i}\}$ (temporal task).}

    \label{fig:spatial-temporal-train}
  \end{figure}

  To evaluate the spatial-temporal memorization capabilities of different recurrent neural network architectures, we conduct a case study using an abstract version of the spatial-temporal memorization task shown in Figure~\ref{fig:spatial_temporal_mapping}
  and Figure~\ref{fig:spatial_mapping_result}. 
  Below, we provide the training details for this task.
  The task simulates a scenario where, at each time step $t$, the recurrent agent receives the following inputs:
  \begin{itemize}
    \item The coordinates of an observed landmark, $l^{i}_{t}$, represented in the robot's current frame.

    \item A binary categorical label, $c^{i}$, associated with the landmark.

    \item The ego-centric motion transformation matrix, $M^{t-1}_{t}$, representing the transformation from the previous frame to the current frame.
  \end{itemize}
  These inputs are concatenated into a 1D vector, passed through a Multi-Layer Perceptron (MLP) layer, and then fed into the recurrent unit. 
  After $T$ steps, the output MLP layer is tasked with:
  \begin{itemize}
    \item Regressing all observed landmark coordinates $\{l^{i}_{T}, i=1,2,\ldots,T\}$ with respect to the robot's frame at the final time step $T$ (spatial memorization task).

    \item Predicting all observed binary labels $\{c^{i}, i=1,2,\ldots,T\}$ associated with the landmarks, which are independent of the observation frame and depend only on the sequential order of observations (temporal memorization task).
  \end{itemize}
  Figure~\ref{fig:spatial-temporal-train} illustrates the network structure used for training. 
  The spatial task is optimized using the Mean Squared Error (MSE) loss, while the temporal task is optimized using the Binary Cross-Entropy (BCE) loss. 
  The network is trained using the Nesterov Momentum Adam optimizer~\citep{dozat2016incorporating} with an initial learning rate of $2 \times 10^{-3}$, which is reduced to $4 \times 10^{-4}$ after 800 epochs, continuing until 1000 epochs are completed.

  To ensure the recurrent network does not overfit or memorize specific patterns of observed landmarks and associated ego-motion trajectories, the following randomization is applied:
  \begin{itemize}
    \item The ego-motion $M^{t-1}_{t}$ for each step is uniformly sampled, with translation in the range $[-2, 2]$ meters and orientation in the range $[-\pi, \pi]$ radians.

    \item The observed landmark coordinates $\{l^{i}_{T}, i=1,2,\ldots,T\}$ are uniformly sampled within the range $[-5, 5]$ meters relative to the observation frame, and the binary categorical labels $\{c^{i}, i=1,2,\ldots,T\}$ are uniformly sampled from the set $\{0, 1\}$.
  \end{itemize}

  \section{Parallelizable Stereo Depth Perception Noise Implementation}
  \label{appendix: stereo-depth-noise-pseudocode}

  The following pseudocode~\ref{alg:depth_noise} injects synthetic stereo depth images with edge noise,
  filling noise, and round noise to simulate realistic sensor artifacts. 
  This implementation is used both during pretraining on synthetic depth data and during online reinforcement learning to better mimic real-world sensor imperfections.

  \textbf{Inputs:} The function takes \texttt{depth} ($\mathbb{R}^{B \times H
  \times W}$), a batch of raw depth image tensors, where $B$ indicates the batch size, and $H$ and $W$ represent the spatial dimensions of the depth image, specifically its height and width.

  \textbf{Parameters:} The parameters include \texttt{f} (focal length of the camera), \texttt{b} (baseline between stereo cameras), \texttt{filt\_size} (local window size for filtering), \texttt{$\tau$\_min} and \texttt{$\tau$\_max} (edge noise threshold range), \texttt{$\rho$\_min} and \texttt{$\rho$\_max} (pseudo-stereo match probability range), and \texttt{invalid\_disp} (value to mark dropped disparities).

  \begin{algorithm}
    \caption{Stereo Depth Noise Algorithm}
    \label{alg:depth_noise}
    \begin{algorithmic}
      \REQUIRE $depth \in \mathbb{R}^{B \times H \times W}$

      \STATE $(K_{mean}, K_{sub}) \leftarrow$ \textsc{ComputeKernels}($filt\_size$)
      \STATE $disp \leftarrow \frac{f \cdot b}{depth}$ \STATE $filtered\_disp \leftarrow$
      \textsc{FilterDisp}($disp$, $K_{mean}$, $K_{sub}$) \STATE $filtered\_depth
      \leftarrow \frac{f \cdot b}{filtered\_disp}$
      \RETURN $filtered\_depth$

      \STATE \STATE \textbf{function} \textsc{ComputeKernels}($s$) \STATE
      Compute kernel $K_{mean}$ \STATE Compute kernel $K_{sub}$ \RETURN
      $(K_{mean}, K_{sub})$

      \STATE \STATE \textbf{function} \textsc{FilterDisp}($disp$, $K_{mean}$, $K_{sub}$)
      \STATE $\rho \sim \text{Uniform}(\rho\_min, \rho\_max)$ \STATE
      $R \sim \text{BernoulliMask}(\text{shape}=disp, p=\rho)$ \STATE
      $m \leftarrow \text{Conv2D}(disp, K_{mean})$ \STATE
      $\tau \sim \text{Uniform}(\tau\_min, \tau\_max)$ \STATE
      $M \leftarrow (|disp - m| < \tau) \wedge R$ \STATE
      $v \leftarrow \text{Quantize}(disp)$ \STATE
      $masked \leftarrow \text{if }M \text{ then }v \text{ else }invalid\_disp$
      \STATE $num \leftarrow \text{Conv2D}(masked, K_{sub})$ \STATE
      $den \leftarrow \text{Conv2D}(M, K_{sub}) + \epsilon$ \STATE
      $fil led \leftarrow \text{if }den > 0 \text{ then }\frac{num}{den}\text{ else
      }m asked$
      \RETURN $filled$
    \end{algorithmic}
  \end{algorithm}

  \section{Training Details for Navigation with Reinforcement Learning}
  \label{appendix: rl-training-details}

  This section provides the training and parameter details for the end-to-end navigation task using reinforcement learning. 
  We utilize an asymmetric actor-critic setup, training with the \textit{NVIDIA IsaacLab} simulation framework. The actor processes noisy observations, including depth input with noise augmentation, using the spatial attention-based recurrent structure. 
  In contrast, the critic has access to additional 360-degree height scan information alongside the depth input. 
  These inputs are processed separately through two attention layers, which are then concatenated before being passed to the SRU unit. 
  Unlike the actor, the critic does not use noise-augmented observations. 
  To handle the height scan input for the critic, we pretrain a height scan encoder with the same architecture as the depth encoder, using height scan images collected from the RL simulation environments. To improve the network's generalization for handling large distance values in long-range navigation, we convert the goal position $p_t \in \mathbb{R}^3$ into a unit directional vector and a log-transformed distance value. 
  This transformation allows the network to generalize better to varying distances. 


  To enhance sim-to-real transfer, we introduce randomization noise to the actor's observations. 
  The noise parameters applied during training are summarized in Table~\ref{tab:noise_randomization}. 
  The critic, in contrast, receives clean observations without any noise or delay, ensuring stable and accurate value estimation during training. 
  \begin{table}[h]
    \centering
    \begin{tabular}{l|c}
      \toprule
      \textbf{Observation Parameter} & \textbf{Noise Range $(\mathcal{U})$} \\
      \midrule
      Linear Velocity ($v_t$)        & $\pm 0.2$ m/s        \\
      Angular Velocity ($\omega_t$)  & $\pm 0.1$ rad/s      \\
      Projected Gravity ($n_t$)      & $\pm 0.1$            \\
      Goal Position ($p_t$)          & $\pm 0.5$ m, $\pm 0.1$ rad \\
      Observation Delay              & 0 ms to 600 ms       \\ 
      \bottomrule
    \end{tabular}
    \vspace{0.5em}
    \caption{Noise parameters (uniform distribution $\mathcal{U}$) applied to the actor's observations during RL training for navigation.}
    \label{tab:noise_randomization}
  \end{table}
  These randomization strategies, combined with the pretraining of encoders and the use of spatial attention-based recurrent structures, enable robust training and effective sim-to-real transfer for the navigation policy.


\end{document}